\RequirePackage{fix-cm}
\documentclass[twocolumn]{svjour3}          

\smartqed
\usepackage{graphicx}
\usepackage{epsfig}
\usepackage{graphicx}
\usepackage{amsmath}
\usepackage{amssymb}
\usepackage{times}
\usepackage{subfigure}
\usepackage{epsfig}
\usepackage{multirow}
\usepackage{graphics}
\usepackage{graphicx}
\usepackage{amsmath}
\usepackage{amssymb}
\usepackage{wrapfig}
\usepackage{enumerate}
\usepackage{setspace}
\usepackage{verbatim}
\usepackage{url}
\usepackage{bm}
\usepackage{amssymb}
\usepackage{color}
\usepackage{multirow}
\usepackage{subfigure}
\usepackage{natbib}

\begin{document}

\title{A Multi-View Embedding Space for\\Modeling Internet Images, Tags, and their Semantics}


\author{Yunchao~Gong \and
        Qifa Ke\and
        Michael Isard\and
        Svetlana~Lazebnik}

\institute{Y. Gong \at
              Department of Computer Science\\
              University of North Carolina at Chapel Hill\\
              \email{yunchao@cs.unc.edu}           \\
              Project website: http://www.unc.edu/$^\sim$yunchao/crossmodal.htm
           \and
           Q. Ke, M. Isard \at
           Microsoft Research Silicon Valley,
           Mountain View, CA\\
           \email{qke@microsoft.com,~misard@microsoft.com}
                      \and
           S. Lazebnik \at
           Department of Computer Science\\
           University of Illinois at Urbana-Champaign\\
           \email{slazebni@illinois.edu}\\
}

\date{~}

\maketitle

\begin{abstract}
This paper investigates the problem of modeling Internet images and associated text or tags
for tasks such as image-to-image search, tag-to-image search, and image-to-tag search (image annotation).
We start with {\em canonical correlation analysis (CCA)}, a popular and successful
approach for mapping visual and textual features to the same latent space,
and incorporate a third view capturing high-level image semantics, represented
either by a single category or multiple non-mutually-exclusive concepts.
We present two ways to train the three-view embedding: supervised, with the third view
coming from ground-truth labels or search keywords; and unsupervised, with semantic themes automatically
obtained by clustering the tags. 
To ensure high accuracy for retrieval tasks while keeping the learning process scalable,
we combine multiple strong visual features and use explicit nonlinear kernel mappings to efficiently approximate kernel CCA.
To perform retrieval, we use a specially designed similarity function in the embedded space, which substantially outperforms the
Euclidean distance. The resulting system produces compelling qualitative results and outperforms
a number of two-view baselines on retrieval tasks on three large-scale Internet image datasets.

\end{abstract}

\section{Introduction}

\begin{figure}
\includegraphics[width=\columnwidth]{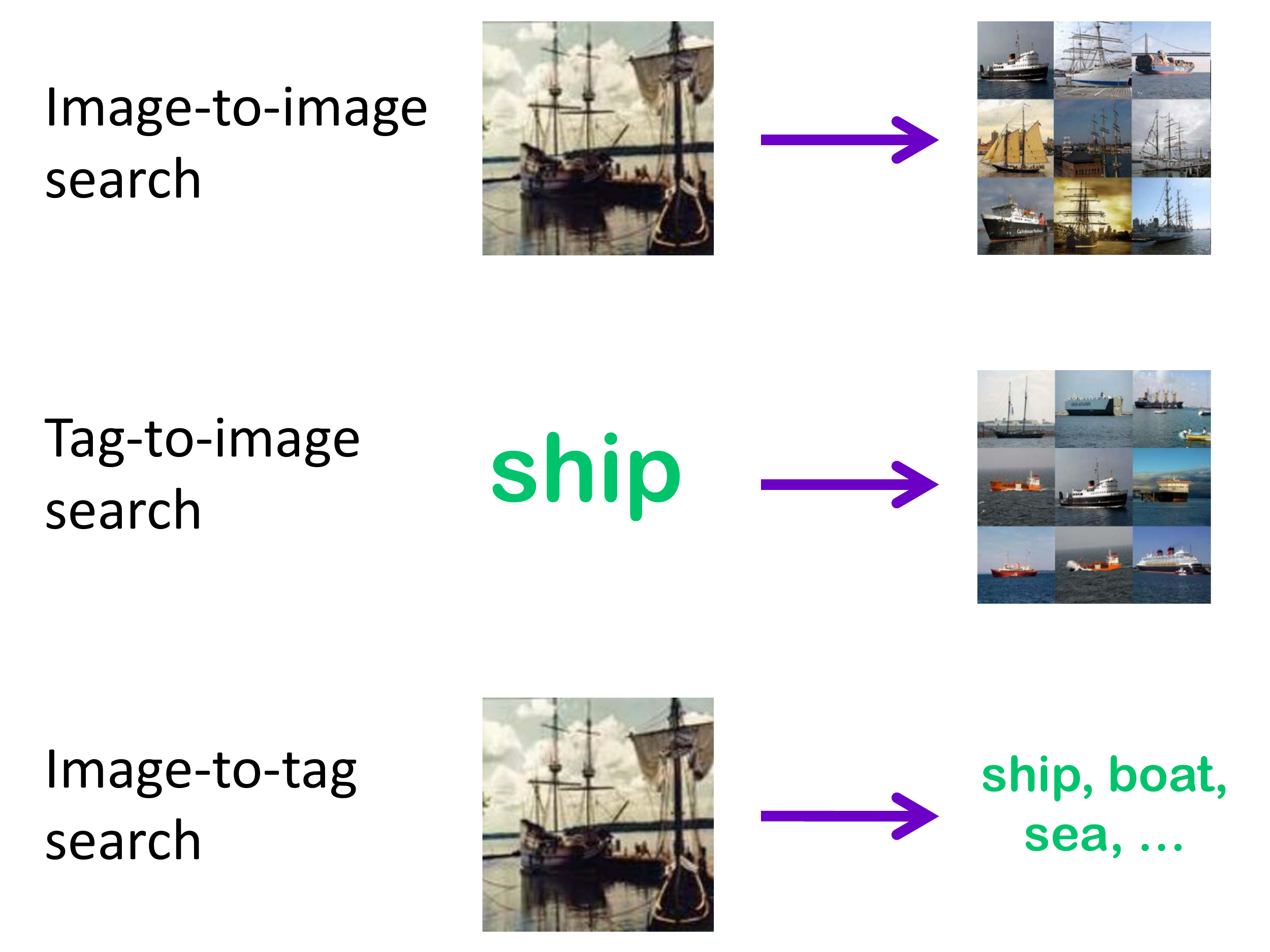}
\caption{Retrieval scenarios considered in this paper. Top: Given a query image, retrieve similar images from the database. Middle: Given a search keyword or tag, retrieve relevant images. Bottom: Given a query image, retrieve keywords or tags describing this image (automatic image annotation).\label{fig:retrieval-scenarios}}
\end{figure}

\begin{figure*}[!hbt] 
   \centering
   \includegraphics[width=6.7in, clip=true, trim=0mm 40mm 17mm
45mm]{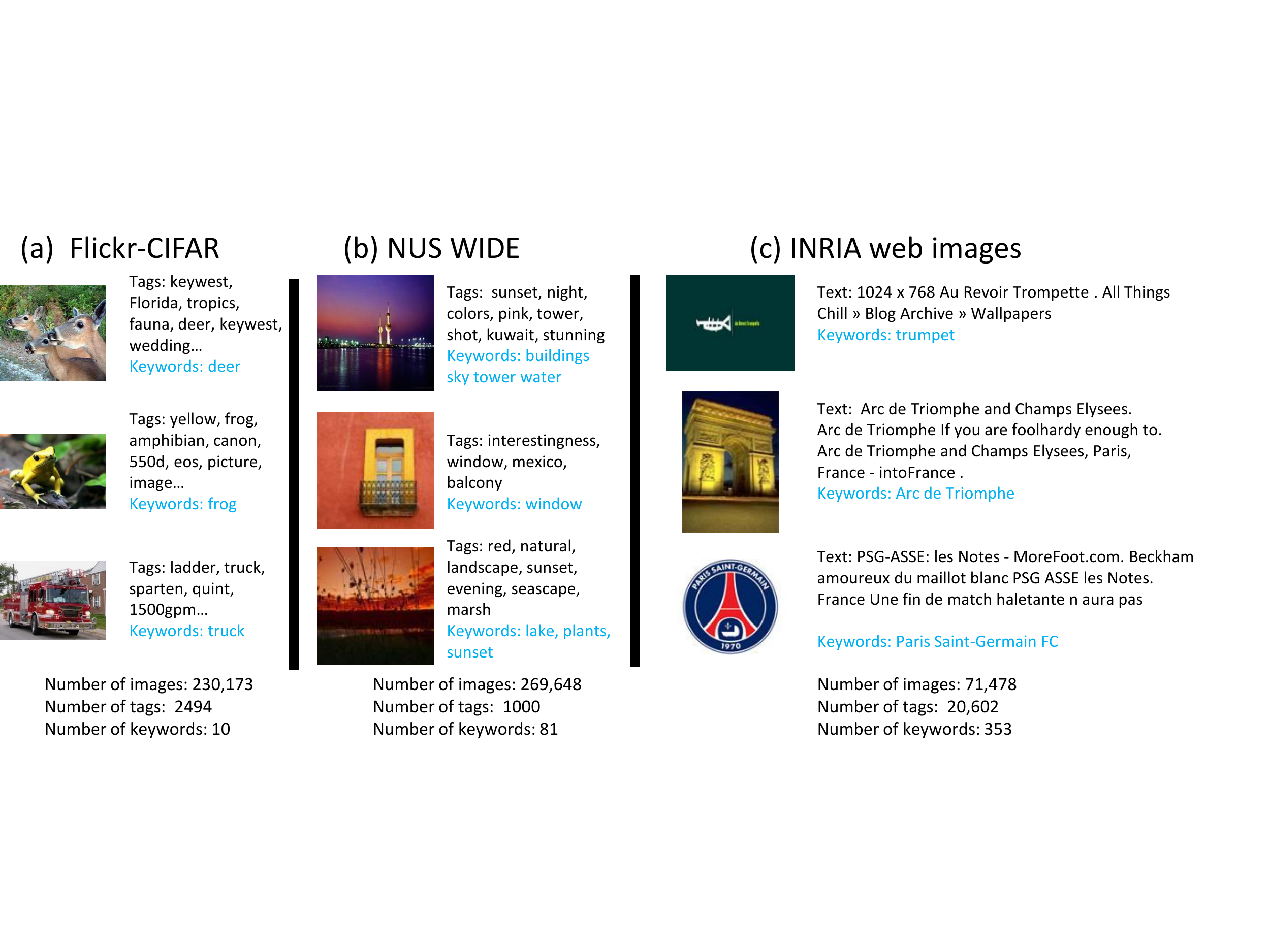}
\caption{An overview of the Internet image datasets used in this paper. Each image has three views associated with it: the visual features; the text or tags; and the semantics or ground-truth keywords. For Flickr-CIFAR dataset (collected by ourselves as described in Section \ref{dataset}) and INRIA-Websearch dataset~\citep{Krapac10}, each image only has one ground-truth keyword. For the NUS-WIDE dataset~\citep{nus-wide-civr09}, each image has multiple keywords. \label{data}}
\end{figure*}

The goal of this work is modeling the statistics of images and associated textual data in large-scale Internet photo collections
in order to enable a variety of retrieval scenarios: similarity-based image search, keyword-based image search, and automatic image annotation (Figure \ref{fig:retrieval-scenarios}).
Practical models for these tasks must meet several requirements.
First, they must be {\em accurate}, which is a big challenge given that the
imagery is extremely heterogeneous and user-provided annotations are noisy.
Second, they must be {\em scalable} to millions of images. Third, they must be {\em flexible},
accommodating {\em cross-modal} retrieval tasks such as tag-to-image or image-to-tag search
in the same framework, and enabling, for example, tag-based search of images without any tags.

Several promising recent approaches for modeling images and associated text~\citep{gong11,cca_survey,Hwang10,hwang2011ijcv,Rasiwasia10} rely on {\em canonical correlation analysis} (CCA), a classic technique that maps two views, given by visual and and textual features, into a common latent space where the correlation between the two views is maximized~\citep{Hotelling36}. This space is {\em cross-modal}, in the sense that embedded vectors representing visual and textual information are treated as the same class of citizens, and thus image-to-image, text-to-image, and image-to-text retrieval tasks can in principle all be handled in exactly the same way.

While CCA is very attractive in its simplicity and flexibility, existing CCA-based approaches have several shortcomings. In particular, the works cited above use classic two-view CCA, which only considers the direct correlations between images and corresponding textual feature vectors. However, as we will show in this paper, significant improvements can be obtained by considering a third view with which the first two are correlated -- that of the underlying semantics of the image.

In this work, we use the term ``semantics'' to refer to high-level labels or topics that characterize the content of an image for the sake of a given application.
For concrete examples, consider Figure \ref{data}, which illustrates the three-view datasets used in our experiments.
In these datasets, the semantic view of an image consists of one or more ground-truth {\em keywords}.
Even though our approach does not rely on a probabilistic generative model,
we can think of the other two views, i.e., {\em visual features} and {\em tags/text}, as being stochastically generated based on the keywords. In particular, the tags tend to come from a larger vocabulary than the keywords and they tend to be noisier.
 As in Figure \ref{data} (a), the semantics of an image may be given by a single object category (``deer''), while the user-provided tags may include a number of additional terms correlated with that category (``keywest, Florida, tropics, fauna, wedding'' etc.). Alternatively, the semantics might be given by multiple keywords corresponding to objects, scene types, or attributes. Thus, as in Figure \ref{data} (b), an image may be annotated by multiple ground-truth keywords ``buildings, sky, tower, water'' and tags ``sunset, night, colors, pink, tower, shot, kuwait, stunning.'' Or, as in Figure \ref{data} (c), the semantics may be given by the name of a logo or landmark, and the text may be taken from the surrounding webpage, and may or may not explicitly mention the ground-truth keyword.

\begin{figure*}[t] 
   \centering
   \includegraphics[width=6.2in, trim=0mm 0mm 10mm -10mm]{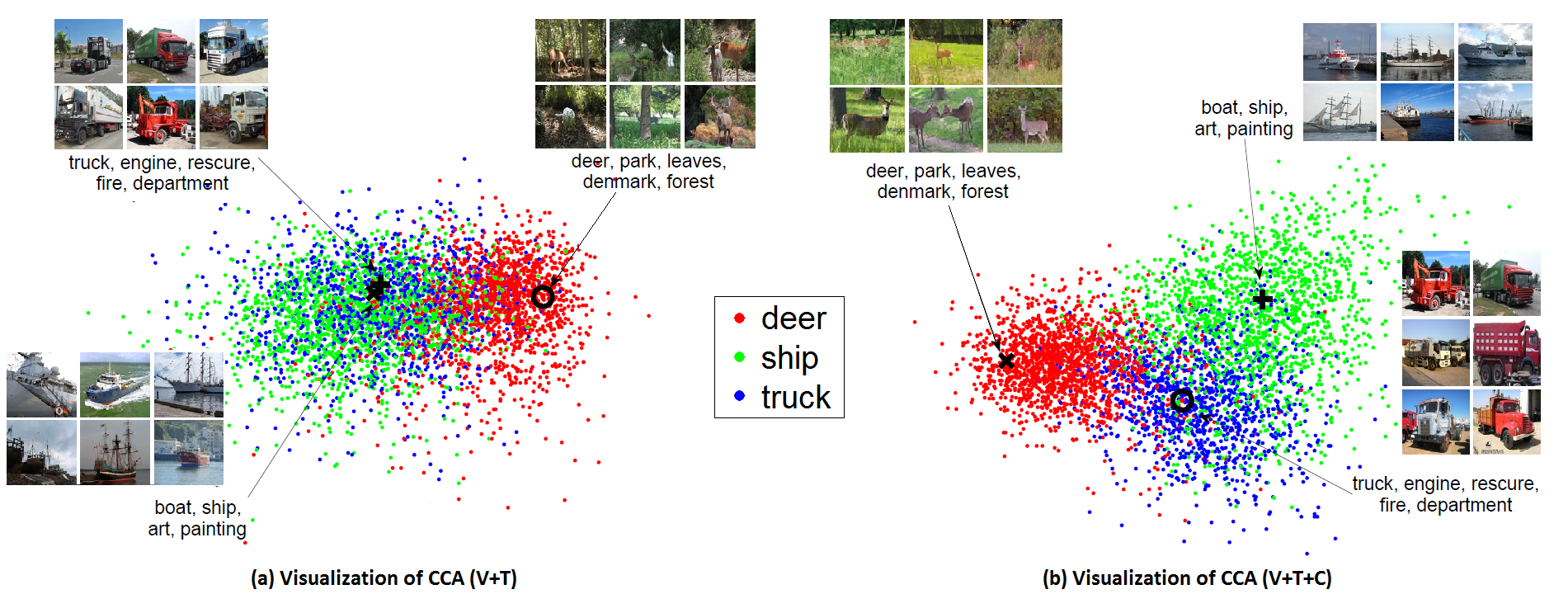}
   \caption{A visualization of the first two directions of the common latent space for (a) standard two-view CCA and (b) our proposed three-view CCA model. Different colors indicate different image categories (though note that category information is not used in learning the three-view embedding). Black points indicate sample tag queries, and the corresponding images are their nearest neighbors in the latent space.}
   \label{2d}
\end{figure*}

In this paper, we present a three-view CCA model that explicitly incorporates the high-level semantic information as a third view. The difference between the standard two-view CCA and our proposed three-view embedding is visualized in Figure \ref{2d}. In the two-view embedding space (Figure \ref{2d} (a)), which is produced by maximizing the correlations between visual features and the corresponding tag features, images from different classes are very mixed. On the other hand, the three-view embedding (Figure \ref{2d} (b)) provides a much better separation between the classes. As our experiments will confirm, a third semantic view -- which may be derived from a variety of sources -- is capable of considerably increasing the accuracy of retrieval on very diverse datasets.

In all the examples of Figure \ref{data}, the ground-truth semantic keywords are defined ahead of time and accurately annotated for the express purpose of training recognition algorithms. However, in most realistic situations, it is easy to gather noisy text and tags, but not so easy to get at the underlying semantics. Fortunately, we will show that even in cases when clean ground-truth annotation for the third view is unavailable, it is still possible to learn a better embedding for the photo collection by representing the semantics explicitly. In some cases, we can get an informative additional signal from search keywords. For example, if we retrieve a number of images together with their tags from Flickr using a search for ``frog,'' then knowing the original search keyword gives us additional modeling power even if many of these images do not actually depict frogs. Furthermore, if ground truth category or search keyword information is absent completely, we will demonstrate that an effective third view can be derived in an unsupervised way by clustering the noisy tag vectors constituting the second view. This approach is inspired by cluster-based information retrieval~\citep{Wei06} and the ``cluster assumption'' in semi-supervised learning \citep{Chapelle03}. In effect, the tag clustering can be thought of as ``reconstructing'' or ``recovering'' the absent topics or distinct types of image content.

To obtain high retrieval accuracy, most modern methods have found it necessary to combine multiple high-dimensional visual features, each of which may come with a different similarity or kernel function. Retrieval approaches of~\cite{Hwang10,hwang2011ijcv,Yakhnenko09} accomplish this combination using nonlinear kernel CCA (KCCA)~\citep{Bach02,cca_survey}, but the standard KCCA formulation scales {\em cubically} in the number of images in the dataset. Instead of KCCA, we use a scalable approximation scheme based on efficient explicit kernel mapping followed by linear dimensionality reduction and linear CCA. Finally, we specifically design a similarity function suitable for our learned latent embedding, and show that it achieves significant improvement over the Euclidean distance. Experiments on the three large-scale datasets of Figure \ref{data} show the promise of the proposed approach.

The following is a preview of the structure and main contributions of this paper:
\begin{itemize}
\item A novel {\em three-view} CCA framework that explicitly incorporates the dependence of visual features and text on the underlying image semantics
(Section \ref{sec:three_view_cca}).
\item A similarity function specially adapted to CCA that improves the accuracy of retrieval in the embedded space (Section \ref{metric}).
\item Scalable yet discriminative representations for the visual and textual views based on multiple feature combination, explicit kernel mappings, and linear dimensionality reduction (Sections \ref{sec:visual_kernel} and \ref{sec:tag_approximation}).
\item Two methods for instantiating the third semantic view: {\em supervised}, or derived from ground-truth annotations by unsupervised clustering;
and {\em unsupervised}, or derived by clustering the tag vectors from the second (textual) view. In both cases, our experiments confirm that adding the
third view helps to improve retrieval accuracy. For the unsupervised case, we perform a comparative evaluation of
several tag clustering methods from the literature (Section \ref{topics}).
\item Extensive evaluation of the proposed three-view models on three tasks -- image-to-image, tag-to-image, and image-to-tag search. Section \ref{dataset} will give an overview of our experimental protocol, and Sections \ref{sec:evaluation}-\ref{sec:evaluation3} will present results on the three large-scale datasets introduced in Figure \ref{data}.
\end{itemize}

\section{Related Work\label{previous}}

In the vision and multimedia communities, jointly modeling images and text has been an active research area. This section gives a non-exhaustive survey of several important lines of research related to our work.

Some of the earliest research on images and text~\citep{Barnard01,blei2003modeling,blei03,Duygulu02,Lavrenko03} has focused on learning the co-occurrences between image regions and tags using a generative model. Since most datasets used for training such models lack image annotation at the region level, learning to associate tags with image regions is a very challenging problem, especially for contaminated Internet photo collections with very large tag vocabularies. Moreover, image tags frequently refer to global properties or characteristics that cannot be easily localized. Therefore, we focus on establishing relationships between whole images and words.

Conceptually, our three-view formulation may be compared to the generative model that attempts to capture the relationships between the image class, annotation tags, and image features. One example of such a model in the literature is \cite{Wang09}. Unlike~\cite{Wang09}, though, we do not concern ourselves with the exact generative nature of the dependencies between the three views, but simply assign symmetric roles to them and model the pairwise correlations between them. Also, while~\cite{Wang09} tie annotation tags to image regions following~\cite{blei2003modeling,blei03}, we treat both the image appearance and all the tags assigned to the image as global feature vectors. This allows for much more scalable learning and inference (the approach of~\cite{Wang09} is only tested on datasets of under 2,000 images and eight classes each).

The major goal of our work is learning a joint latent space for images and tags, in which corresponding images and tags are mapped to nearby locations,
so that simple nearest-neighbor methods can be used to perform cross-modal tasks, including image-to-image, tag-to-image, and image-to-tag search.
A number of successful recent approaches to learning such an embedding rely on Canonical Correlation Analysis (CCA)~\citep{Hotelling36}. \cite{cca_survey} and \cite{Rasiwasia10} have applied CCA to map images and text to the same space for cross-modal retrieval tasks. \cite{Hwang10,hwang2011ijcv} have presented a cross-modal retrieval approach that models the relative importance of words based on the order in which they appear in user-provided annotations. \cite{Blaschko08} have used KCCA to develop a cross-view spectral clustering approach that can be applied to images and associated text. CCA embeddings have also been used in other domains, such as cross-language retrieval~\citep{udupa10,Vinokourov02}.
Unlike all the other CCA-based image retrieval and annotation approaches, ours adds a third view that explicitly represents the latent image semantics.

Our approach also has connections to supervised {\em multi-view learning}, in which images are characterized by visual and textual views, both of which are linked to the underlying semantic labels. The literature contains a number of sophisticated methods for multi-view learning, including generalizations of CCA/KCCA~\citep{Rai09,Sharma12,Yakhnenko09}, metric learning~\citep{Quadrianto11} and large-margin formulations~\citep{chenPAMI12}. Fortunately, we have found that our basic CCA formulation already gives very promising results without having to pay the price of increased complexity for learning and inference.

Since learning a projection for the data is equivalent to learning a Mahalanobis metric in the original feature space, our work is related to {\em metric learning}~\citep{Globerson05,Goldberg05,Weinberger05}. For example, the large-margin nearest neighbor (LMNN) approach \citep{Weinberger05} learns a distance metric that is optimized for nearest neighbor classification, and neighborhood component analysis (NCA)~\citep{Goldberg05} optimizes leave-one-out loss for nearest neighbor classification. Metric learning has been used for image classification and annotation~\citep{Guillaumin09,Mensink12,Verma12}. However, all of these approaches learn an embedding or a metric for visual features only, so they cannot be used to perform cross-modal retrieval.

The two main tasks we use for evaluating our system are image-to-image search,
which has been traditionally studied as {\em content-based image retrieval}~\citep{Datta08,Smeulders00}, and
tag-to-image search, or image retrieval using text-based queries~\citep{Grangier08,Krapac10,liuMM09,Lucchi12}.
A task related to tag-to-image search, though one we do not consider directly, is re-ranking of contaminated image search results for the purpose of dataset collection~\citep{berg06,Fan10,Frankel97,Schroff07}.

The third task we are interested in evaluating is image-to-tag search or automatic image annotation~\citep{Carneiro2007,li2008,Monay04}.
This task has traditionally been addressed with the help of sophisticated generative models such as~\cite{blei2003modeling,Carneiro2007,Lavrenko03}.
More recently, a number of publications have reported better results with simple data-driven schemes based on retrieving database images similar to a query and transferring the annotations from those images~\citep{nus-wide-civr09,Guillaumin09,Makadia08,Verma12,Wang08}. We will adopt this strategy in our experiments and demonstrate that retrieving similar images in our embedded latent space can improve the accuracy of tag transfer.

The data-driven image annotation approaches of~\cite{Guillaumin09,Makadia08,Verma12} use discriminative learning to obtain a metric or a weighting of different features to improve the relevance of database images retrieved for a query. Unfortunately, the learning stage is very computationally expensive -- for example, in the TagProp method of~\cite{Guillaumin09}, it scales quadratically with the number of images. In fact, the standard datasets used for image annotation by~\cite{Makadia08,Guillaumin09,Verma12} consist of 5K-20K images and have 260-290 tags each.
By contrast, our datasets (shown in Figure \ref{data}) range in size from 71K to 270K and have tag vocabularies of size 1K-20K. While it is possible to develop scalable metric learning algorithms using stochastic gradient descent (e.g.,~\cite{Mensink12}), our work shows that learning a linear embedding using CCA can serve as a simpler attractive alternative.

Perhaps the largest-scale image annotation system in the literature is the {\em Wsabie} (Web Scale Annotation by Image Embedding) system by \cite{Weston11}. It uses stochastic gradient descent to optimize a ranking objective function and is evaluated on datasets with ten million training examples. Like our approach, Wsabie learns a common embedding for visual and tag features. Unlike ours, however, it has only a two-view model and thus does not explicitly represent the distinction between the tags used to describe the image and the underlying image content. Also, Wsabie is not explicitly designed for multi-label annotation, and evaluated on datasets whose images come with single labels (or single paths in a label hierarchy). 

One of the shortcomings of data-driven annotation approaches~\citep{Guillaumin09,Makadia08,Verma12} as well as Wsabie is that they not account for co-occurrence and mutual exclusion constraints between different tags for the same image. If the retrieved nearest neighbors of an image belong to incompatible semantic categories (e.g., ``bird'' and ``plane''), then the tags transferred from them to the query may be incoherent as well (see Figure \ref{tagging} (a) for an example). To better exploit constraints between multiple tags, it is possible to treat image annotation as a multi-label classification problem~\citep{Chen11ICCV,Zhu05multi}. In the present work, we limit ourselves to learning the joint visual-textual embedding. It would be interesting to impose multi-label prediction constraints in the joint latent space -- in fact, \cite{Zhang11} have recently proposed an approach combining CCA with multi-label decoding -- but doing so is outside the scope of our paper.

Finally, our work has connections to approaches that use Internet images and accompanying text as auxiliary training data to improve performance on tasks such as image classification, for which cleanly labeled training data may be scarce~\citep{Guillaumin10,Quattoni07,WangForsyth09b}. In particular, \cite{Quattoni07} use the multi-task learning framework of \cite{Ando05} to learn a discriminative latent space from Web images and associated captions. We will use this embedding method as one of our baselines, though, unlike our approach, it can only be applied to images, not to tag vectors. Apart from multi-task learning, another popular way to obtain an intermediate embedding space for images is by mapping them to outputs of a bank of concept or attribute classifiers~\citep{Rasiwasia07,WangForsyth09a}. Once again, unlike our method, this produces an embedding for images only; also, training of a large number of concept classifiers tends to require more supervision and be more computationally intensive than training of a CCA model.



\section{Modeling Images, Tags, and High-Level Semantics\label{cca}}

\subsection{Scalable three-view CCA formulation} \label{sec:three_view_cca}

In this section, we introduce a three-view kernel CCA formulation for learning a joint space for visual, textual, and semantic information. Then we show how to obtain a scalable approximation using explicit kernel embeddings and linear CCA.

We assume we have $n$ training images each of which is associated with a $v$-dimensional visual feature vector and a $t$-dimensional tag feature vector (our specific feature representations for both views will be discussed in Section \ref{approximation}). The respective vectors are stacked as rows in matrices $V \in \mathbb{R}^{n\times v}$ and $T \in \mathbb{R}^{n\times t}$. In addition, each training image is also associated with semantic class or topic information, which is encoded in a matrix $C \in \mathbb{R}^{n\times c}$, where $c$ is the number of classes or topics. Each image may be labeled with exactly one of the $c$ classes (in which case only one entry in each row of $C$ is 1 and the rest are 0); alternatively, each image may be described by several of the $c$ keywords (in which case, multiple entries in each row of $C$ may be 1). Another possibility is that $C$ is a soft indication matrix, where the $i,j$th entry indicates the degree (or posterior probability) with which image $i$ belongs to the $j$th class or topic. In the supervised learning scenario,  $C$  is obtained from (possibly noisy) annotations that come with the training data. In the unsupervised scenario (where only images and tags are initially given), $C$ is ``latent'' and must be obtained by clustering the tags, as will be discussed in Section \ref{topics}. To simplify the notation in the following, we will also use $X_1,X_2,X_3$ to denote $V, T, C$ respectively.

Let $\boldsymbol x, \boldsymbol y$ denote two points from the $i$th view. The similarity between these points is defined by a kernel function $K_i$ such that $K_i(\boldsymbol x,\boldsymbol y) = \varphi_i(\boldsymbol x)\varphi_i(\boldsymbol y)^\top$, where $\varphi_i(\cdot)$ is a function embedding the original feature vector into a nonlinear kernel space. Practical kernel-based learning schemes do not work in the embedded space directly, relying on the kernel function instead. However, we will formulate KCCA as solving for a linear projection from the kernel space, because this leads directly to our scalable approximation scheme based on explicit embeddings.

\begin{figure}[] 
   \centering
   \subfigure[Two-view model.]{
   \includegraphics[width=1.3in, trim=30mm 165mm 70mm
10mm]{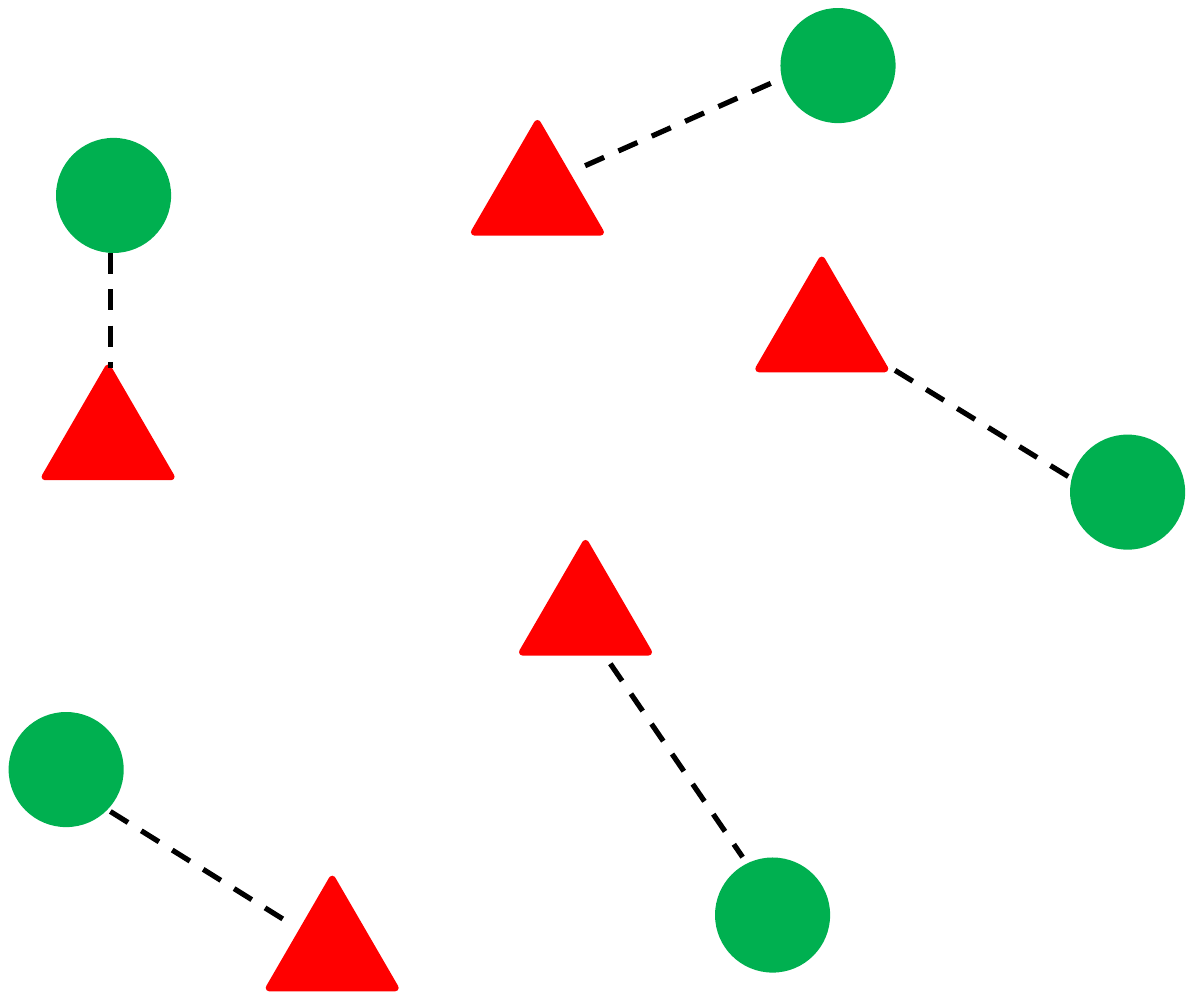}}
   \subfigure[Three-view model.]{
   \includegraphics[width=1.3in, trim=15mm 168mm 90mm
10mm]{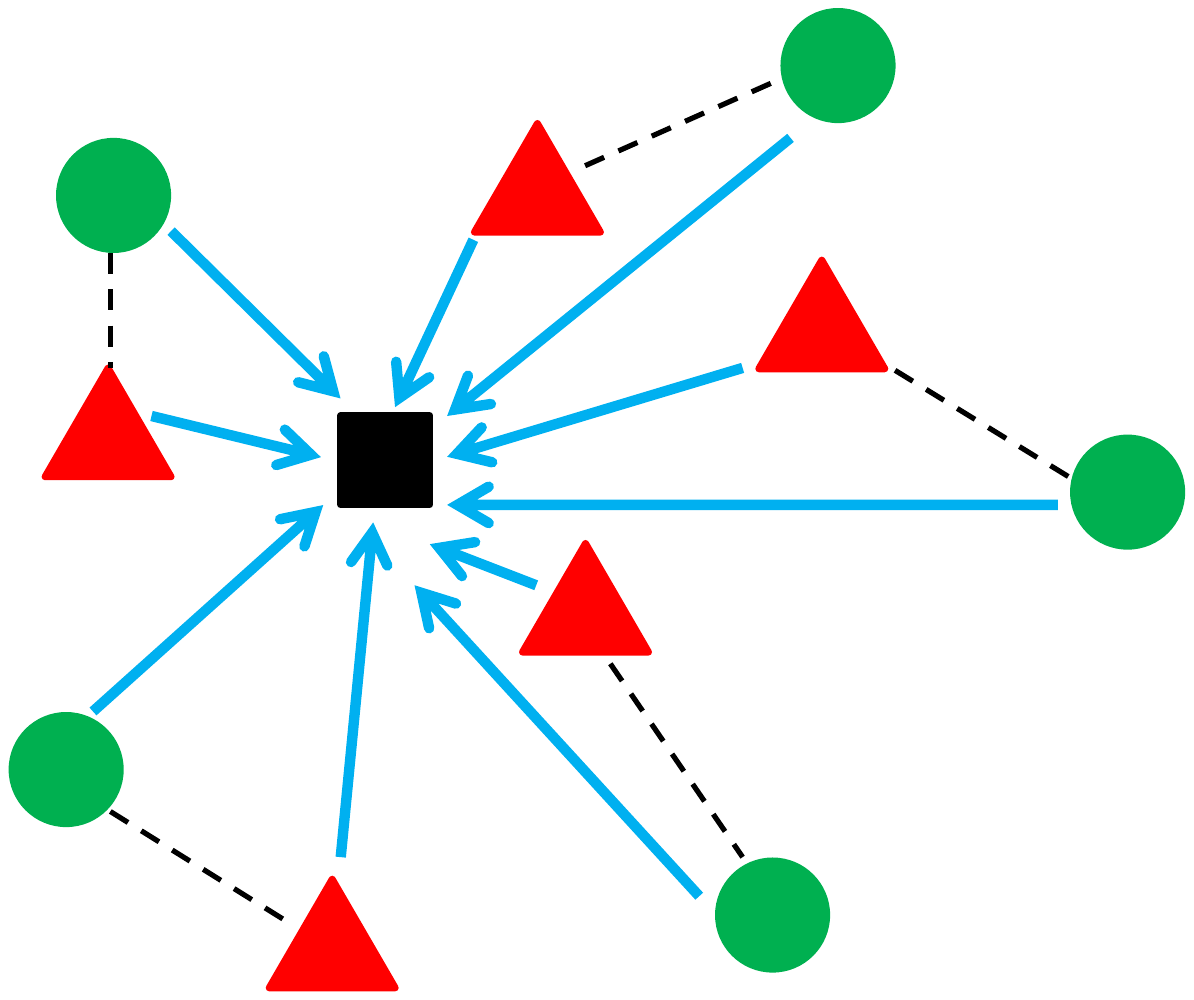}}
   \caption{(a) Traditional two-view CCA minimizes the distance (equivalently, maximizes the correlation) between images (triangles) and their corresponding tags (circles). (b) Our proposed approach is to incorporate semantic classes or topics (black squares) as a third view. Images and tags belonging to the same semantic cluster are forced to be close to each other, imposing additional high-level structure. See also Figure \ref{2d} for a visualization of two embeddings on real data.}
   \label{toy}   \vspace{-10pt}
\end{figure}

In KCCA, we want to find matrices $W_i$ that project the embedded vectors $\varphi_i(\boldsymbol x)$ from each view into a low-dimen-sional common space
such that the distances in the resulting space between each pair of views for the same data item are minimized.
The objective function for this formulation is given by
 \begin{gather}
\min_{\substack{W_1, W_2, W_3}} \sum_{i,j=1}^3 \| \varphi_i(X_i)W_i - \varphi_j(X_j)W_j\|_F^2 \label{eq:cca} \\
\mathrm{subject~to~~~} W_i^\top \Sigma_{ii} W_i = {I}, ~~\boldsymbol{w}_{ik}^\top \Sigma_{ij} \boldsymbol{w}_{jl} = 0, \nonumber \\
i,j=1,\ldots,3,~~i\neq j\,,~~ k,l = 1, \ldots, d,~~ k \neq l, \nonumber
\end{gather}
where $\Sigma_{ij}$ is the covariance matrix between $\varphi(X_i)$ and $\varphi(X_j)$, and $\boldsymbol{w}_{ik}$ is the $k$th column of $W_i$ (the number of columns in each $W_i$ is equal to the dimensionality of the resulting common space). To better understand this objective function, let us consider its three terms:
 \begin{gather}
\min_{W_1,W_2,W_3} \|\varphi_1(V) W_1-\varphi_2(T) W_2\|^2_F + \nonumber \\
 \|\varphi_1(V) W_1-\varphi_3(C) W_3\|^2_F +\|\varphi_2(T) W_2-\varphi_3(C) W_3\|^2_F. \nonumber
\end{gather}
The first term tries to align corresponding images and tags, and it is the sole term in the standard two-view CCA objective~\citep{cca_survey}. The remaining two terms, which are introduced in our three-view model, try to align images (resp. tags) with their semantic topic. Figure \ref{toy} illustrates the difference between the two- and three-view formulations graphically.

In the standard KCCA formulation, instead of directly solving for linear projections of data explicitly mapped into the kernel space by $\varphi_i$, one applies the ``kernel trick'' and expresses the coordinates of a data point in the CCA space as linear combinations of kernel values of that point and several training points.
To find the weights in this combination, one must solve a $3n \times 3n$ generalized eigenvalue problem (see \cite{Bach02,cca_survey} for details), which is infeasible for large-scale data.

To handle large numbers of images and high-dimensional features, we propose a scalable approach based on the idea of approximate kernel maps~\citep{maji09,Perronnin10,Rahimi07,Vedaldi10}. Let $\hat{\varphi}(\boldsymbol x)$ denote an approximate kernel mapping such that $K_i(\boldsymbol x,\boldsymbol x') \simeq \hat{\varphi}_i(\boldsymbol x)\hat{\varphi}_i(\boldsymbol x')^\top$. The dimensionality of $\hat{\varphi}(\boldsymbol x)$ needs to be much lower than $n$ to reduce the complexity of the problem.  The specific kernel mappings used in our implementation will be described in Section \ref{sec:visual_kernel}. Then, instead of using the kernel trick, we can directly substitute $\hat{\varphi}(\boldsymbol x)$ into the linear CCA objective function (\ref{eq:cca}). The solution is given by the following generalized eigenvalue problem:
  \begin{eqnarray*}
\begin{pmatrix}
S_{11} & S_{12} & S_{13}\\
S_{21} & S_{22} & S_{23}\\
S_{31} & S_{32} & S_{33}\\
\end{pmatrix}
\begin{pmatrix}
\boldsymbol w_1 \\
\boldsymbol w_2 \\
\boldsymbol w_3 \\
\end{pmatrix}
 =\lambda
\begin{pmatrix}
S_{11} & 0 & 0\\
0 & S_{22} &  0\\
0 &0& S_{33}\\
\end{pmatrix}
\begin{pmatrix}
\boldsymbol w_1 \\
\boldsymbol w_2 \\
\boldsymbol w_3 \\
\end{pmatrix},
 \end{eqnarray*}
where $S_{ij} = \hat{\varphi}_i(X_i)^\top \hat{\varphi}_j(X_j)$ is the covariance matrix between the $i$th and $j$th views, and $\boldsymbol w_i$ is a column of $W_i$.
The size of this problem is $(d_1+d_2+d_3)\times (d_1+d_2+d_3)$, where the $d_i$ are the dimensionalities of the respective explicit mappings $\hat{\varphi}_i(\cdot)$. This is independent of training set size, and much smaller than $3n \times 3n$. To regularize the problem, we add a small constant ($10^{-4}$ in the experiments) to the diagonal of the covariance matrix.

In order to obtain a $d$-dimensional embedding for different views, we form projection matrices $W_i \in \mathbb{R}^{d_i \times d}$ from the top $d$ eigenvectors corresponding to each $\boldsymbol w_i$. Then the projection of a data point $\boldsymbol x$ from the $i$th view into the latent CCA space is given by $\hat{\varphi}_i(\boldsymbol x) W_i$. Note that once they are learned, the respective projection matrices are applied to each view individually, which means that at test time, we can compute the embedding for data for which one or two views are missing (e.g., an image without tags or ground-truth semantic labels). In the latent CCA space, points from different views are directly comparable, so we can do image-to-image, image-to-tag, and tag-to-image retrieval by nearest neighbor search.

In the implementation, we select the embedding dimensionality $d$ by measuring the retrieval accuracy in embedded spaces with different values of $d$ on validation images set aside from each of our datasets (details will be given in Sections 6-8). We search a range from 16 to 1,024, doubling the dimensionality each time, and the resulting values typically fall around 128-256 on all our datasets.

\subsection{Similarity function in the latent space\label{metric}}

In the CCA-projected latent space, the function used to measure the similarity between data points is important. An obvious choice is the Euclidean distance between embedded data points, as used in \cite{foster10,Hwang10,Rasiwasia10}. However, for our learned embedding, we were able to find a similarity function that produces better empirical results. In particular, we scale the dimensions in the common latent space by the magnitude of the corresponding eigenvalues~\citep{Chapelle03}, and then compute normalized correlation between projected vectors. Indeed, the CCA objective can be reformulated as maximizing the normalized correlation between different views \citep{cca_survey}.

Let $\boldsymbol x$ and $\boldsymbol y$ be points from the $i$th and $j$th views, respectively (we can have $i=j$). Then we define the similarity function between $\boldsymbol x$ and $\boldsymbol y$ as
\begin{gather}\label{distancefunction}
\operatorname{sim}(\boldsymbol x, \boldsymbol y) = \frac{(\hat{\varphi}_i(\boldsymbol x)W_iD_i) (\hat{\varphi}_j(\boldsymbol y)W_jD_j)^\top}{\| \hat{\varphi}_i(\boldsymbol x)W_iD_i \|_2 \| \hat{\varphi}_j(\boldsymbol y)W_jD_j \|_2} \,,
\end{gather}
where $W_i$ and $W_j$ are the CCA projections for data points $\boldsymbol x$ and $\boldsymbol y$,  and $D_i$ and $D_j$ are diagonal matrices whose diagonal elements are given by the $p$-th power of the corresponding eigenvalues~\citep{Chapelle03}. We fix $p=4$ in all our experiments as we have found this leads to the best performance. Section \ref{distancemetric} will experimentally confirm that the above similarity measure leads to much higher retrieval accuracy than Euclidean distance.

\section{Representations of the Three Views \label{approximation}}

In Sections \ref{sec:visual_kernel} and \ref{sec:tag_approximation}, we will present our visual and text features with their respective kernel mappings. Next, in Section \ref{topics}, we will discuss different text clustering approaches that can be used to extract semantic topics in the unsupervised scenario, where the third view is not given for the training data.

\subsection{Visual feature representation \label{sec:visual_kernel}}

We represent image appearance using a combination of nine different visual cues:
\smallskip

\noindent \textbf{GIST} \citep{gist}: We resize each image to $200\times 200$ and use three different scales $[8,8,4]$ to filter each RGB channel, resulting in 960-dimensional ($320 \times 3$) GIST feature vectors.
\smallskip

\noindent \textbf{SIFT}: We extract six different texture features based on two different patch sampling schemes: dense sampling and Harris corner detection. For each local patch, we extract \textbf{SIFT} \citep{lowe04}, \textbf{CSIFT} \citep{Sande10}, and \textbf{RGBSIFT} \citep{Sande10}. For each feature, we form a codebook of size 1,000 using k-means clustering and build a two-level spatial pyramid \citep{lazebnik06}, resulting in a 5000-dimensional vector. We will refer to these six features as D-SIFT, D-CSIFT, D-RGBSIFT, H-SIFT, H-CSIFT, and H-RGBSIFT.
\smallskip

\noindent \textbf{HOG} \citep{Dalal05}: To represent texture and edge information on a larger scale, we use $2\times2$ overlapping HOG as described in \cite{xiao10}. We quantize the HOG features to a codebook of size 1,000 and use the same spatial pyramid scheme as above, once again resulting in 5,000-dimensional feature vectors.\smallskip

\noindent \textbf{Color}: We use a joint RGB color histogram of 8 bins per dimension, for a 512-dimensional feature.\smallskip

Recall from Section \ref{sec:three_view_cca} that we transform all the features by nonlinear kernel maps $\hat{\varphi}(\boldsymbol x)$ and then apply linear CCA to the result. We discuss the specific feature maps we use here.
For GIST features, we use the random Fourier feature mapping~\citep{Rahimi07} that approximates the Gaussian kernel. We compute this mapping with 3,000 dimensions and set its standard deviation equal to the average distance to the 50th nearest neighbor in each dataset. All the other descriptors above are histograms, and for them we adopt the exact Bhattacharyya kernel mapping given by term-wise square root~\citep{Perronnin10}.
To combine different features, we simply average the respective kernels, which has been proven to be quite effective in \cite{Gehler09}.
This corresponds to concatenating all the different visual features after putting them through their respective explicit kernel mappings.
However, the resulting concatenated feature has 38,512 dimensions, necessitating additional dimensionality reduction. To do this, we perform PCA on top of each kernel-mapped feature $\hat{\varphi}_i(\cdot)$. This is essentially using the low-rank approximation of kernel PCA (KPCA)~\citep{kpca} to approximate the combined multiple feature kernel matrix.

In our experiments, we reduce each kernel-mapped feature to $500$ dimensions and the final concatenated feature is a $4,500$-dimensional vector. As validated in Section \ref{kccamodel}, this dimensionality achieves good balance between efficiency and accuracy. Note that for multiple feature combination, we have found it important to center all feature dimensions at the origin.

\subsection{Tag feature representation} \label{sec:tag_approximation}

For the tags associated with the images, we construct a dictionary consisting of $t$ most frequent tags (the vocabulary sizes used for the different datasets are summarized in Figure \ref{data} and will be further detailed in Section \ref{dataset}) and manually remove a small set of stop words. These include camera brands (e.g., ``canon,'' ``nikon,'' etc.), lens characteristics (e.g., ``eos,'' ``70-200mm,'' etc.), and words like ``geo.'' The tag feature matrix $T$ is binary: $T_{ij} = 1$ if image $i$ is tagged with tag $j$ and 0 otherwise. Note that even though the dimensionality of the tag feature may be high (our vocabularies range from 1,000 to over 20,000 on the different datasets), this representation is highly sparse.

Like~\cite{Guillaumin10}, we use the linear kernel for $T$, which corresponds to counting the number of common tags between two images. However, because of the high dimensionality of the tag features, additional compression is required, just as with the concatenated visual features. We apply sparse SVD~\citep{sparseSVD} to the tag feature $T$ to obtain a low-rank decomposition as $T=U_1SU_2^\top$. It is easy to show that $U_1S$ is actually the PCA embedding for $T$, but directly applying sparse SVD to $T$ is more efficient. In our implementation, the compressed representation of $T$ is given by the top 500 columns of $U_1S$.

We have also investigated more sophisticated tag features such as the ranking-based representation of~\cite{Hwang10}, which is based on the idea that tags listed earlier by users are more salient to the image content. However, we have found almost no improvement from our chosen representation.

\subsection{Semantic view representation \label{topics}}

As initially discussed in Section \ref{cca}, the third view of our CCA model is given by the class or topic indicator matrix $C \in \mathbb{R}^{n \times c}$ for $n$ images and $c$ topics. In the supervised training scenario, $C$ is straightforwardly given by ground-truth annotations or noisy search keywords used to download the data. In the more interesting unsupervised scenario, training images come with noisy text or tags, but no additional semantic annotations. In this case, we choose to obtain $C$ by clustering the tags. Given the raw tag feature $T$ ({\em prior} to the application of sparse SVD), our goal is to find $c$ semantic clusters. For this purpose, we investigate several models that have proven successful for text clustering. We briefly describe these models below; quantitative and qualitative evaluation results will be presented in Section \ref{compareTopicModel}.
\smallskip

\noindent \textbf{K-means clustering}: The simplest baseline approach is k-means clustering on raw tag feature $T$ using $c$ centers. The resulting matrix $C$ has a 1 in the $i,j$th position if the $i$th tag feature vector belongs to the $j$th cluster.
\smallskip

\noindent \textbf{Normalized cut (NC)}~\citep{ng01,shi00}:  For text clustering, the normalized cut model is usually formulated as computing the eigenvectors of
\[L = I-D^{-1/2}TT^\top D^{-1/2}\,,\]
in which $D=\mathrm{diag}(T(T^\top\mathbf{1}))$. This is equivalent to computing the first $c$ singular vectors of the sparse matrix $D^{-1/2}T$. Following \cite{ng01}, we normalize each row of the matrix of top $c$ eigenvectors to have unit norm and perform k-means clustering of rows of the resulting matrix $\widetilde{U}$. Directly using $\widetilde{U}$ as $C$ would represent a ``soft'' version of NC, but we have found that the ``hard'' version obtained by k-means produces better results.
\smallskip

\noindent \textbf{Nonnegative matrix factorization (NMF)}~\citep{NMF03}:  The data matrix is normalized as $D^{-1/2}T$, where $D$ is defined the same way as for NC, and then factorized into two nonnegative matrices $U$ and $V$ such that $T = U^\top V$ (if $T$ is $n \times t$, then $U$ is $c \times n$ and $V$ is $c \times t$). Then, as in~\cite{NMF03}, we obtain a normalized matrix $\widetilde{U}$ with entries $U_{ij} / \sqrt{\sum_j V_{ij}^2}$. Finally, we do hard cluster assignment based on the highest value of $\widetilde{U}$ in each row. Just as with NC, this produces better results than using $\widetilde{U}$ as a ``soft'' cluster indicator matrix directly.
\smallskip

\noindent \textbf{Probabilistic latent semantic analysis (pLSA)}~\citep{Hofmann99}: This approach models each document (vector of tags for an image) as a mixture of topics. The output of pLSA is the posterior probability of each topic given each document. Directly using this matrix of posterior probabilities as $C$ leads to ``soft'' pLSA clustering. However, once again, we get better performance with ``hard'' pLSA where we map each document to the cluster index with the highest posterior probability. We have also investigated latent Dirichlet allocation (LDA)~\citep{blei03} and found the performance to be similar to pLSA, so we omit it.
\smallskip

Because the number of topics used in this work is not very high (from 10 to 100), we simply use a linear kernel on $C$ with no further dimensionality reduction.

\section{Overview of Experimental Evaluation \label{dataset}}

This section will present the components of our experimental evaluation, including datasets, retrieval tasks, multi-view models being compared, and baselines. Subsequently, Sections \ref{sec:evaluation}-\ref{sec:evaluation3} will present results on our three datasets.

\subsection{Datasets}

Our selection of datasets is motivated by two considerations. First, we want datasets that are as large as possible, both in the number of images and in the number of tags. Second, we want datasets that have the right kind of annotations for evaluating our method -- specifically, images that are accompanied both by noisy text or tags, and ground-truth labels.

We have considered a number of datasets used in recent related papers, but unfortunately, most of them are unsuitable for our goals.
In particular, standard image annotation datasets used by \cite{Makadia08,Guillaumin09,Rasiwasia07,Verma12} -- namely, Corel5K \citep{Duygulu02}, ESP Game \citep{VonAhn04}, and IAPR-TC \citep{Grubinger06} -- have only two views and are rather small-scale (5K-20K images and 260-290 tags).
\cite{Rasiwasia10}, who have first proposed a two-view CCA model for cross-modal retrieval of Internet images, perform experiments on a Wikipedia dataset that has rich textual views as well as ground-truth labels, but it consists of only 2,866 documents. \cite{Weston11} evaluate their Wsabie annotation system on millions of images. However, one of their datasets is drawn from ImageNet~\citep{deng09}, which is more appropriate for image classification, and the other one is proprietary; neither has the three-view structure we are looking for.

The three datasets we have chosen are shown in Figure \ref{data}. The first one is collected by ourselves, while the other two are publicly available. \smallskip

\noindent{\textbf{Flickr-CIFAR dataset}}: We have downloaded 230,173 images from Flickr by running queries for categories from the CIFAR10 dataset~\citep{cifar}: airplane, automobile, bird, cat, deer, dog, frog, horse, ship, and truck. We keep tags that appear at least 150 times, resulting in a tag dictionary with dimensionality 2,494. On average, there are 6.84 tags per image.
The Flickr images come with search keywords and user-provided tags, but no ground-truth labels. To quantitatively evaluate retrieval performance we need another set of cleanly labeled test images. We get this set by collecting the same ten categories from ImageNet~\citep{deng09}, resulting in 15,167 test images with no tags but ground-truth class labels.
\smallskip

\noindent{\textbf{NUS-WIDE dataset}}: This dataset~\citep{nus-wide-civr09} was collected at the National University of Singapore. It also originates from Flickr, and contains 269,648 images. The dataset is manually annotated with 81 ground truth concept labels, e.g., animal, snow, dog, reflection, city, storm, fog, etc. One important difference between NUS-WIDE and other datasets is that NUS-wide images may be associated with multiple ground truth labels.  For the tags, we use the list of 1,000 words provided by~\cite{nus-wide-civr09}; on average, each image has 5.78 tags and 1.86 ground truth annotations. Each ground truth concept is also in the tag dictionary. \smallskip

\noindent{\textbf{INRIA-Websearch dataset}}: Finally, we use the INRIA Web query dataset \citep{Krapac10}, which contains 71,478 web images and 353 different concepts or categories, which include famous landmarks, actors, films, logos, etc. Each concept comes with a number of images retrieved via Internet search, and each image is marked as either relevant or irrelevant to its query concept. This dataset is especially challenging in that it contains a very large number of concepts relative to the total number of images. The second view for this dataset consists of text surrounding images on web pages, not tags. We keep words that appear more than 20 times and remove stop words using a standard list for text document analysis, which gives us a tag dictionary of size 20,602. On this dataset, we also apply \emph{tf-idf} weighting to the tag feature. \smallskip

The above three datasets have different characteristics and present different challenges for our method. Flickr-CIFAR has the fewest classes but the largest number of images per class. It is also the only dataset whose training images have no ground-truth semantic annotation, and whose test images come from a different distribution than the training images. We use this dataset for detailed comparative validation of the different implementation choices of our method (Section \ref{sec:evaluation}). NUS-WIDE images are fully manually annotated and come with multiple ground truth concepts per image. INRIA-Websearch is the only one not collected from Flickr, and its images are the most inconsistent in quality. It has the largest number of classes but the smallest number of images per class. It also has by far the largest vocabulary for the second view and the noisiest statistics for this view.

\subsection{Retrieval tasks}

For evaluation, we consider the following tasks. \smallskip

\noindent \textbf{Image-to-image search (I2I)}: Given a query image, project its visual feature vector into the CCA space, and use it to retrieve the most similar visual features from the database. Recall that our similarity function in the CCA space is given by eq. (\ref{distancefunction}). \smallskip

\noindent \textbf{Tag-to-image search (T2I)}: Given a search tag or combination of tags, project the corresponding feature vector into the CCA space and retrieve the most similar database images. This is a cross-modal task, in that the CCA-embedded tag query is used to directly search CCA-embedded visual features. Note that with our method, we can use tags to search database images that do not initially come with any tags.
In scenarios where ground-truth labels or keywords are available for the database images, we also consider a variant of this task where we use the keywords as queries, which we refer to as \textbf{keyword-to-image search (K2I)}.
\smallskip

\noindent \textbf{Image-to-tag search (I2T):} Given an image, retrieve a set of tags that accurately describe it. This task is more challenging than the other two because going from a feature vector in CCA space to a coherent set of tags requires a sophisticated reconstruction or decoding algorithm (see, e.g., \cite{Hsu09,Zhang11}). The design of such an algorithm is beyond the scope of our present work, but to get a preliminary idea of the promise of our latent space representation for this task, we evaluate a simple data-driven scheme similar to that of \cite{Makadia08}. Namely, given a query image, we first find the fifty nearest neighbor tag vectors in CCA space, and then return the five tags with the highest frequencies in the corresponding database images. Note that \cite{Makadia08} return tags according to their global frequencies, while for our larger and more diverse datasets, we have found that local frequency works better. Because our method for I2T is somewhat preliminary and because proper evaluation of this task requires human annotators (see Section \ref{sec:tagging}), our experiments on this task will be smaller-scale and more limited than on the other two.
\smallskip

The precise evaluation protocols and performance metrics used for each task are dataset-specific, and will be described in Sections \ref{sec:evaluation}-\ref{sec:evaluation3}.

\subsection{Multi-view models}


In the subsequent presentation, we will denote visual features as {\bf V}, tag features as {\bf T}, the keyword or ground truth annotations as {\bf K}, and the automatically computed topics as {\bf C}. {\bf CCA (V+T)} will refer to the two-view baseline model based on visual and tag features; {\bf CCA (V+T+K)} to the three-view model based on visual features, tags, and supervised semantic information (ground truth labels or search keywords); and {\bf CCA (V+T+C)} to the three-view model with the unsupervised third view (automatically computed tag clusters). All of these models will be evaluated for both I2I and T2I retrieval. For completeness, we will also evaluate the two-view {\bf CCA (V+C)} and {\bf CCA (V+K)}\footnote{It can be shown that CCA with labels as one of the views is equivalent to Linear Discriminant Analysis (LDA)~\citep{Bartlett38}.} models for I2I retrieval. However, because these models do not give an embedding for the tags, they cannot be used for cross-modal retrieval (i.e., T2I). In addition, we will evaluate K2I and I2T on subsets of models as appropriate (see Sections \ref{sec:evaluation}-\ref{sec:evaluation3} for details).

\subsection{Baselines}


It is important to evaluate how CCA compares to alternative methods for obtaining intermediate embeddings for visual features supervised by tag or keyword information. For this, we have implemented two embedding methods from the recent literature, as described below.
\smallskip

\noindent{\bf Structural learning.} Our first baseline is given by the structural or multi-task learning method of~\cite{Ando05,Quattoni07}. In their formulation, the tag matrix $T$ is treated as supervisory information for the visual features $V$ and a matrix of image-to-tag predictors $W$ is obtained by ridge regression: $\|T-VW\|^2 + \rho \|W\|^2$. Next, since the tasks of predicting multiple tags are assumed to be correlated, we look for low-rank structure in $W$ by computing its SVD. If $W = U_1SU_2^\top$, then we use $U_1$ (or more precisely, its top $d$ columns) as the embedding matrix for the visual features: $E = VU_1$. We select $d$ by validation just as with our CCA-based methods. Note that the structural learning method does not produce an embedding for tags, so unlike CCA (V+T) and our three-view models, it is not suitable for cross-modal retrieval.
\smallskip

\begin{table*}[]
{\footnotesize
\hfill{}
\begin{tabular}{c||cccccccccc}
\hline
 &  D-SIFT &    D-CSIFT  &  D-RGBSIFT  &  H-SIFT &    H-CSIFT  &   H-RGBSIFT  & HOG    &    GIST  &  RGB Hist.  &  All    \\
\hline
& \multicolumn{10}{c}{Image-to-image search} \\
No PCA  &      42.51   &     42.58   &     41.54     &  43.57      &  41.27    &    43.02     &   43.13      &  40.48  	&	24.32    &         --  \\
PCA (500D)  &    42.03   &         42.01   &         40.86      &      42.51     &       40.98    &        42.21       &     42.41     &       39.96  	  &  24.38	   & \textbf{54.90}    \\
\hline
& \multicolumn{10}{c}{Tag-to-image search} \\
No PCA  &           53.68   &    50.34 &      52.02   &    55.67     &  52.18 &      54.84   &    54.06  &     48.75   &   26.88	    &       --   \\
PCA (500D)  &          55.24 &       53.71 &       53.83  &      58.42  &      54.18  &      57.29  &      56.12    &    51.41  &   28.52&  \textbf{64.07}     \\
\hline
\end{tabular}}
\hfill{}
\caption{Precision@50 for full-dimensional vs. PCA-reduced data for \textbf{image-to-image} (top) and \textbf{tag-to-image} (bottom) retrieval for CCA (V+T). We did not obtain the result for combined features without PCA due to excessive computational cost (see text).}
\label{approximation2}
\end{table*}

\noindent{\bf Wsabie.} As a second baseline, we use Wsabie~\citep{Weston11}. This method learns a discriminative model of the form
$f(\boldsymbol x) = \boldsymbol xUW$ where $U$ is the embedding and $W$ is the matrix of weights for a set of classifiers, which in our case correspond to keywords (the K view). Once we obtain the embedding matrix $U$ for the visual features, we evaluate the accuracy of I2I in the embedded space. Note that Wsabie, just as the structural learning method described above, cannot be used for cross-modal retrieval.

We have implemented Wsabie as described in~\cite{Weston11}, using stochastic gradient descent (SGD) to optimize a ranking-based loss function.
We use random initialization and a fixed learning rate of 0.01. The only difference from~\cite{Weston11} is that instead of explicit regularization, we use early stopping as suggested in \cite{Gordo2012}. Namely, we run the SGD training for $20n$ iterations, where $n$ is the number of training points, and validate the performance on the I2I task after processing every 1,000 points. At the end, the parameters with the highest validation accuracy are picked. For the dimensionality of the embedding $U$, we use a value of 128, which is relatively efficient and achieves good performance (it is also comparable to the typical dimensionalities that get selected by validation for our CCA model). Overall, the training of Wsabie is slower and more complicated than our approach, as it involves learning rate tuning and validation for early stopping. By contrast, the only parameter in our approach is the regularization constant in the covariance matrix (see Section \ref{sec:three_view_cca}) and it is set to a fixed value in all our experiments.

\section{In-depth Analysis on Flickr-CIFAR} \label{sec:evaluation}

In this section, we use the Flickr-CIFAR dataset to conduct a detailed study of the various components of our proposed multi-view models, including feature combination and compression, different methods for tag clustering, and the proposed similarity function. For this purpose, we use the I2I, T2I, and K2I tasks. At the end, in Section \ref{sec:tagging} we perform a smaller-scale evaluation of I2T or image tagging.

\subsection{Experimental protocol} \label{sec:protocol}

Remember from Section \ref{dataset} that the Flickr-CIFAR dataset consists of 230,173 Flickr images that are used to learn the CCA embedding and 15,167 ImageNet images that are used for quantitative evaluation. The ImageNet images are split into 13,167 ``database'' images against which retrieval is performed, 1,000 validation images, and 1,000 test images. One fixed split is used for all experiments. The validation images are run as queries against the database in order to select the dimensionality $d$ of the embedding.

For I2I search, we use the test images as queries and report precision at top $p$ retrieved images (Precision@$p$) -- that is, the fraction of the $p$ returned images having the same ImageNet label as the query. For T2I search, we need a different set of queries as the ImageNet images are not tagged. For this, we take the tag feature vectors of 1,000 randomly chosen Flickr images (which are excluded from the set used to learn the embedding). The ground truth label of each query is given by the search keyword used to download the corresponding image. We have manually examined 50 of these queries and
found that about 75\% percent of the tags are closely related to the ground-truth keyword --
for example, for the ``car'' keyword, the tags contain ``car,'' ``cars,'' and most of the other tags are also loosely related
to the ground-truth keyword (``auto''). Thus, we expect this evaluation scheme to be reasonably accurate.
Just as for image-to-image search, the evaluation metric for tag-to-image search is Precision@$p$.

\begin{table*}[]
{\hfill{}
\begin{tabular}{c||cccccc||cccccc}
\hline
& \multicolumn{6}{c||}{Image-to-image search} & \multicolumn{6}{c}{Tag-to-image search} \\
\# clusters &   10  &  20  & 30  &  40   &  50    & 100 & 10 & 20 & 30 & 40 & 50 & 100 \\
\hline
visual k-means &   49.12  &    48.73  &     48.73    &   48.52  &     48.50    &   47.58 &
  57.35    &  56.67    &  56.47    &  56.07  &    56.16    &  56.20    	          \\
\hline
k-means &      51.60     &  56.18  &        57.36    &   57.45     &  57.34     &   \textbf{57.40} &
63.23   &   65.23  &    67.36    &  66.76     & 68.29     & \textbf{70.29}      \\
NC &           	 54.86    &   \textbf{62.33}    &    \textbf{61.90}  &      \textbf{61.21}   &    \textbf{60.65}    &   56.65	&
63.75  &    \textbf{76.05}    &  \textbf{74.90}  &    \textbf{72.58}   &  \textbf{72.45}     & 67.71      \\
NMF &           54.01   &    55.45     &  57.51   &   56.63   &    55.98  &     53.07   &
\textbf{64.44}     & 66.58    &  67.03     & 67.41   &   66.33   &  62.54      \\
pLSA &    	      \textbf{55.40}    &   56.43   &   57.33    &   57.62   &    56.89    &   54.88   &
62.45    &  64.71    &  64.92   &   65.76  &   67.32     & 66.83     \\
\hline
\end{tabular}
}\hfill{}
\caption{Precision@50 for different clustering methods for \textbf{image-to-image} and \textbf{tag-to-image} retrieval with the CCA (V+T+C) model. The second row shows results for visual clusters and the remaining rows for semantic (tag-based) clusters. The results are averaged over five different random initializations of the clustering methods. The performance of the CCA (V+T) baseline is 54.90\% for image-to-image search, and 64.02\% for tag-to-image search.}
\label{topic1}
\end{table*}

\subsection{Evaluation of features and training set size \label{kccamodel}}

Recall from Section \ref{sec:visual_kernel} that we apply nonlinear kernel maps to nine different visual features, reduce each of them to 500 PCA dimensions, and concatenate them together. As for the tag features (Section \ref{sec:tag_approximation}), we use sparse SVD to compress them to 500 dimensions. In this section, we evaluate these transformations. Since no dimensionality reduction is involved in the third view (K or C), for simplicity, we perform the evaluation with the standard two-view CCA (V+T) model.

Table \ref{approximation2} reports the effect of PCA on image-to-image and tag-to-image search for individual and combined visual features. For image-to-image retrieval, applying PCA to the original feature vectors may slightly hurt performance, but the decrease is less than 1\%. More importantly, combined features significantly outperform each individual feature. On the other hand, for tag-to-image retrieval, PCA consistently helps to improve performance. This is possibly because Flickr tags are noisy, and reducing the dimensionality smooths the data.

To motivate our use of dimensionality reduction, it is instructive to give some running times on our platform, a 4-core Xeon 3.33GHz workstation with 48GB RAM. For a single feature, it takes 2.5 seconds to obtain the CCA solution for approximated data (500 V + 500 T dimensions), versus 20 minutes for the full-dimensional kernel-mapped data (5,000 V + 2,494 T). For all nine combined visual features, it took around five minutes to get the approximated solution (4,500 V + 500 T); because the computation scales cubically with the number of dimensions, we have not tried to obtain the full-dimensional solution for all features (38,512 V + 2,494 T). Likewise, while one would ideally want to compare our results to an exact KCCA solution using kernel matrices, for hundreds of thousands of training points this is completely infeasible on our platform (for $n$ training points, exact two-view KCCA involves solving a $2n \times 2n$ eigenvalue problem).

We conclude that combining multiple visual cues is indeed necessary to get the highest absolute levels of accuracy, and that dimensionality reduction of kernel-mapped features can satisfactorily address memory and computational issues with negligible loss of accuracy.

\begin{figure}[]
   \centering
   \includegraphics[width=2.2in, trim= 25mm 50mm 25mm 65mm]{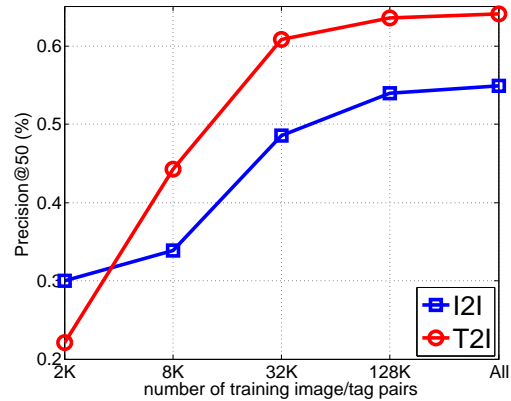}
   \caption{Performance with increasing number of training images on the Flickr-CIFAR dataset.}
   \label{fig:size}
\end{figure}

Finally, Figure \ref{fig:size} reports the retrieval accuracy of the two-view CCA (V+T) as a function of training dataset size. We can see that a fairly large amount of data, above 100K, is needed to converge to good performance.

\begin{figure*}[t]
   \centering
   \includegraphics[width=1.6in, trim= 25mm 70mm 25mm 0mm]{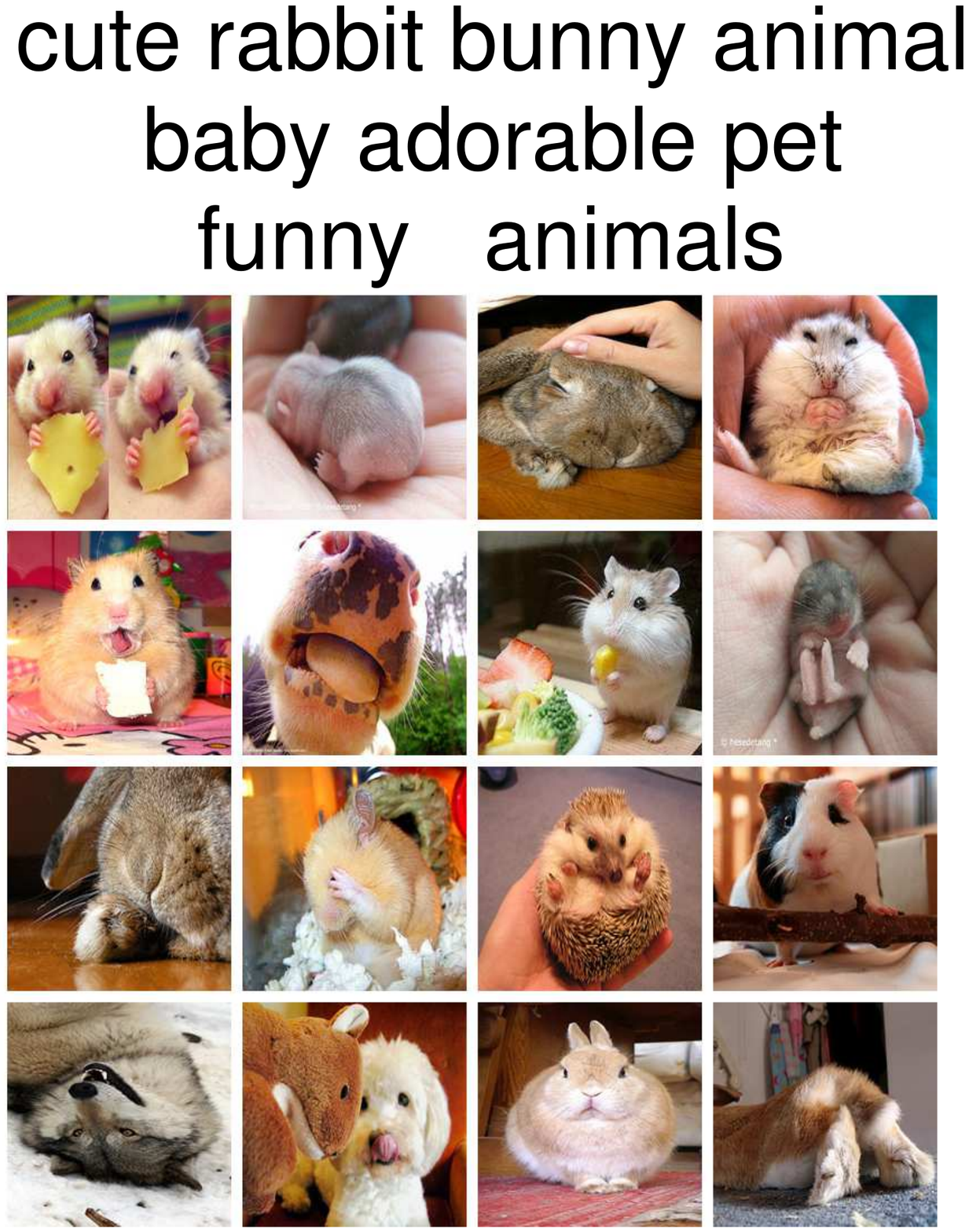}
   \includegraphics[width=1.6in, trim= 25mm 70mm 25mm 0mm]{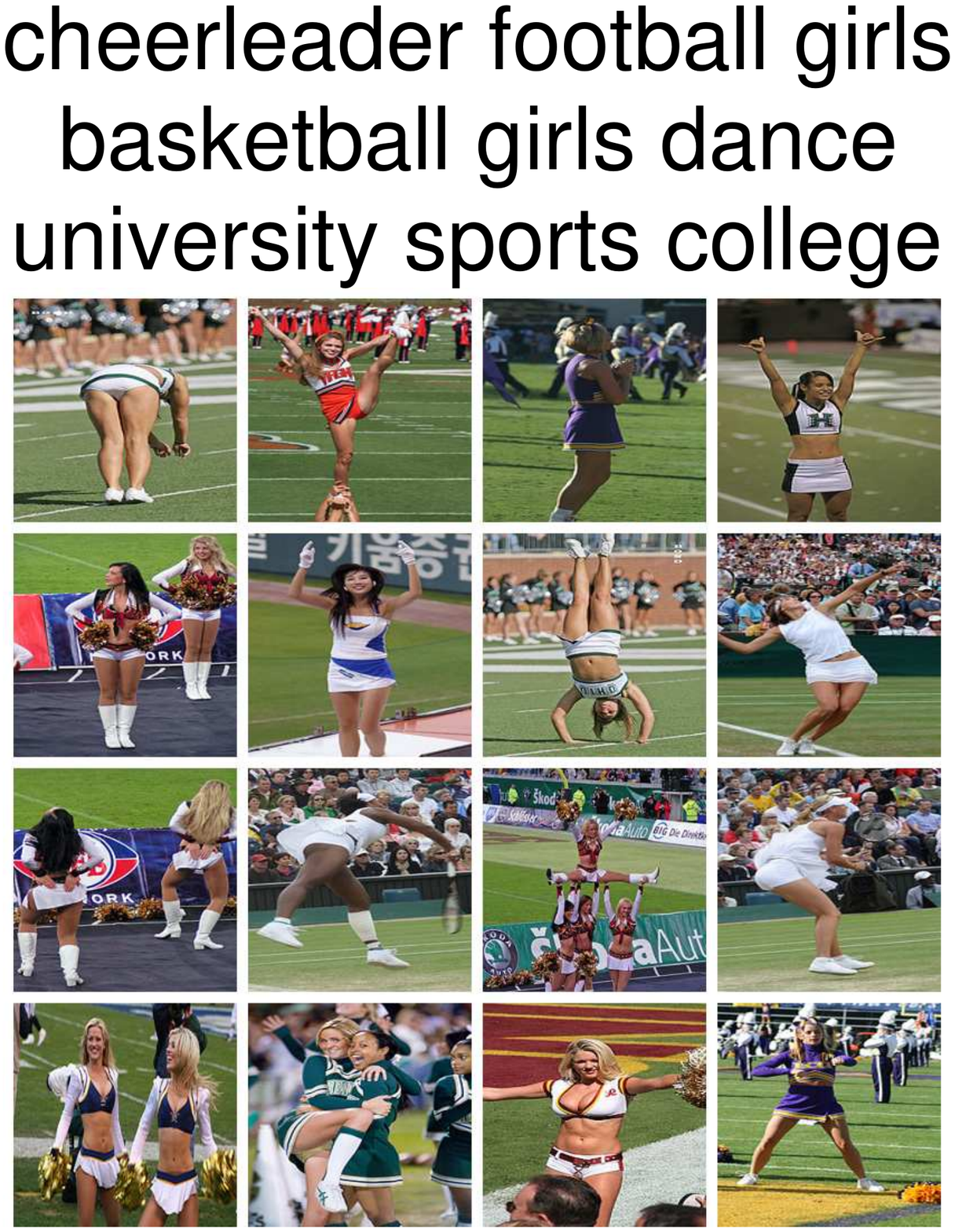}
   \includegraphics[width=1.6in, trim= 25mm 70mm 25mm 0mm]{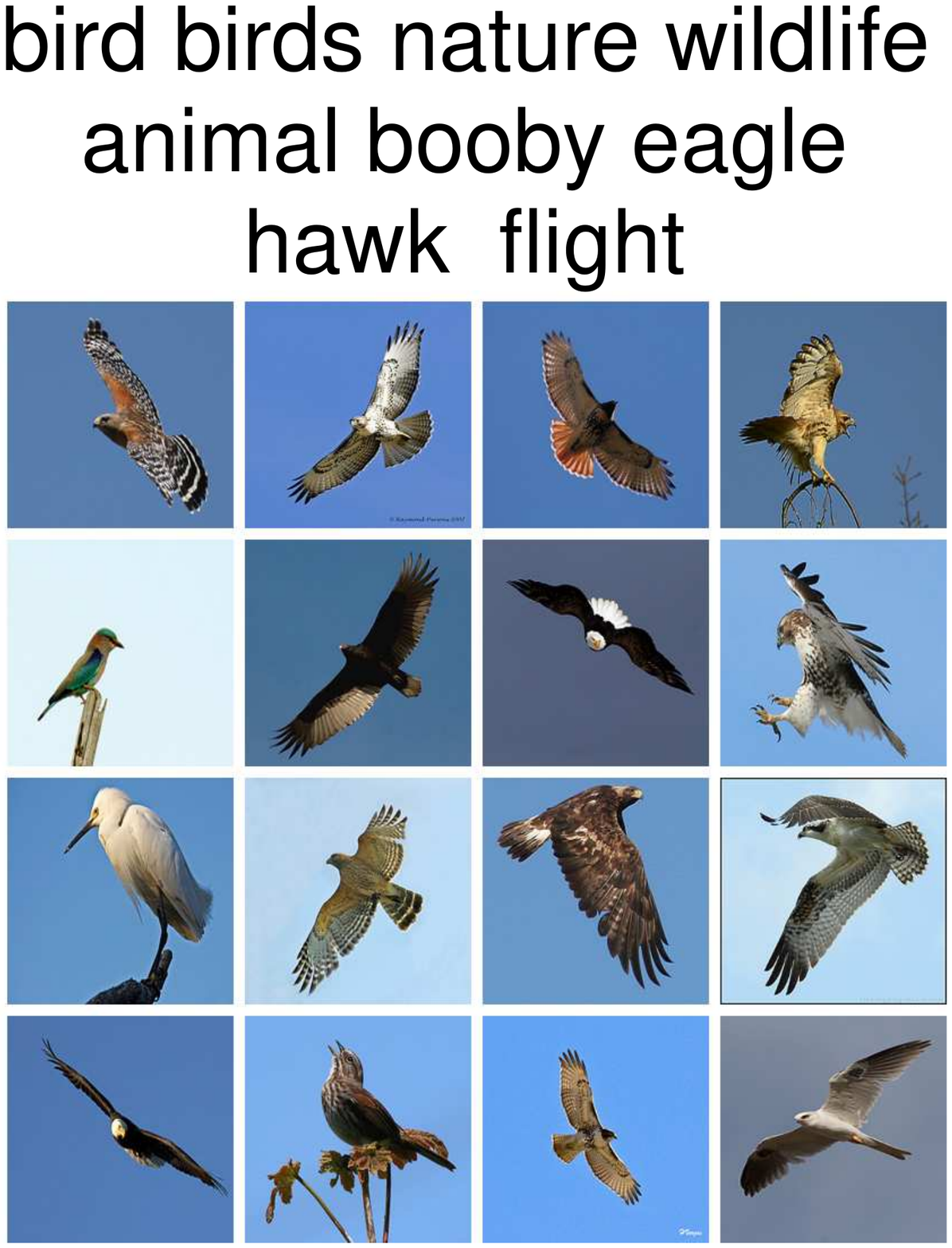}
   \includegraphics[width=1.6in, trim= 25mm 70mm 25mm 0mm]{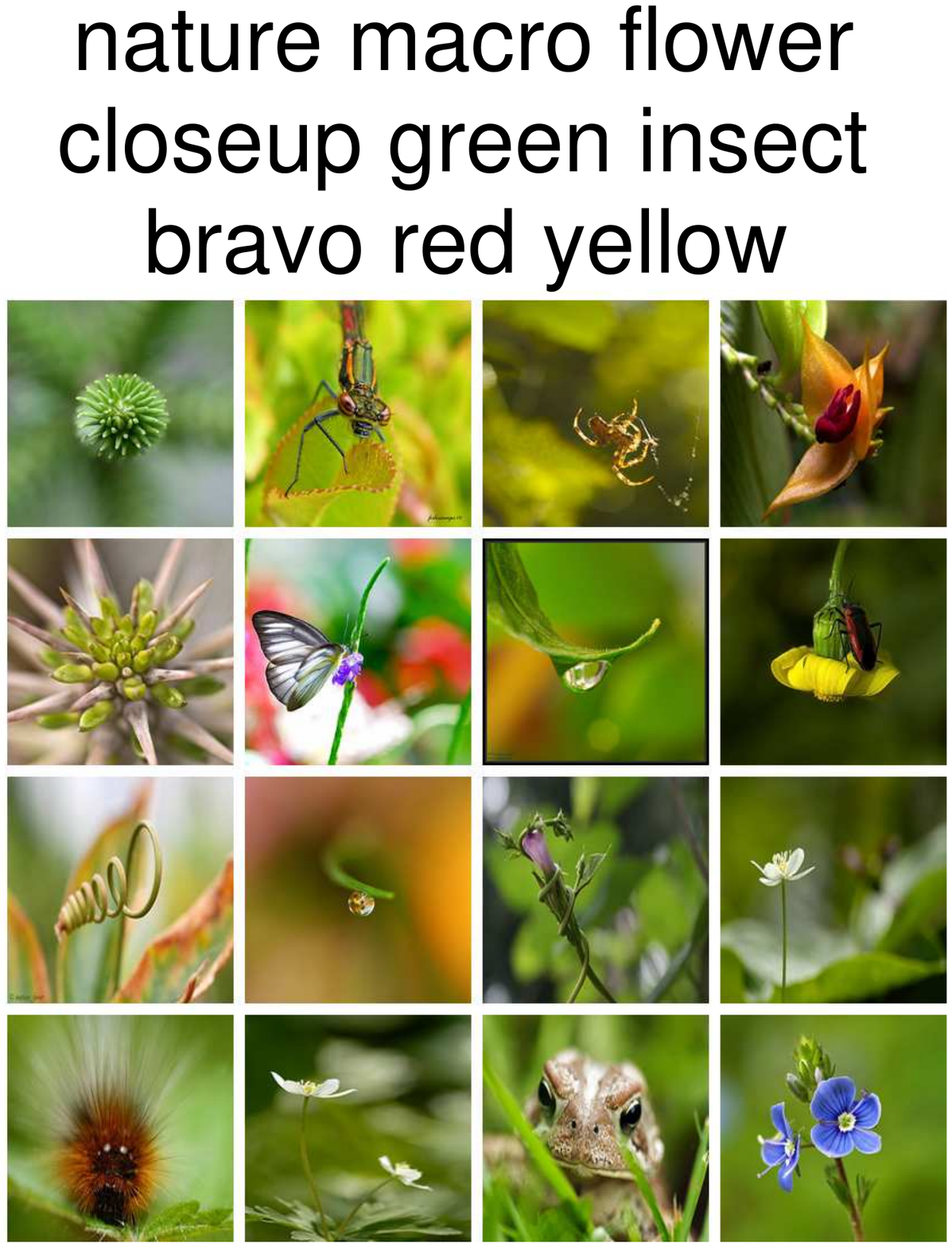}
   \includegraphics[width=1.6in, trim= 25mm 70mm 25mm 0mm]{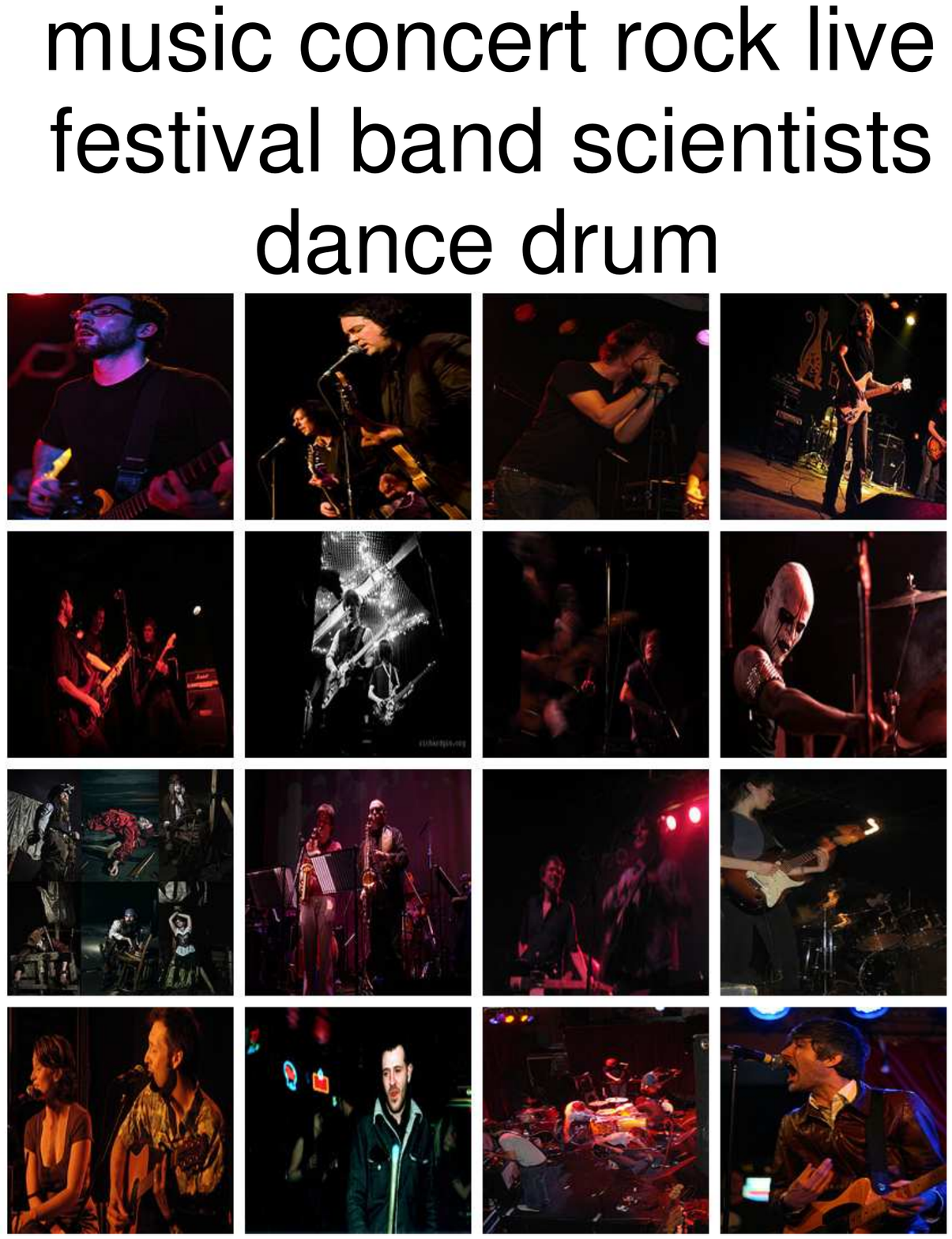}
   \includegraphics[width=1.6in, trim= 25mm 70mm 25mm 0mm]{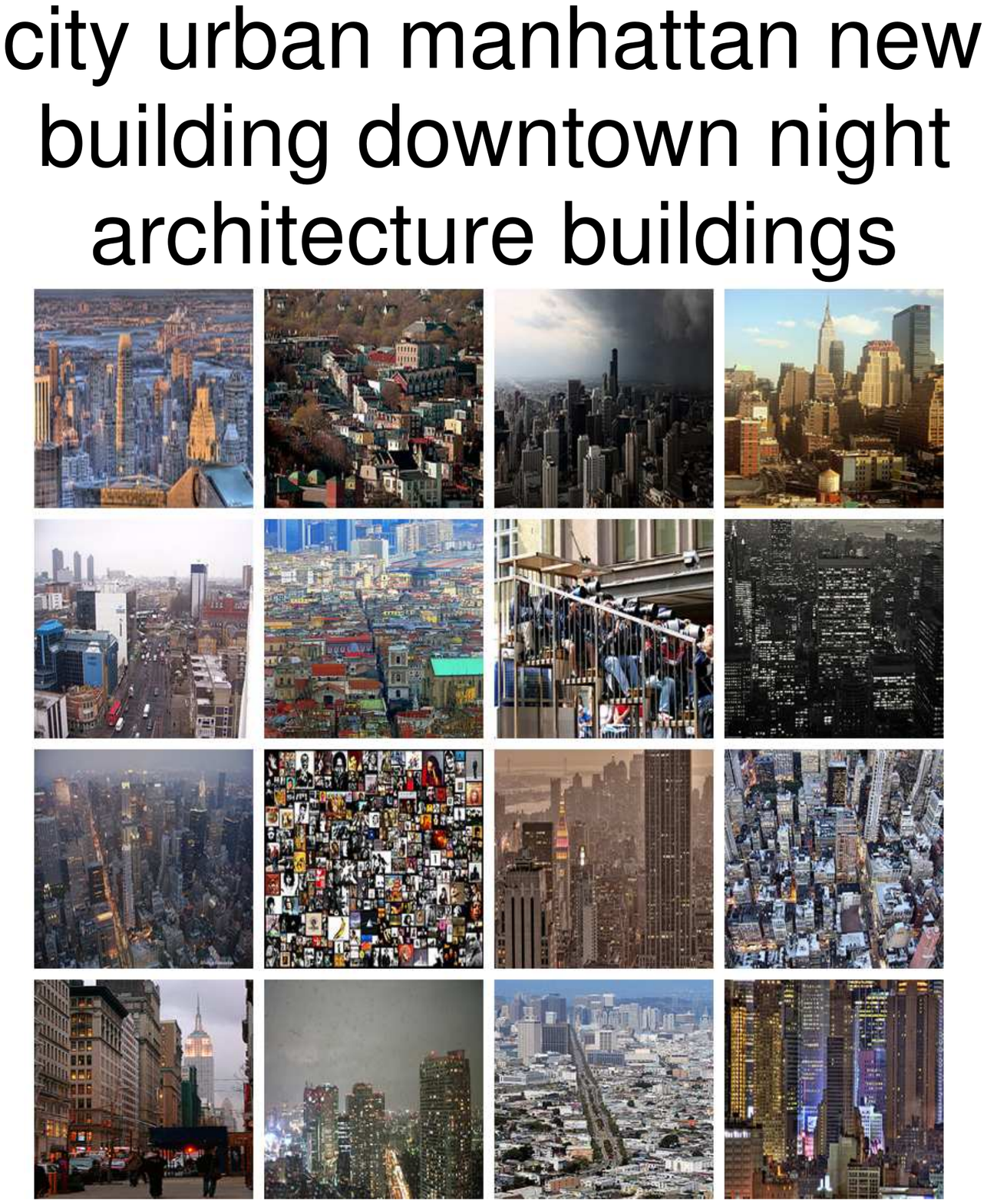}
   \includegraphics[width=1.6in, trim= 25mm 70mm 25mm 0mm]{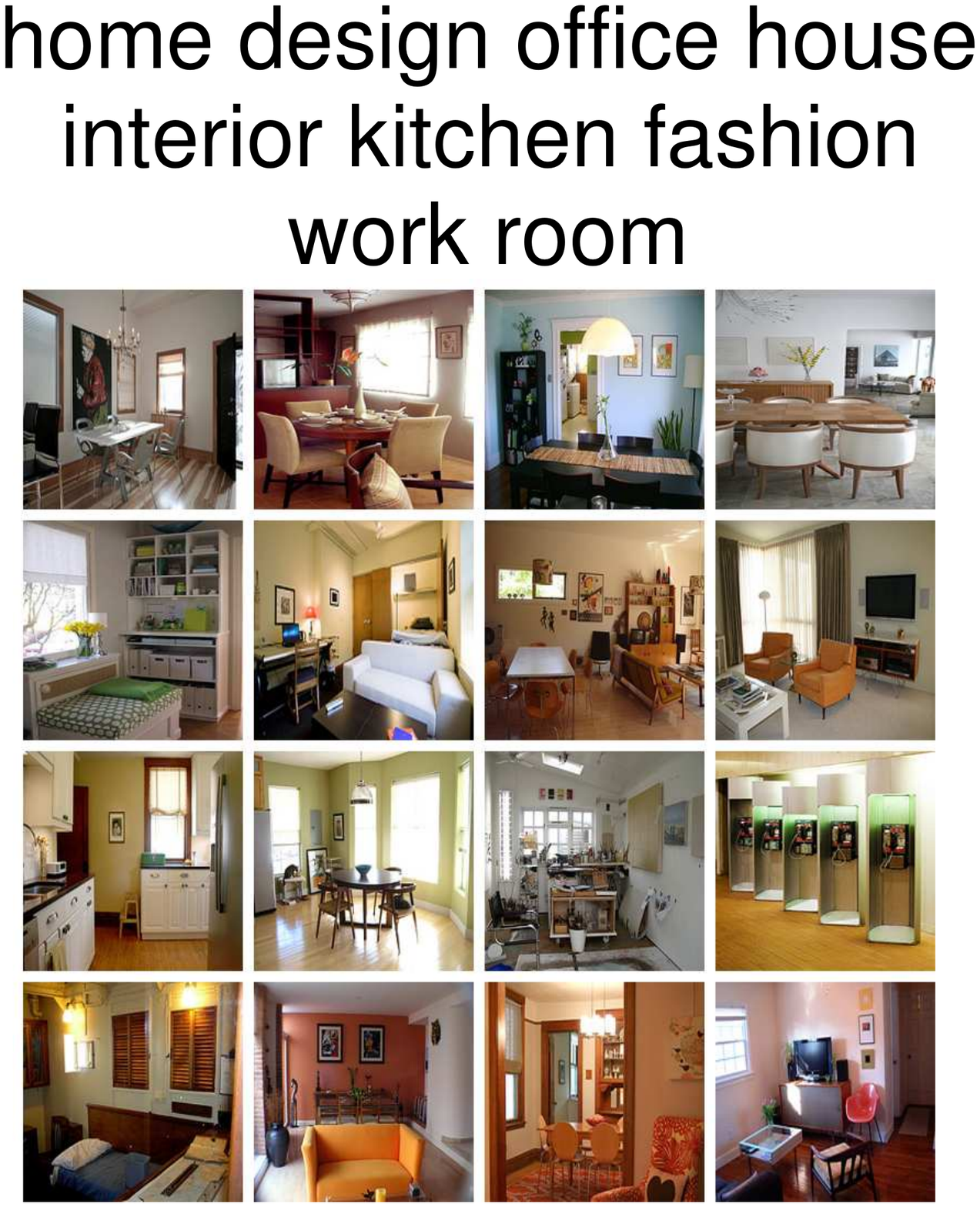}
   \includegraphics[width=1.6in, trim= 25mm 70mm 25mm 0mm]{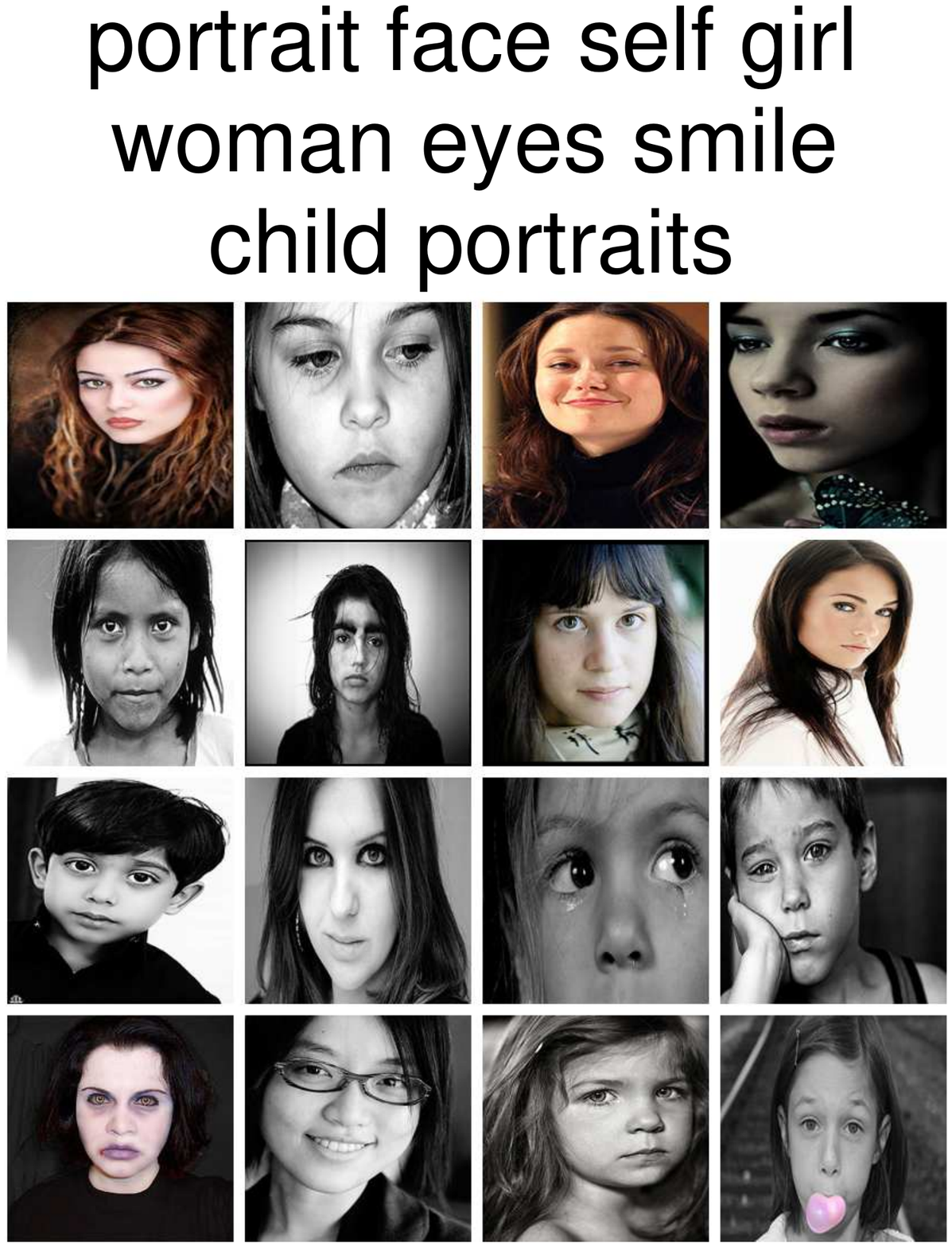}
   \includegraphics[width=1.6in, trim= 25mm 70mm 25mm 0mm]{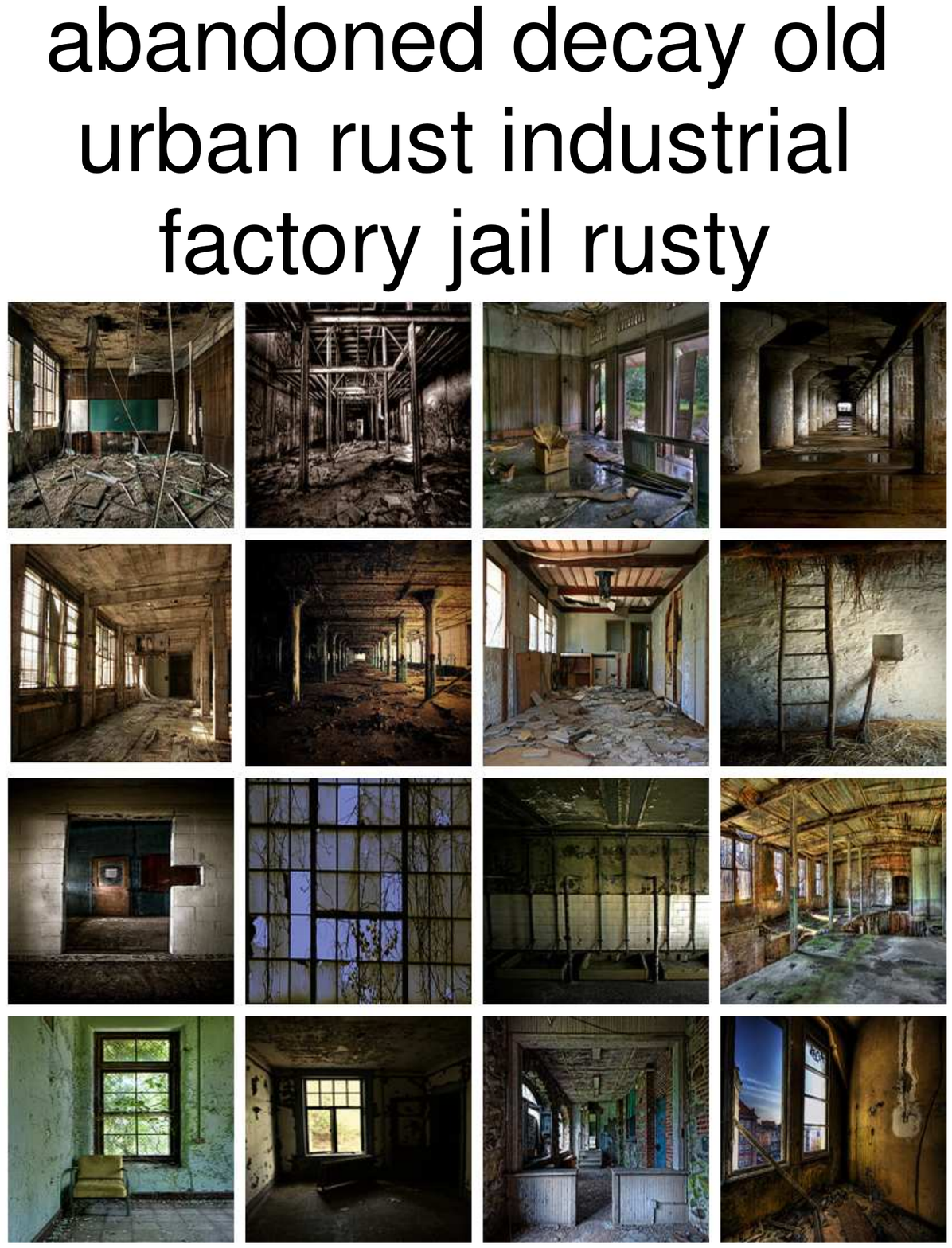}
   \includegraphics[width=1.6in, trim= 25mm 70mm 25mm 0mm]{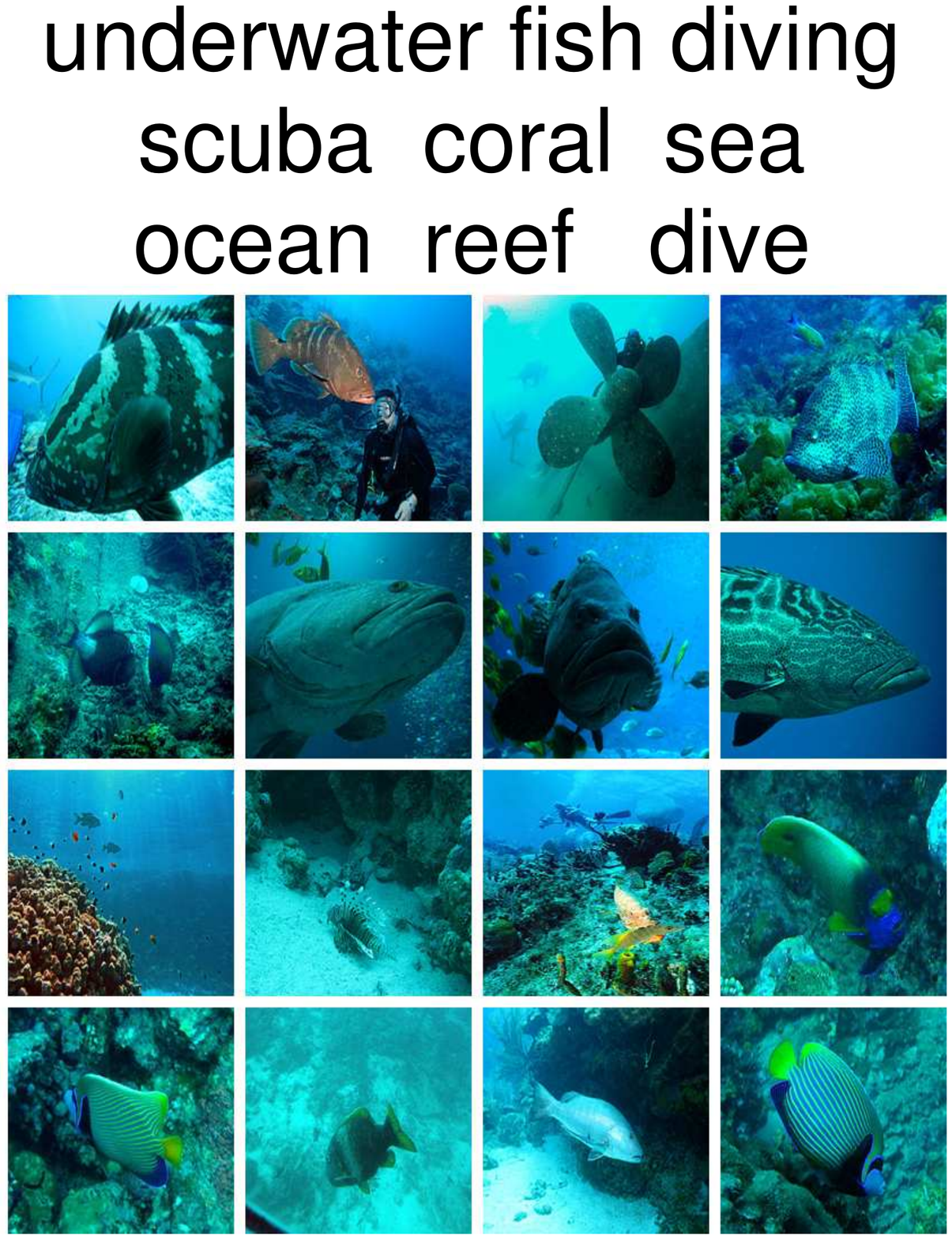}
   \includegraphics[width=1.6in, trim= 25mm 70mm 25mm 0mm]{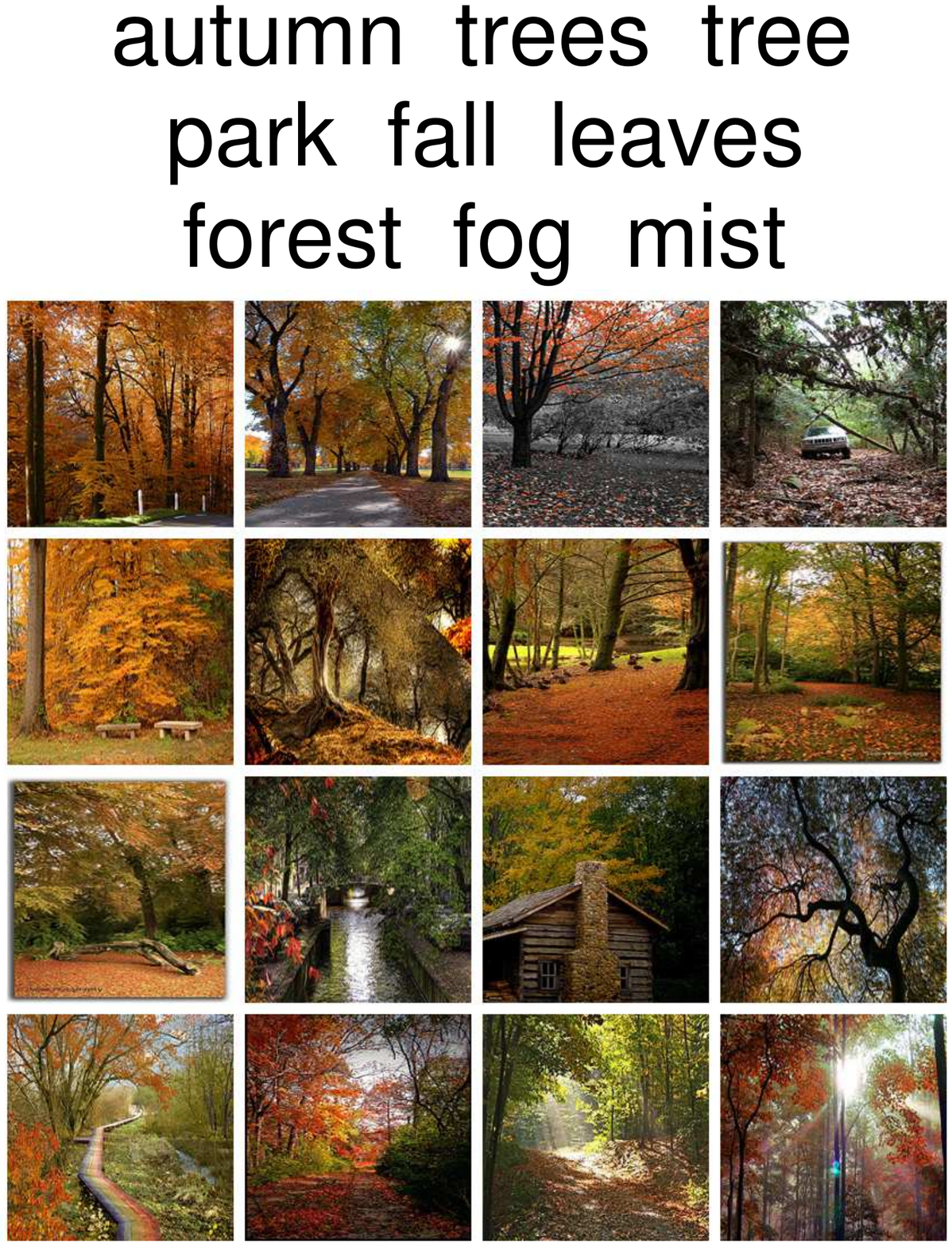}
   \includegraphics[width=1.6in, trim= 25mm 70mm 25mm 0mm]{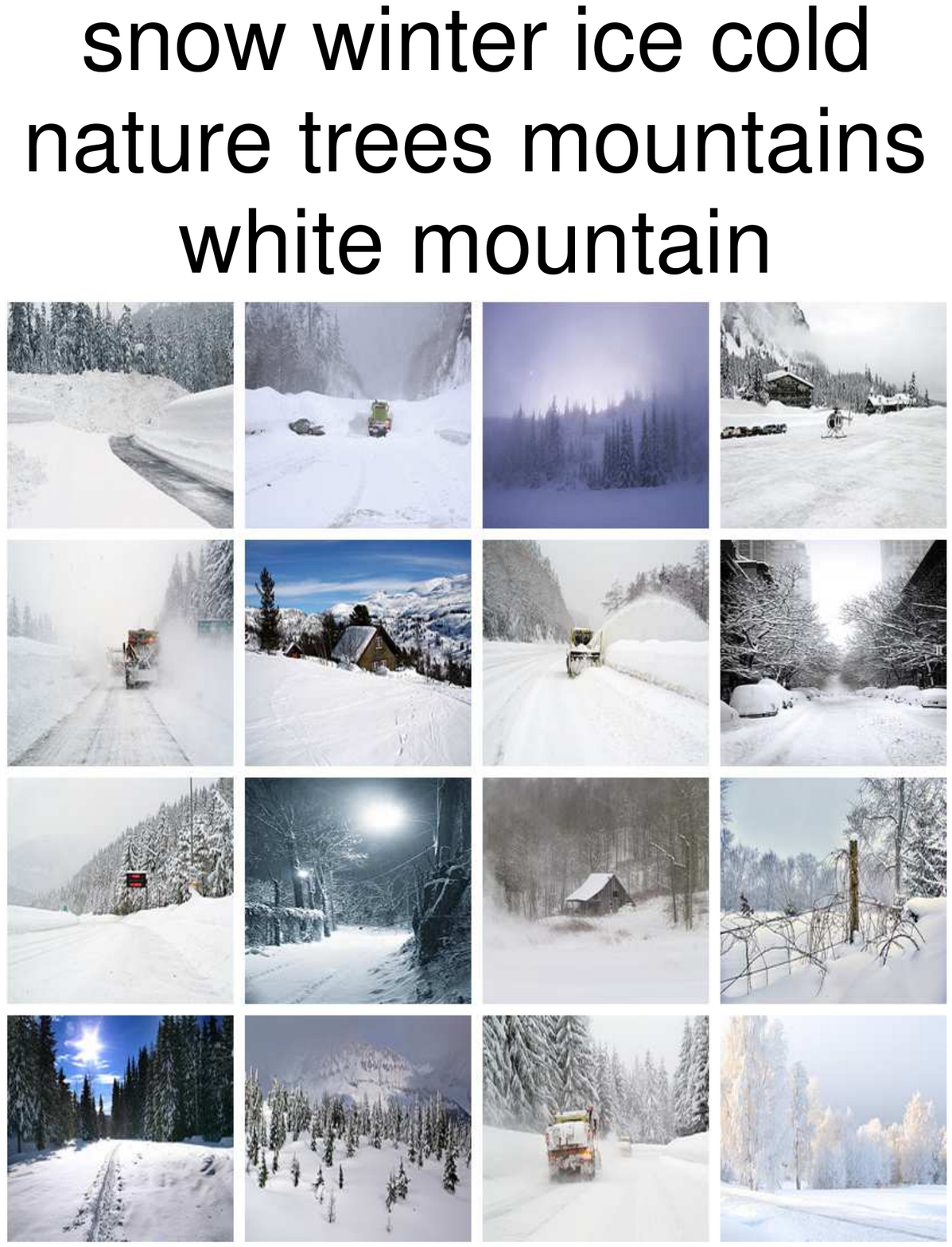}
   \caption{Example semantic clusters on the NUS-WIDE dataset. For each cluster, we show the most frequent tags and the images closest to the cluster center in the CCA (V+T+C) space.}
   \label{fig:NUS_semantic}
\end{figure*}

\begin{figure*}[t]
   \centering
   \includegraphics[width=1.6in, trim= 25mm 70mm 25mm 20mm]{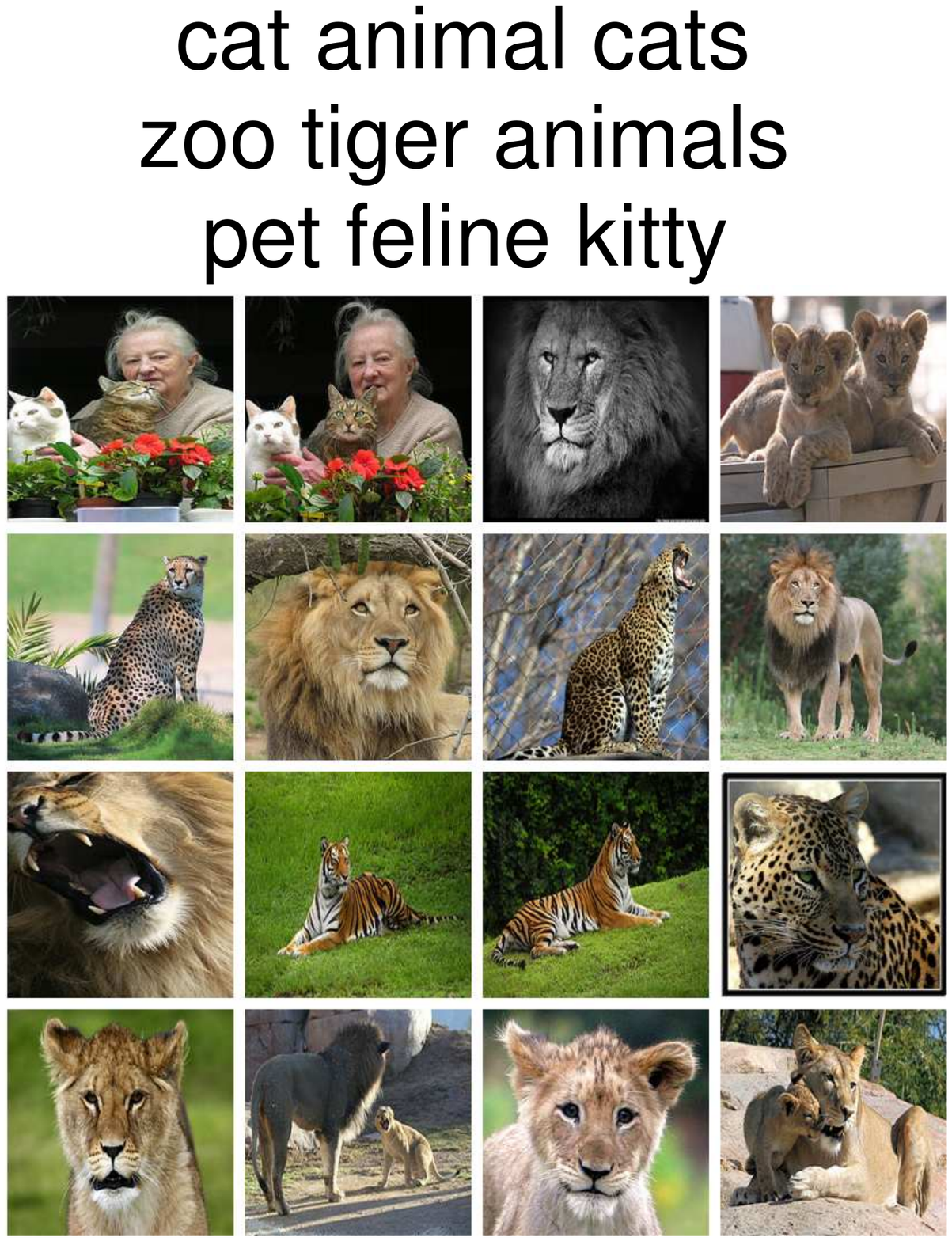}
   \includegraphics[width=1.6in, trim= 25mm 70mm 25mm 20mm]{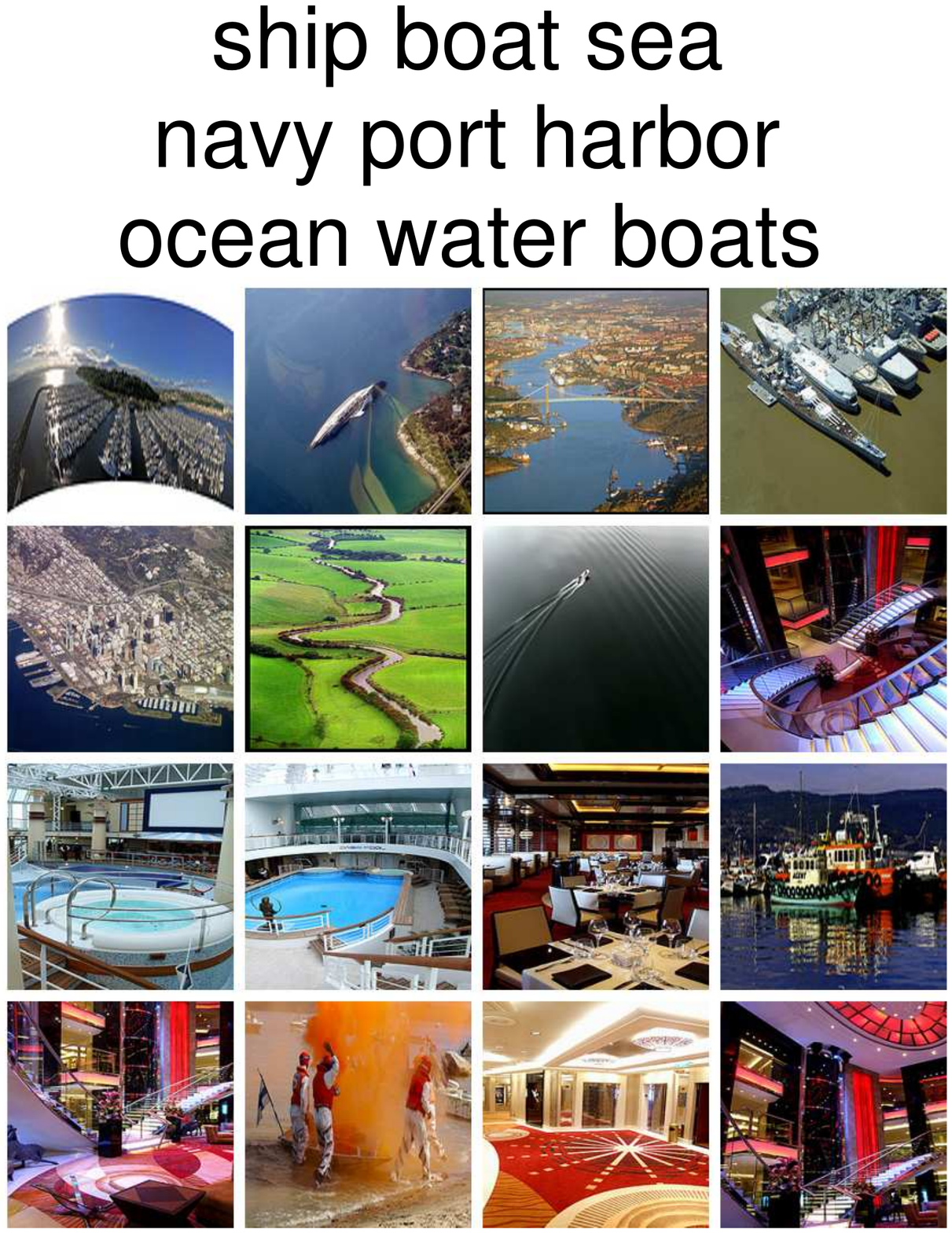}
   \includegraphics[width=1.6in, trim= 25mm 70mm 25mm 20mm]{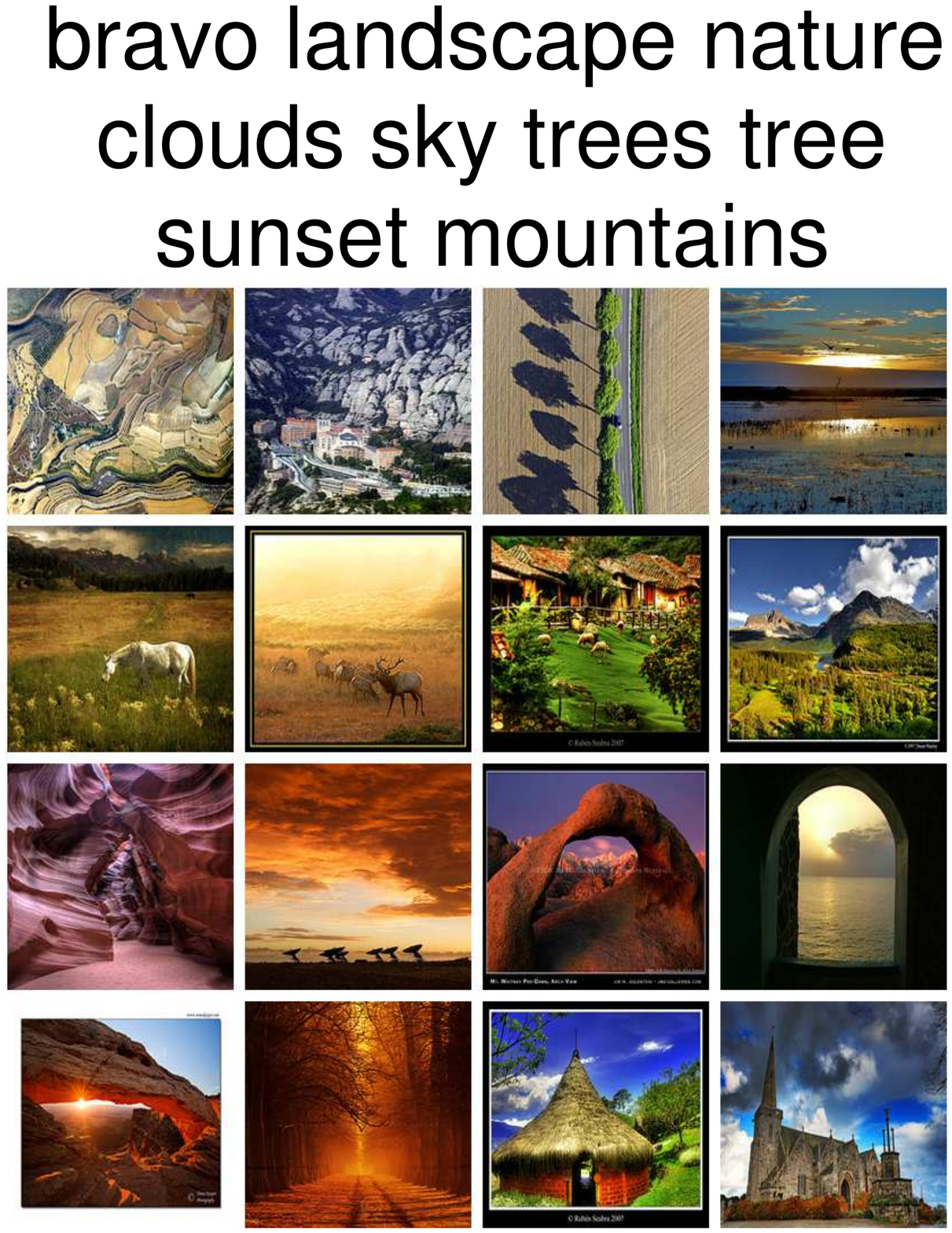}
   \includegraphics[width=1.6in, trim= 25mm 70mm 25mm 20mm]{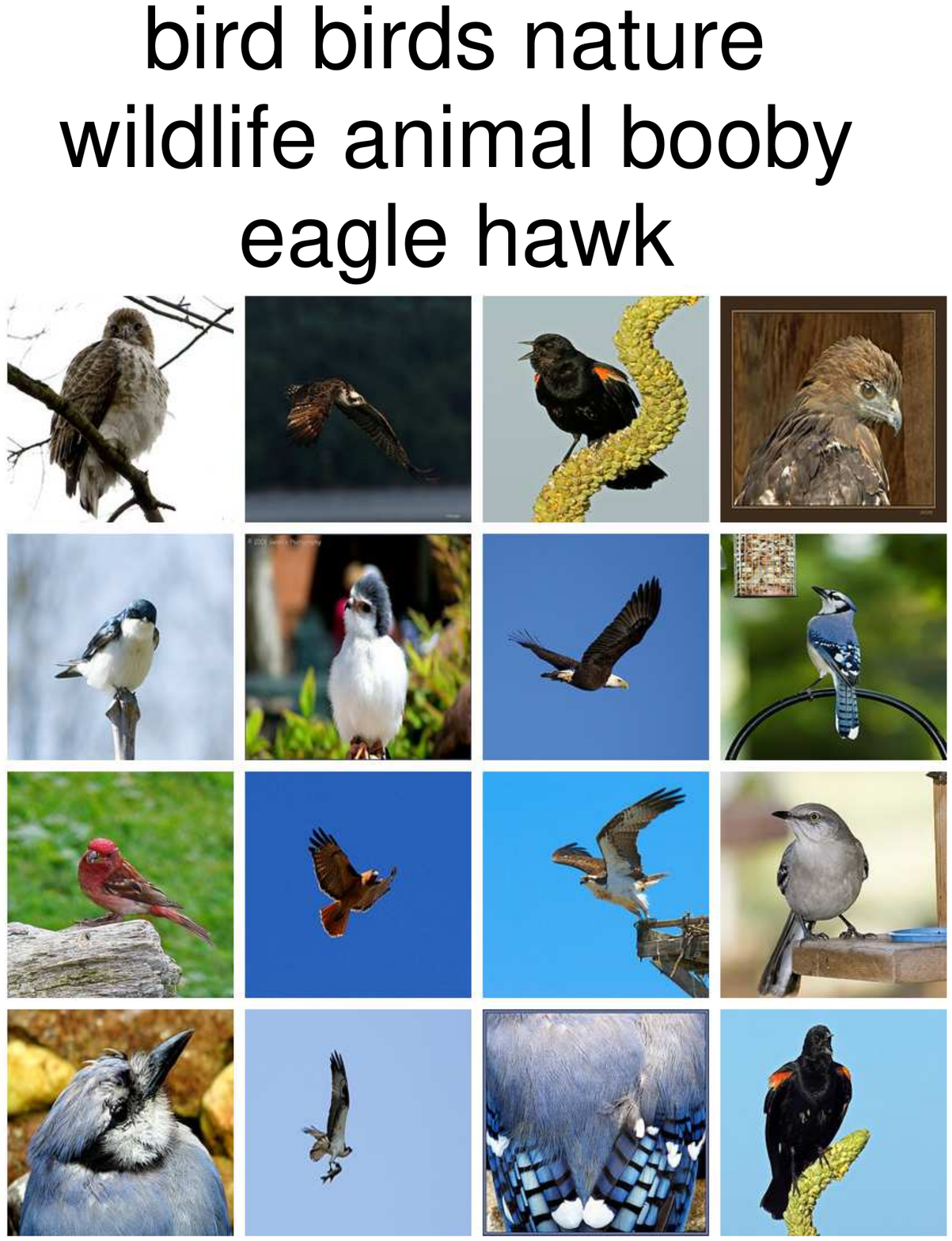}
   \caption{Example tag clusters produced by k-means clustering on top of normalized cuts (Section \ref{topics}). For each cluster center, the sixteen images with the closest tag vectors are shown. The most frequent tags in the cluster are shown above the central cluster images.}
   \label{fig:NUS_tag}
\end{figure*}

\begin{figure*}[t]
   \centering
   \includegraphics[width=1.6in, trim= 25mm 70mm 25mm 0mm]{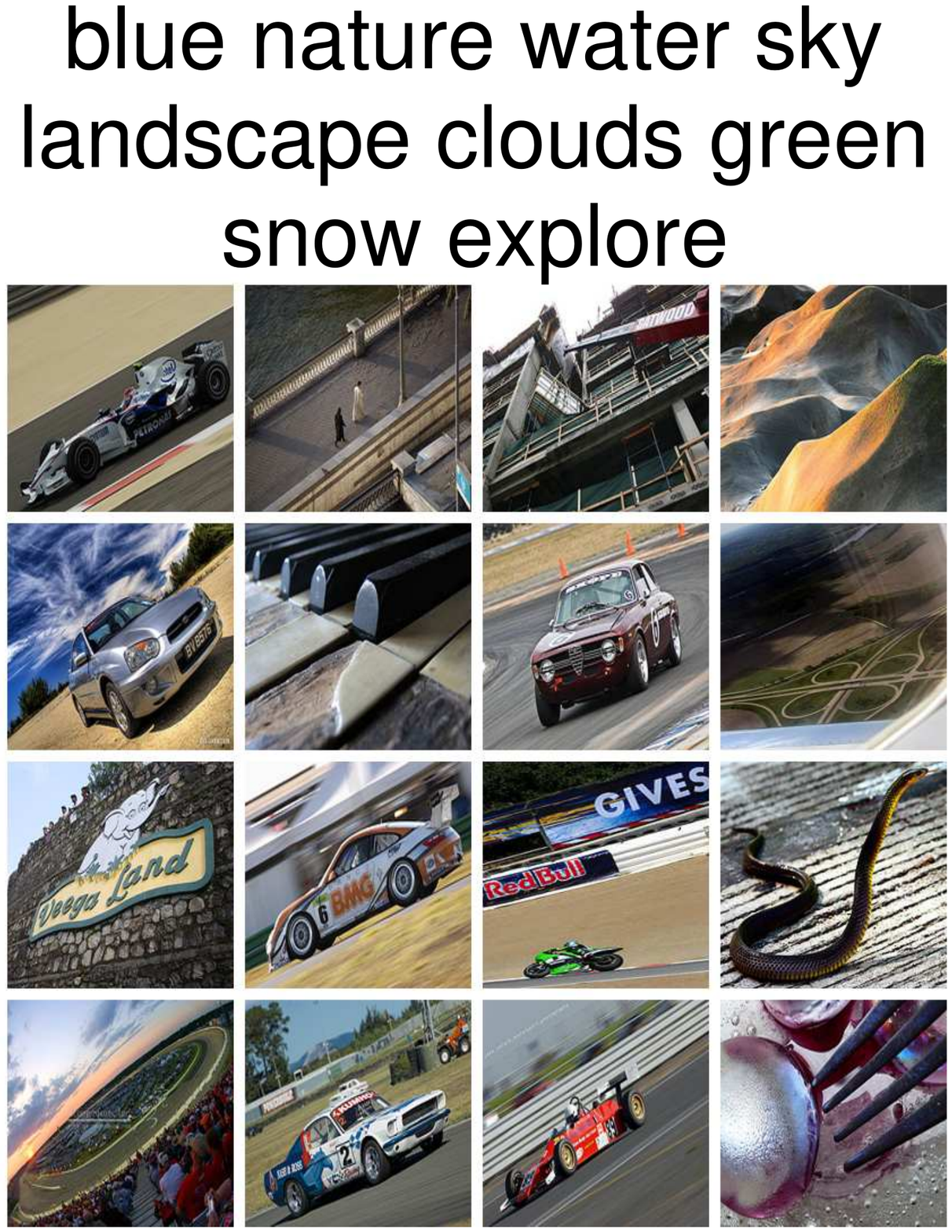}
   \includegraphics[width=1.6in, trim= 25mm 70mm 25mm 0mm]{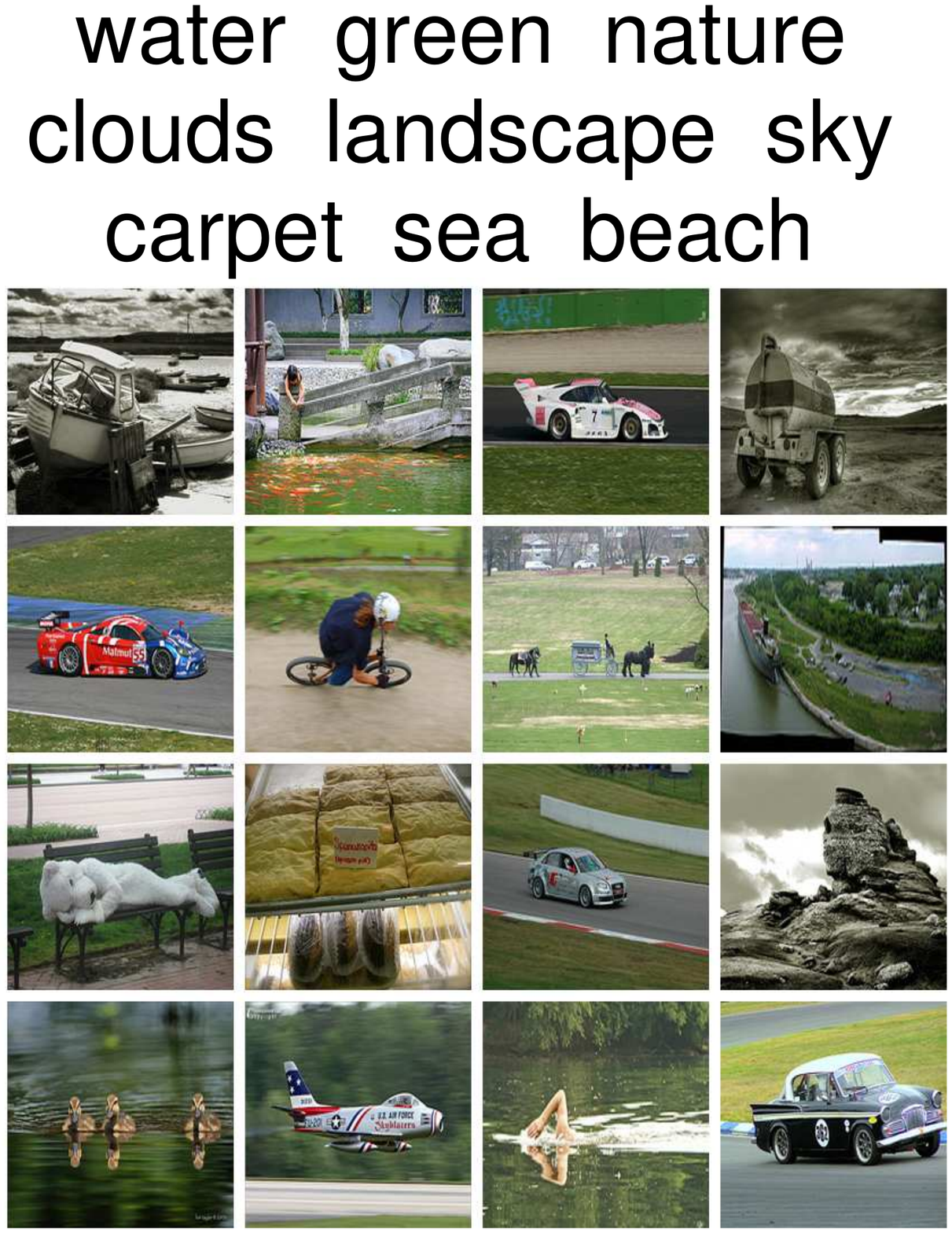}
   \includegraphics[width=1.6in, trim= 25mm 70mm 25mm 0mm]{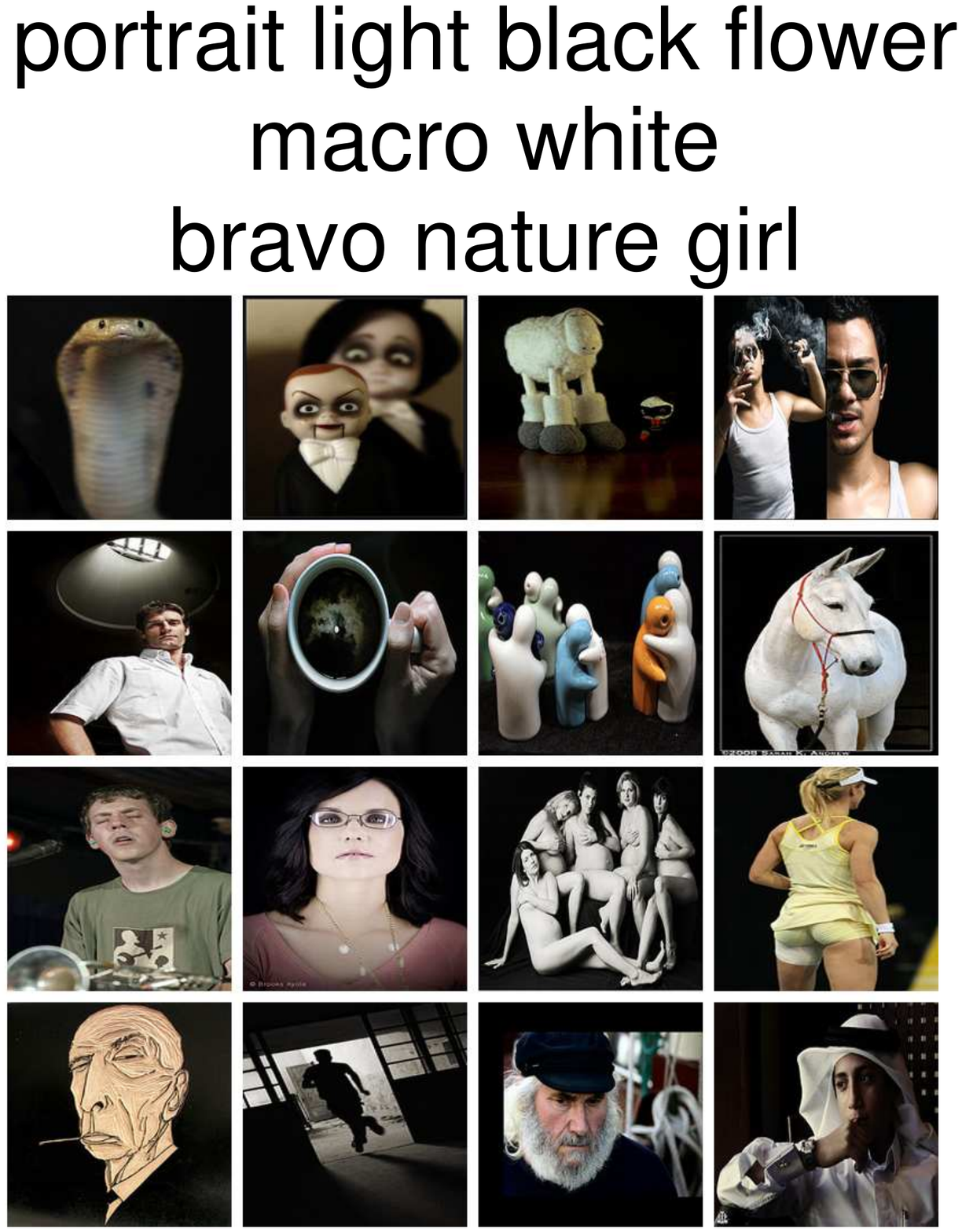}
   \includegraphics[width=1.6in, trim= 25mm 70mm 25mm 0mm]{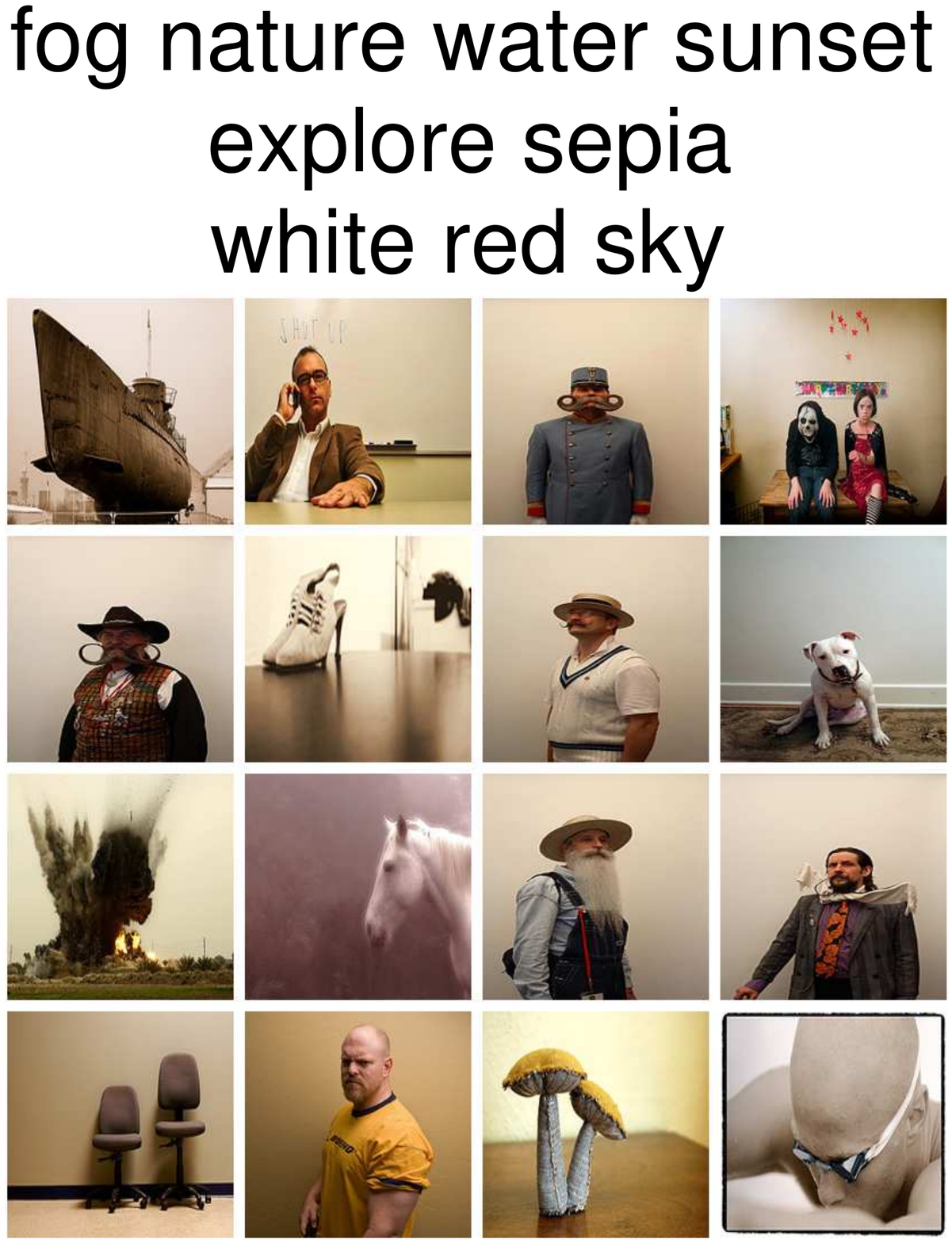}
   \caption{Example visual k-means clusters for the NUS-WIDE dataset.}
   \label{fig:NUS_visual}
\end{figure*}

\subsection{Comparison of tag clustering methods\label{compareTopicModel}}

In Section \ref{topics}, we have presented a number of tag clustering methods that can be used to obtain the semantic topic matrix $C$ for the third view when the training images do not come with ground-truth semantic information. Table \ref{topic1} compares the performance of these methods. We use our proposed CCA (V+T+C) model, where $V$ and $T$ are low-dimensional approximations to the visual and tag features as discussed in Sections \ref{sec:visual_kernel} and \ref{sec:tag_approximation}, and $C$ is generated by the different clustering methods being compared. As a baseline, we also include results for k-means clustering based solely on visual features. It is clear that visual clusters have worse performance than all of the tag-based clustering methods, thus confirming that the visual features are too ambiguous for unsupervised extraction of high-level structure (this will also be demonstrated qualitatively in Figure \ref{fig:NUS_visual}). In fact, the performance of the three-view CCA (V+T+C) model with visual clusters is even worse than that of the CCA (V+T) baseline.

Among the tag-based clustering methods, normalized cut (NC) method achieves the best performance across a wide range of cluster sizes, followed by NMF, k-means and pLSA in decreasing order of accuracy. Thus, NC will be our method of choice for computing the CCA (V+T+C) embedding. As a function of the number of clusters, accuracy tends to increase up to a certain point, after which overfitting sets in. The best number of clusters to use depends on the breadth of coverage and the semantic structure of the dataset; we select it using validation data.

Figure \ref{fig:NUS_semantic} visualizes a few tag clusters in the joint CCA space for the NUS-WIDE dataset. For each example cluster, it shows the most frequent tags associated with the images in that cluster, as well as the sixteen images closest to the center of the cluster in the CCA (V+T+C) space. We can see that the clusters have a good degree of both visual and semantic coherence. For comparison, Figure \ref{fig:NUS_tag} shows some clusters in just the tag view, i.e., before the joint CCA projection. We can see the images in each cluster correspond to the same semantic concept, though they are not visually similar. Finally, for completeness, Figure \ref{fig:NUS_visual} shows k-means clusters on just the visual features. Because our visual features are relatively powerful, the cluster images are still perceptually similar, but they are no longer semantically consistent (in particular, note the poor correspondence between the most frequent tags for the entire clusters and the top sixteen images in the clusters).
Figure \ref{fig:NUS_visual} confirms the difficulty of finding good semantic themes by visual clustering alone, and helps to explain why visual clusters decrease the retrieval performance when incorporated as a third view in the learned CCA model (Table \ref{topic1}).

\subsection{Comparison of similarity functions \label{distancemetric}}

\begin{table}[]
{
\hfill{}
\begin{tabular}{l||cc}
\hline
method   &    I2I        &   T2I  \\
\hline
 CCA (V+T) (Eucl)  &      45.69         &      51.32 	      \\
  CCA (V+T) (scale+Eucl)  &     48.60      &    54.43 	      \\
 CCA (V+T) (scale+corr)  &   	\textbf{54.90}      &  	\textbf{64.07}  	      \\
 \hline
CCA (V+T+C)  (Eucl) &        53.61     &    69.69    \\
  CCA (V+T+C) (scale+Eucl)  &   57.06           &    72.42 	      \\
CCA (V+T+C) (scale+corr) &    \textbf{ 	62.44	}     &     \textbf{	75.92 }    \\
 \hline
CCA (V+T+K)  (Eucl)  &	    52.41	        &      71.47               	\\
  CCA (V+T+K) (scale+Eucl)  &   57.47         &    75.23     \\
CCA (V+T+K)  (scale+corr)  &	    \textbf{62.56}	 	        &          	\textbf{78.88}	      	\\
\hline
\end{tabular}}
\hfill{}
\caption{Evaluation of different components of our proposed similarity function (eq. \ref{distancefunction}) on three multi-view setups. ``Eucl'' denotes Euclidean distance, ``scale'' denotes scaling of the feature dimensions by the CCA eigenvalues, and ``corr'' denotes normalized correlation. CCA (V+T+C) is computed using 20 NC clusters and CCA (V+T+K) is the supervised three-view model with K given by search keywords of the Flickr images.}
\label{distance}
\end{table}

\begin{table}[t]
{
\hfill{}
\begin{tabular}{l||ccc}
\hline
method   &    I2I        &   T2I    & K2I \\
\hline
V-full   &   	 33.68       &   --       &      --      \\
V   &   	41.65      &   --       &      --     \\
\hline
 CCA (V+T)  &   	54.90      &  	64.07 	       &         95.60     \\
 CCA (V+K)   &	    	61.77         &          --          &    92.60      	\\
CCA (V+T+K)   &	   	 62.56	        &    78.88    	       &    97.20         	\\
CCA (V+C)  &     	61.69	     &   --       &      --     \\
CCA (V+T+C)  &     62.44    &    75.92         & 97.80  \\
\hline
Structural learning &	 	57.77	 	&	--	      &      --     \\
Wsabie & 	57.15 		 & 	 --     &    --  \\
\hline
\end{tabular}}
\hfill{}
\caption{Results on Flickr-CIFAR for image-to-image ({\bf I2I}), tag-to-image ({\bf T2I}), and keyword-to-image({\bf K2I}) retrieval.
The protocols for I2I and T2I are described in Section \ref{sec:protocol}. For K2I, each of the 10 ground truth classes is used as a query once.
The evaluation metric is average precision (\%) at top 50 retrieved images.
{\bf V-full} refers to the concatenated 38,512-dimensional visual features. In all the other approaches, {\bf V} refers to the 4,500-dimensional PCA-reduced features, {\bf T} to the 500-dimensional sparse SVD-reduced tag features, and {\bf C} is computed based on 20 NC clusters. {\bf Structural learning} refers to the method of~\cite{Ando05,Quattoni07} and {\bf Wsabie} refers to the method of~\cite{Weston11}.
We have obtained standard deviations from five random database/query splits, and they are around 0.25\% - 1\%.}
\label{main}
\end{table}

Section \ref{metric} has presented our similarity function for nearest neighbor retrieval in the CCA space.
Table \ref{distance} compares this function (eq. \ref{distancefunction}) to plain Euclidean distance for three different multi-view setups. We separately evaluate the effects of its two main components: eigenvalue scaling and normalized correlation. From the table, we can find that both these components give signficant improvements over the Euclidean distance.
We have consistently observed similar pattern on other datasets, so we adopt the proposed similarity function in all subsequent experiments.

\subsection{Comparison of multi-view models} \label{sec:retrieval}

Table \ref{main} evaluates the performance of several multi-view models on three tasks: image-to-image (I2I), tag-to-image (T2I), and keyword-to-image (K2I) retrieval. As explained in Section \ref{sec:protocol}, our performance metric for all tasks is class label (keyword) precision at top 50 images.

The most naive baselines for our approach are given by the single-view representations consisting only of visual features -- either raw 38,512-dimensional ones (V-full) or PCA-compressed 4,500-dimensional ones (V). Both of these representations can only be used directly for image-to-image similarity search (I2I). As can be seen from Table \ref{main}, the PCA-compressed feature gets higher precision for this task, but in absolute terms, both perform poorly.

A stronger baseline for our three-view models is given by the two-view CCA (V+T) representation, which can be used for all three retrieval tasks we are interested in (it can be used for K2I because the ten class labels or keywords in this dataset are a subset of the tag vocabulary). For I2I, the CCA (V+T) embedding improves the precision over non-embedded image features (V) from 41.65\% to 54.9\%. Thus, projecting visual features into a space that maximizes correlation with Flickr tags greatly improves the semantic coherence of similarity-based image matches (i.e., in the CCA space, ``truck'' query images are much more likely to have top matches that are also ``truck'' images).

\begin{figure*}[] 
   \centering
\includegraphics[width=0.7in, trim=85mm 80mm 65mm
80mm]{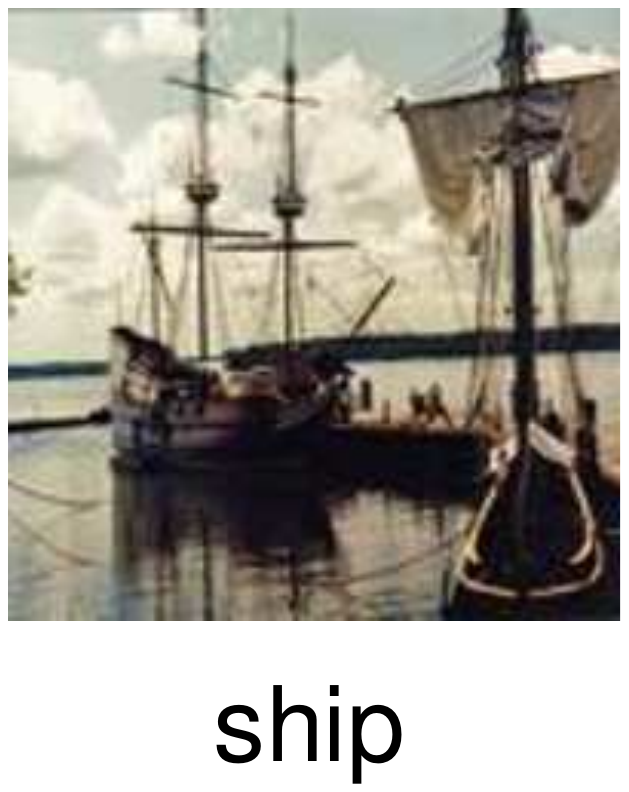}
  \includegraphics[width=1.9in, trim= 20mm 85mm 10mm
80mm]{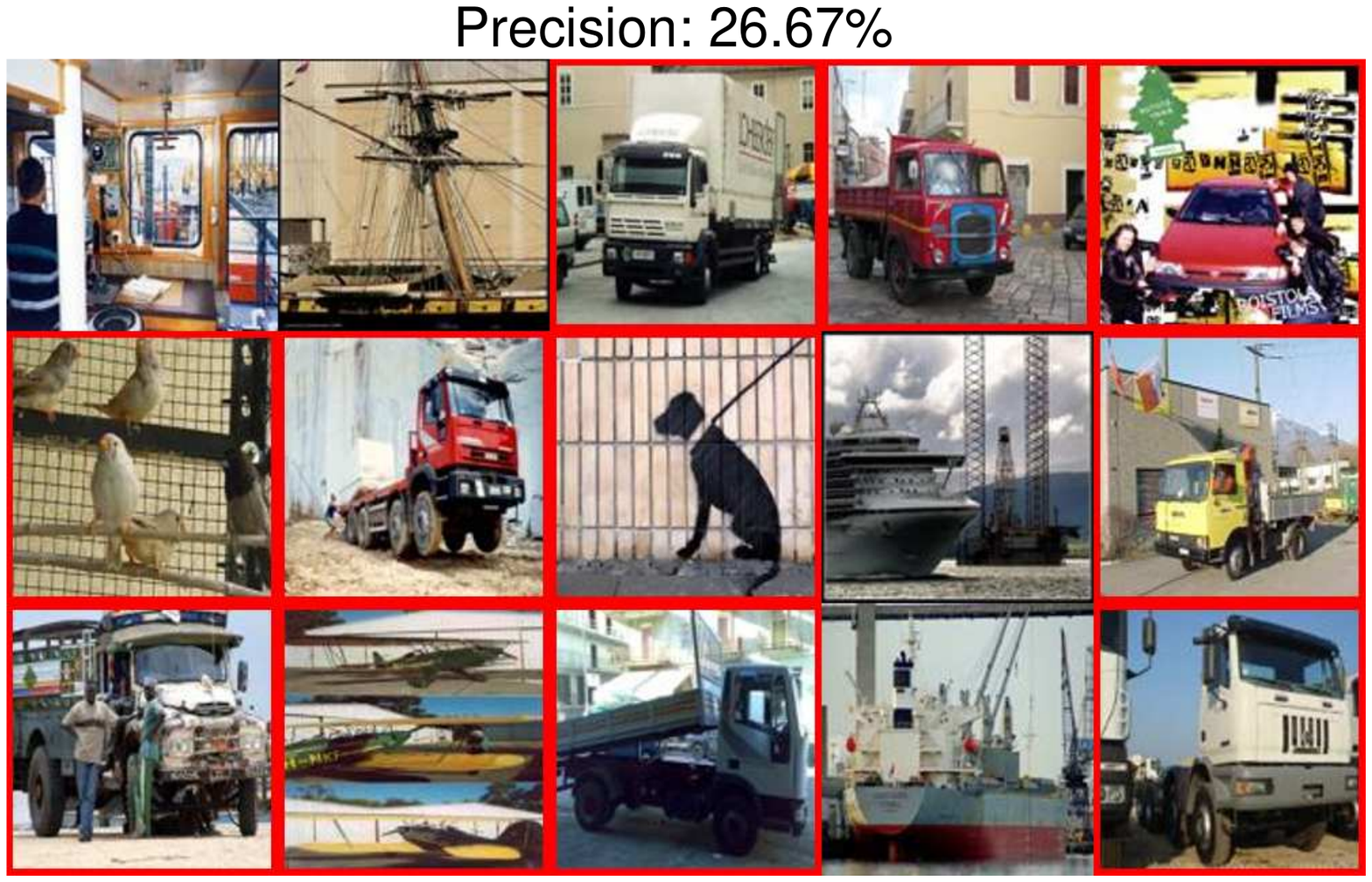}
   \includegraphics[width=1.9in, trim= 17mm 85mm 18mm
80mm]{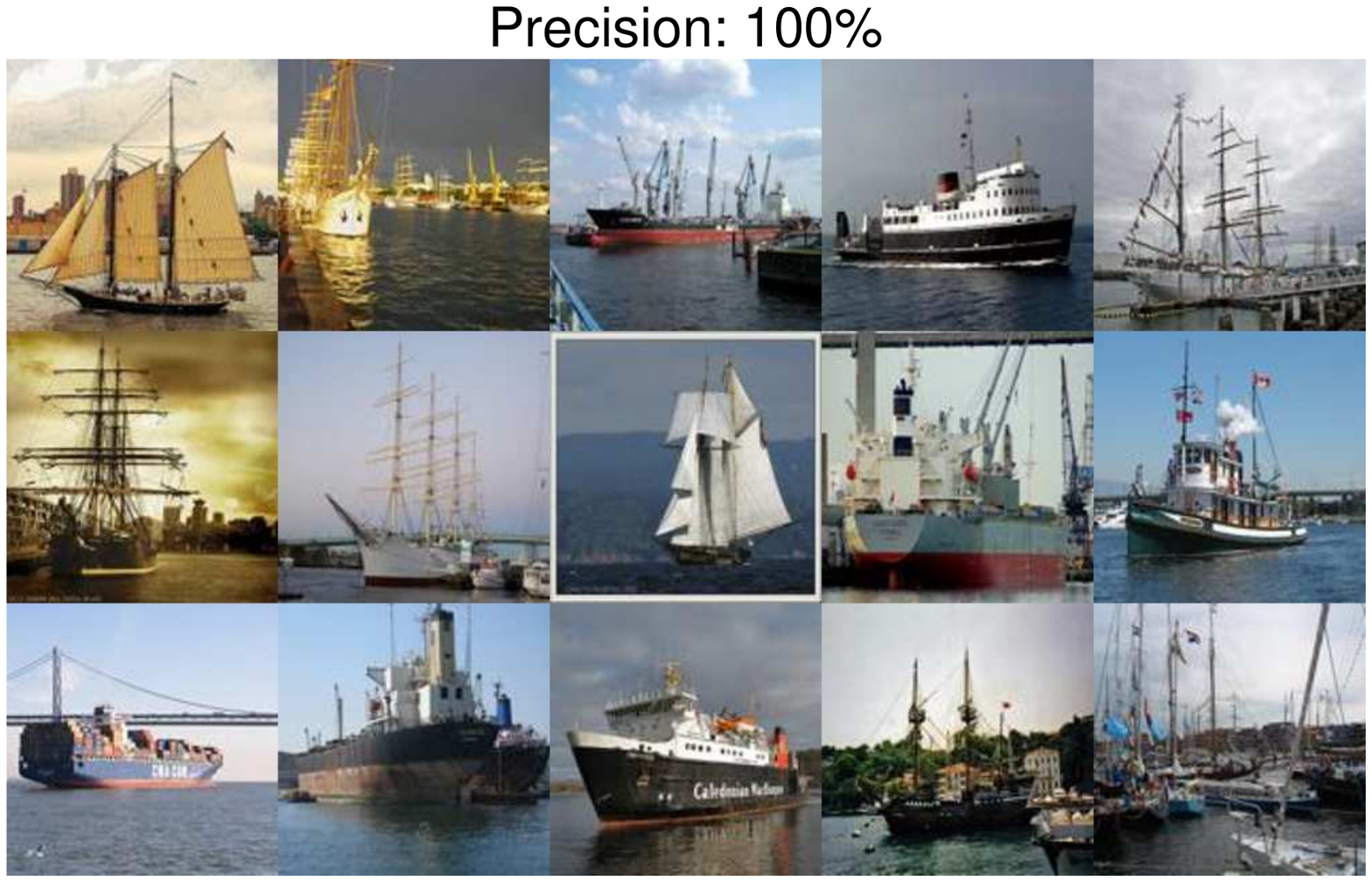}
   \includegraphics[width=1.9in, trim= 15mm 85mm 20mm
80mm]{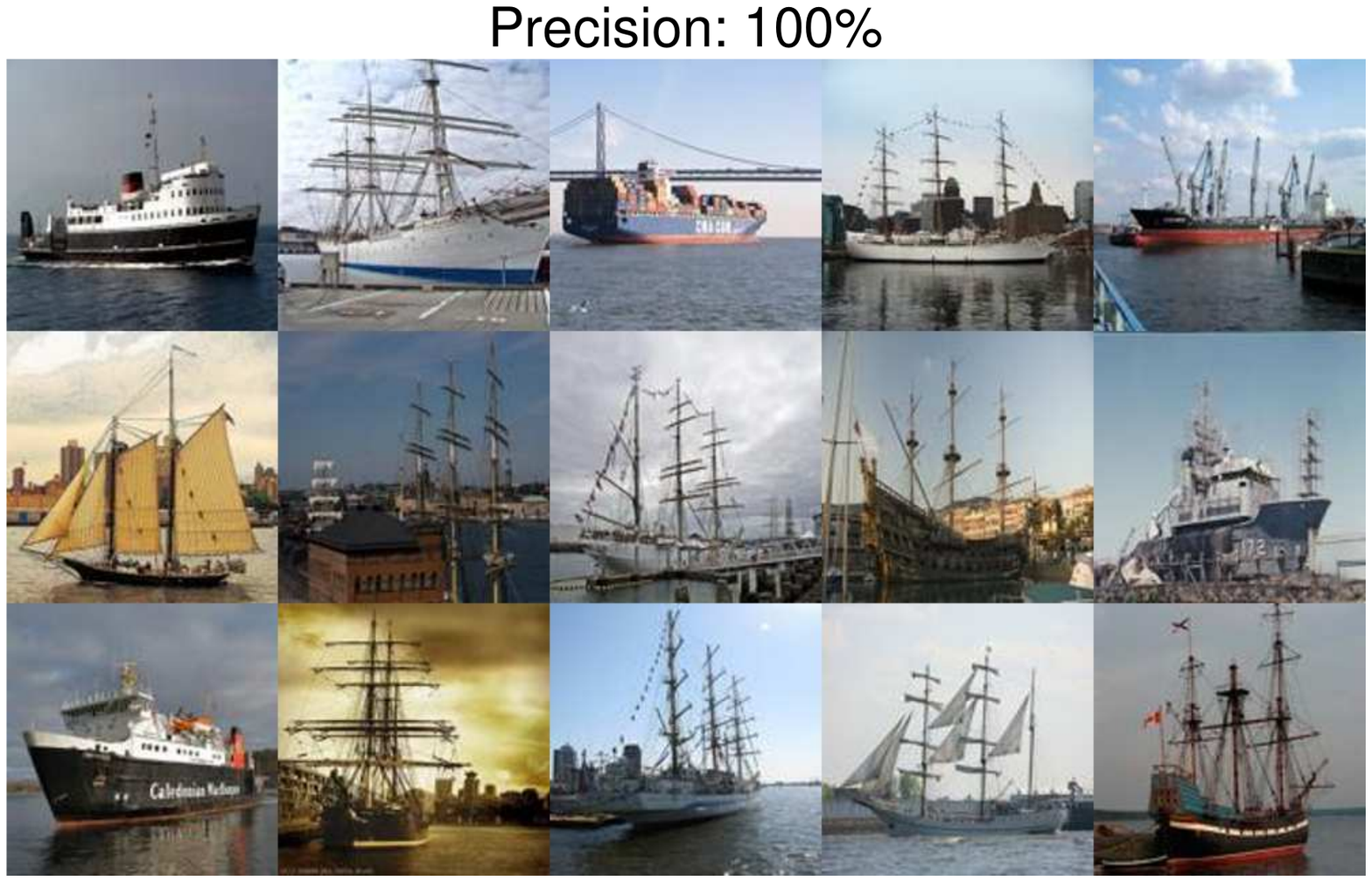}
   \includegraphics[width=0.7in, trim=75mm 80mm 75mm
80mm]{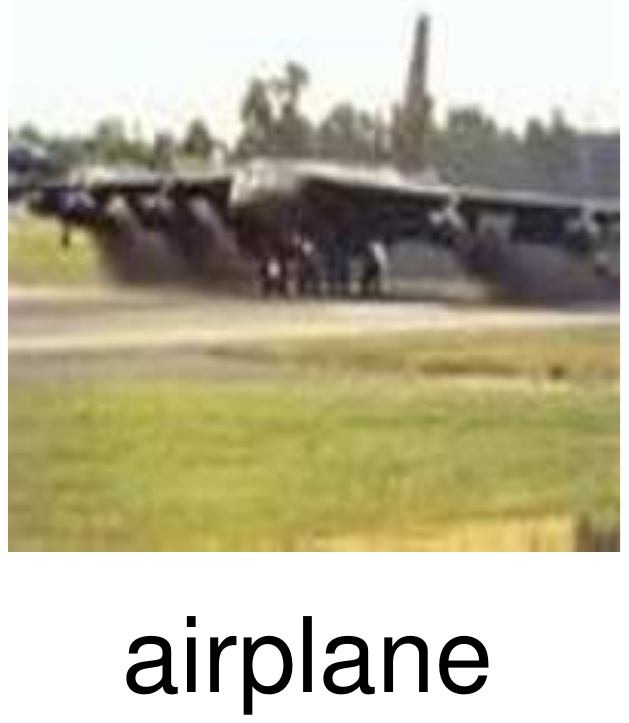}
\subfigure[Original visual feature.]{
   \includegraphics[width=1.9in, trim= 15mm 85mm 15mm
80mm]{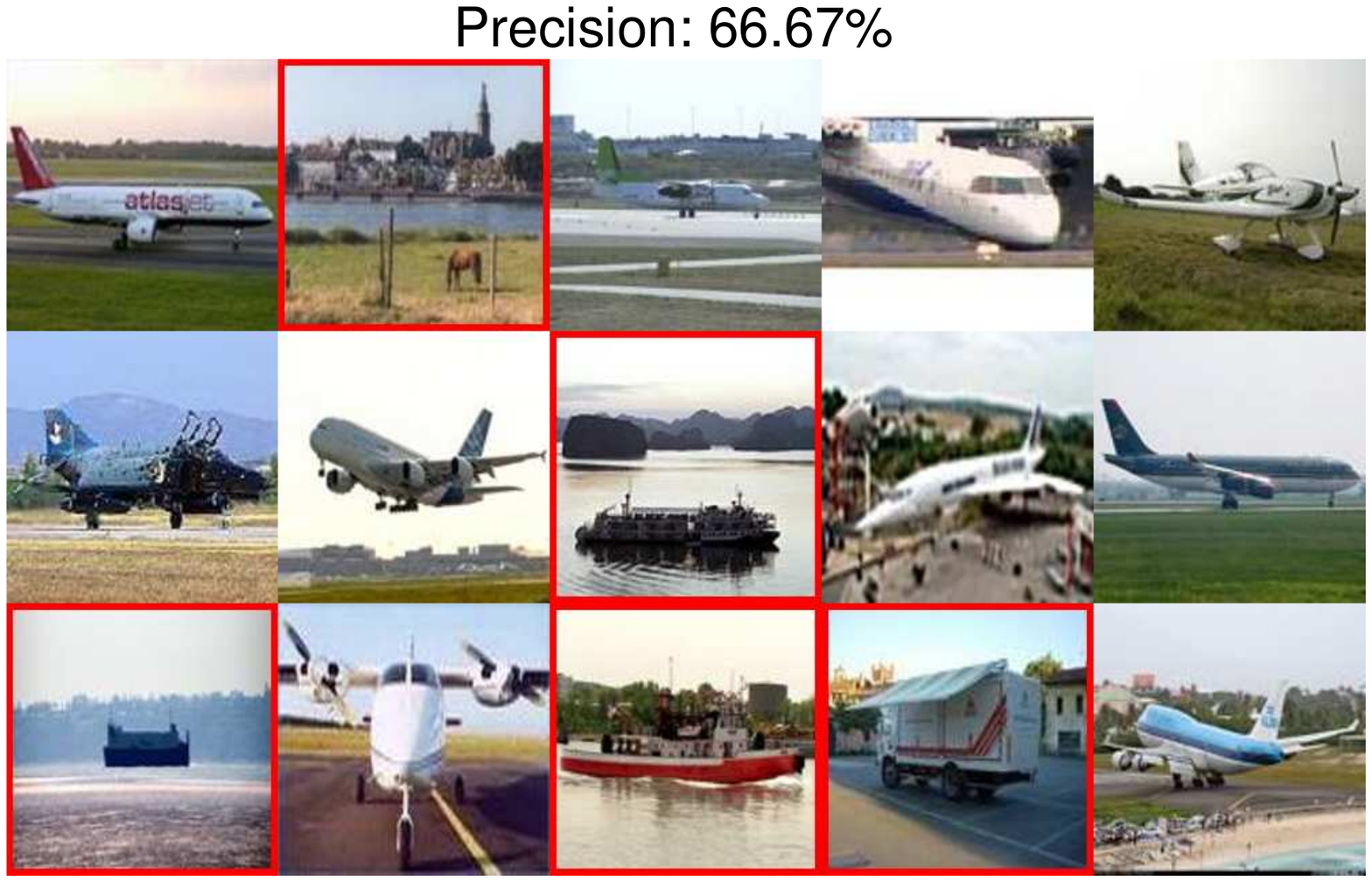}}
\subfigure[CCA (V+T).]{
   \includegraphics[width=1.9in, trim= 15mm 85mm 15mm
80mm]{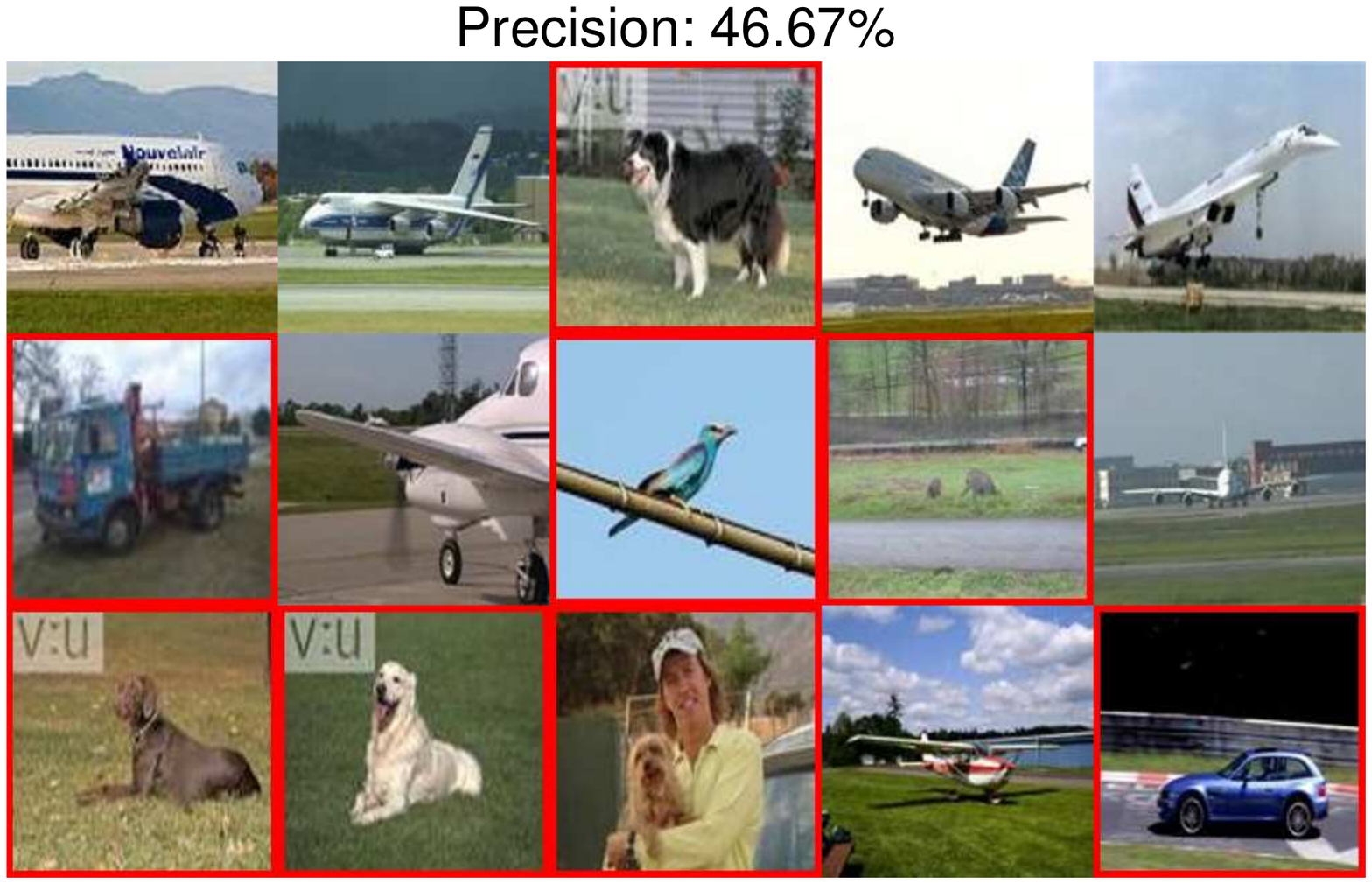}}
\subfigure[CCA (V+T+C).]{
   \includegraphics[width=1.9in, trim= 15mm 85mm 15mm
80mm]{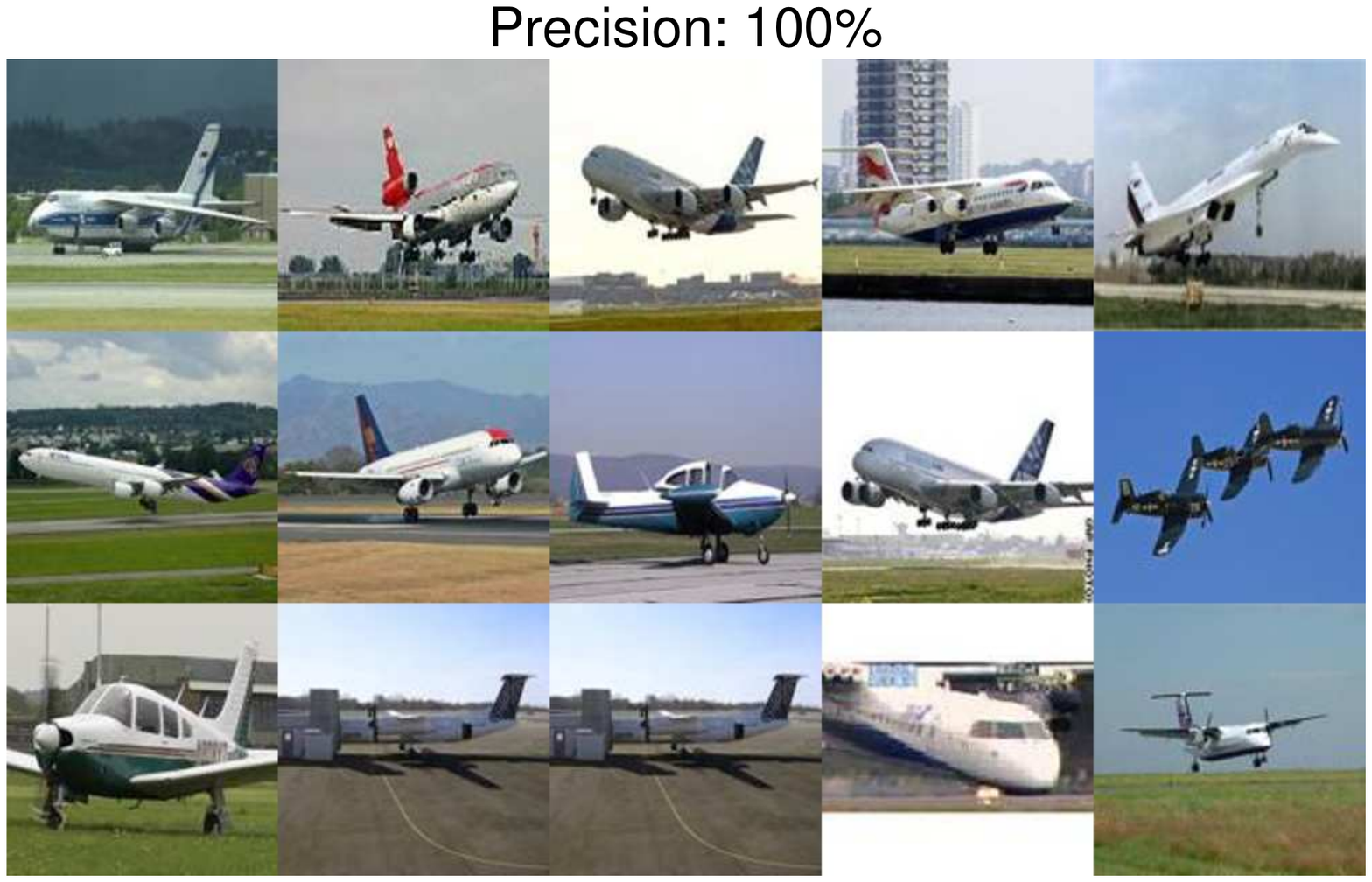}}
   \caption{Image-to-image retrieval results for two sample queries. The leftmost image is the query. Red border indicates a false positive.}
      \label{sample_i2i2}
\end{figure*}

\begin{figure*}[] 
   \centering
   \subfigure[yellow]{
   \includegraphics[width=1.5in, trim= 50mm 90mm 50mm
80mm]{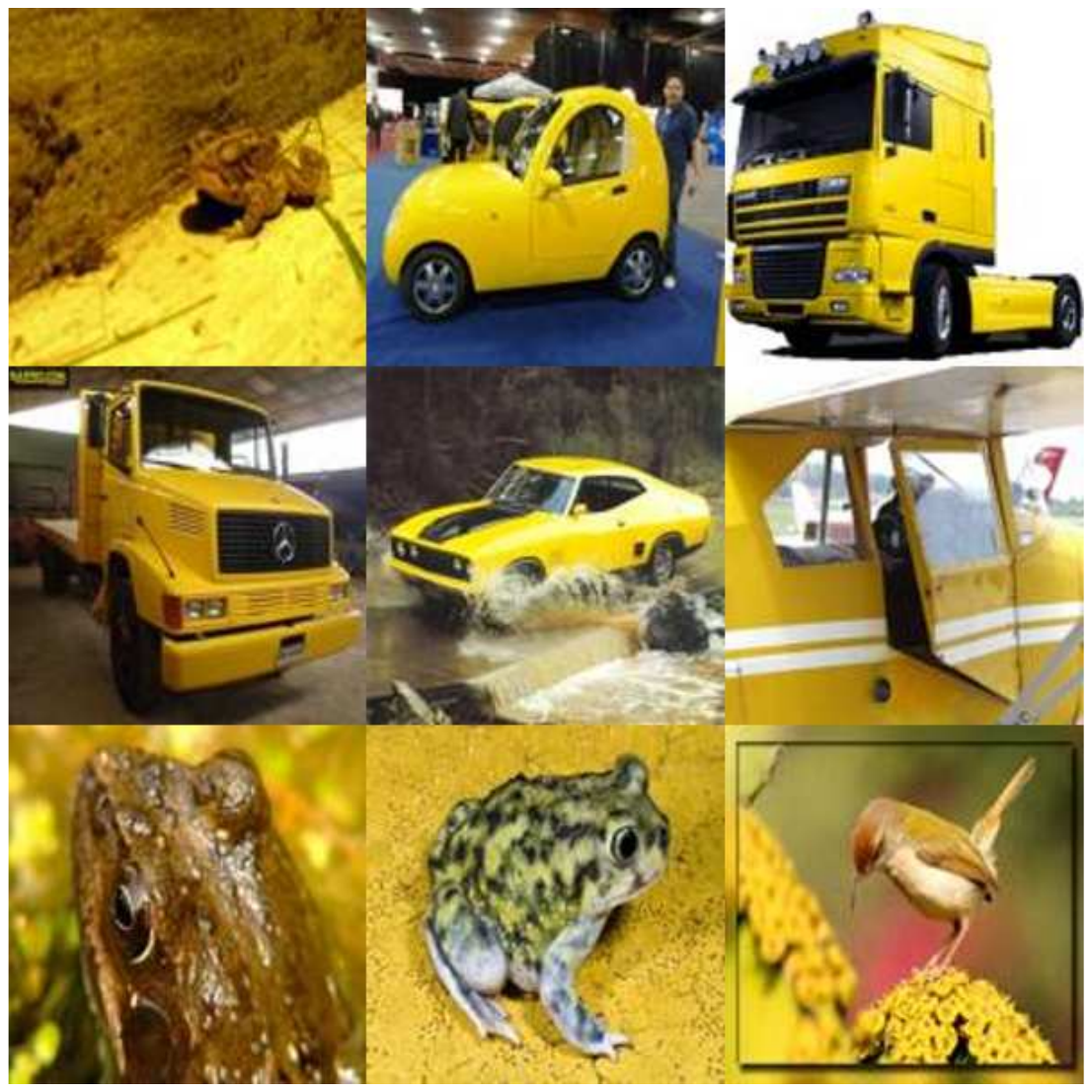}}
\subfigure[red]{
  \includegraphics[width=1.5in, trim= 50mm 90mm 50mm
80mm]{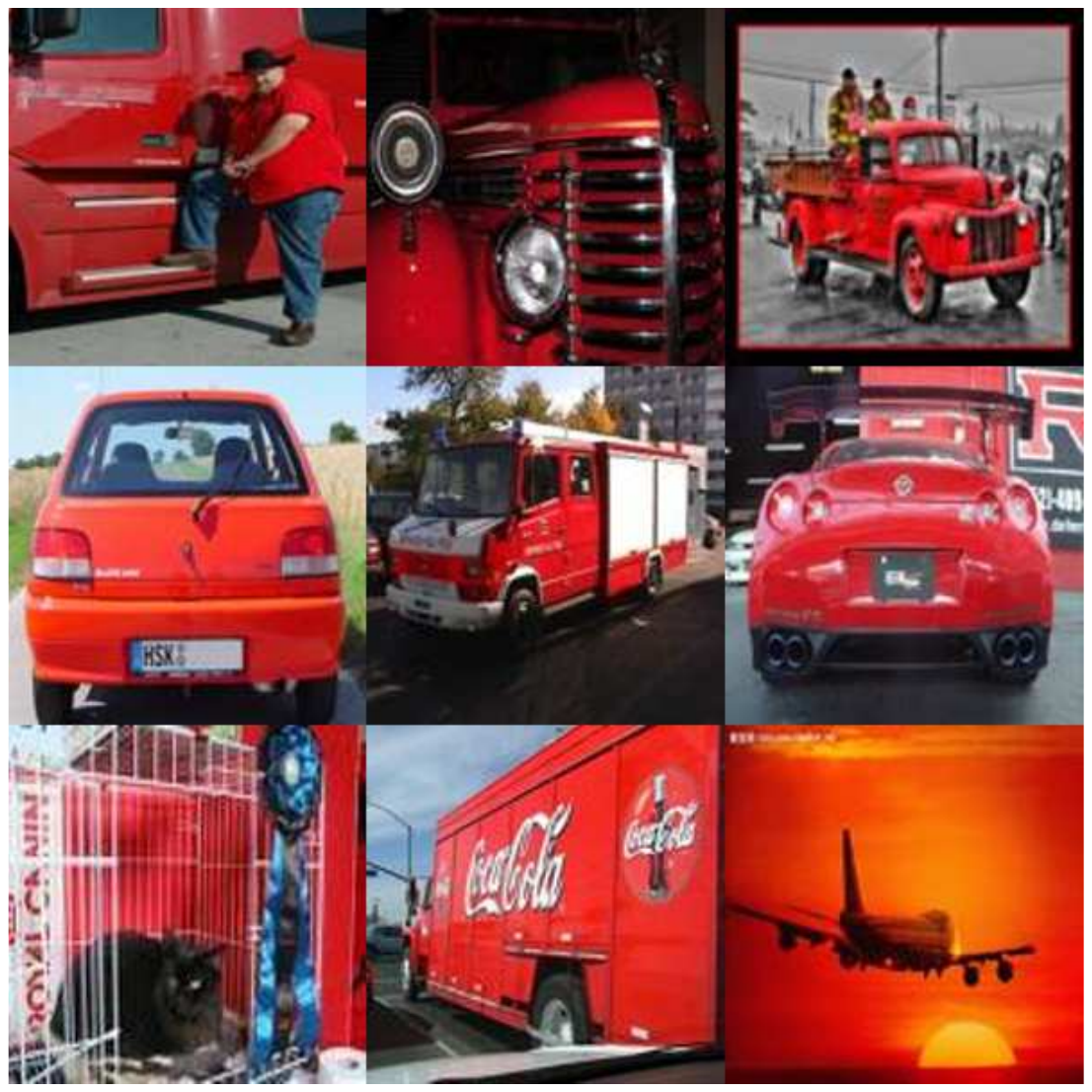}}
\subfigure[sail]{
 \includegraphics[width=1.5in, trim= 50mm 90mm 50mm
80mm]{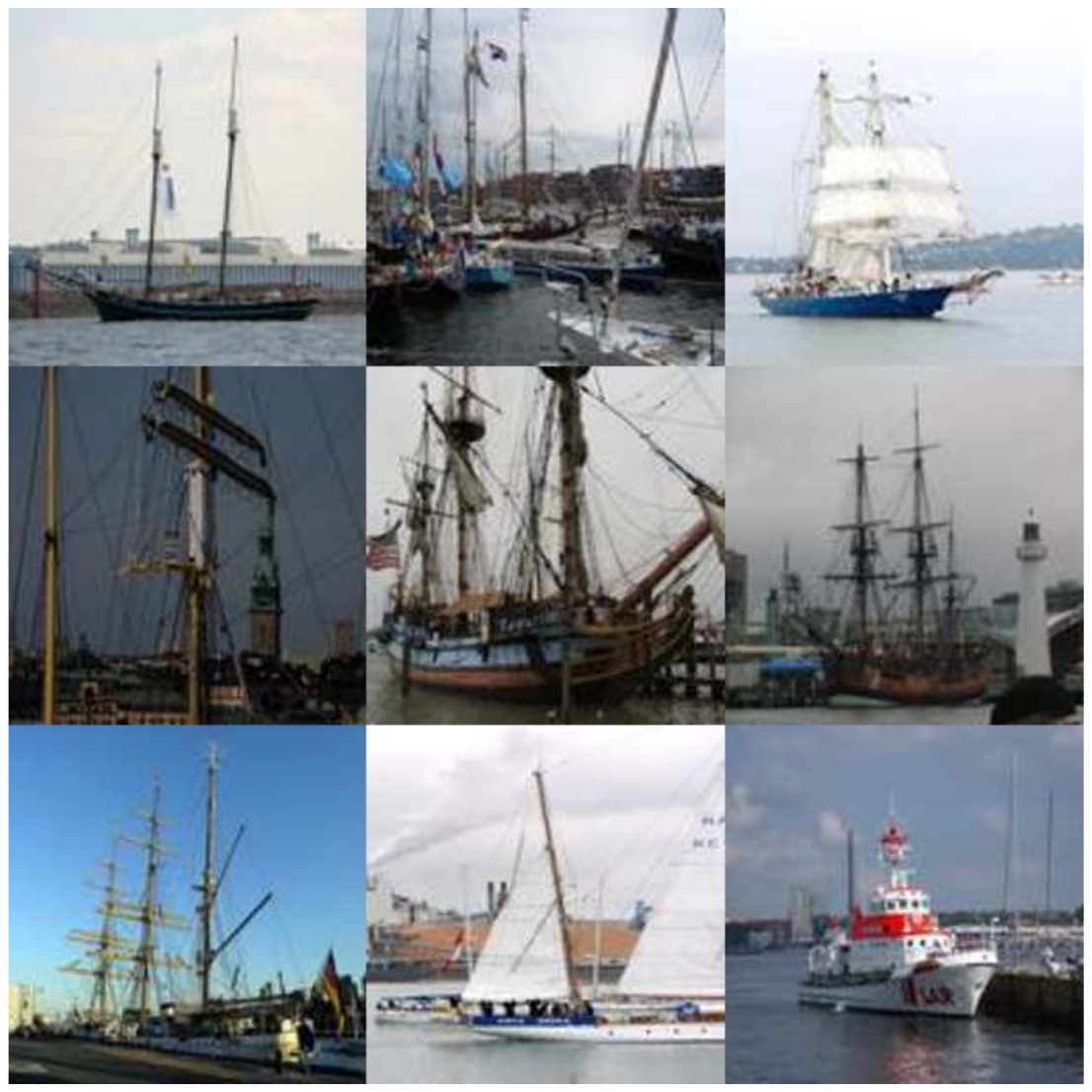}}
\subfigure[ocean]{
   \includegraphics[width=1.5in, trim= 50mm 90mm 50mm
80mm]{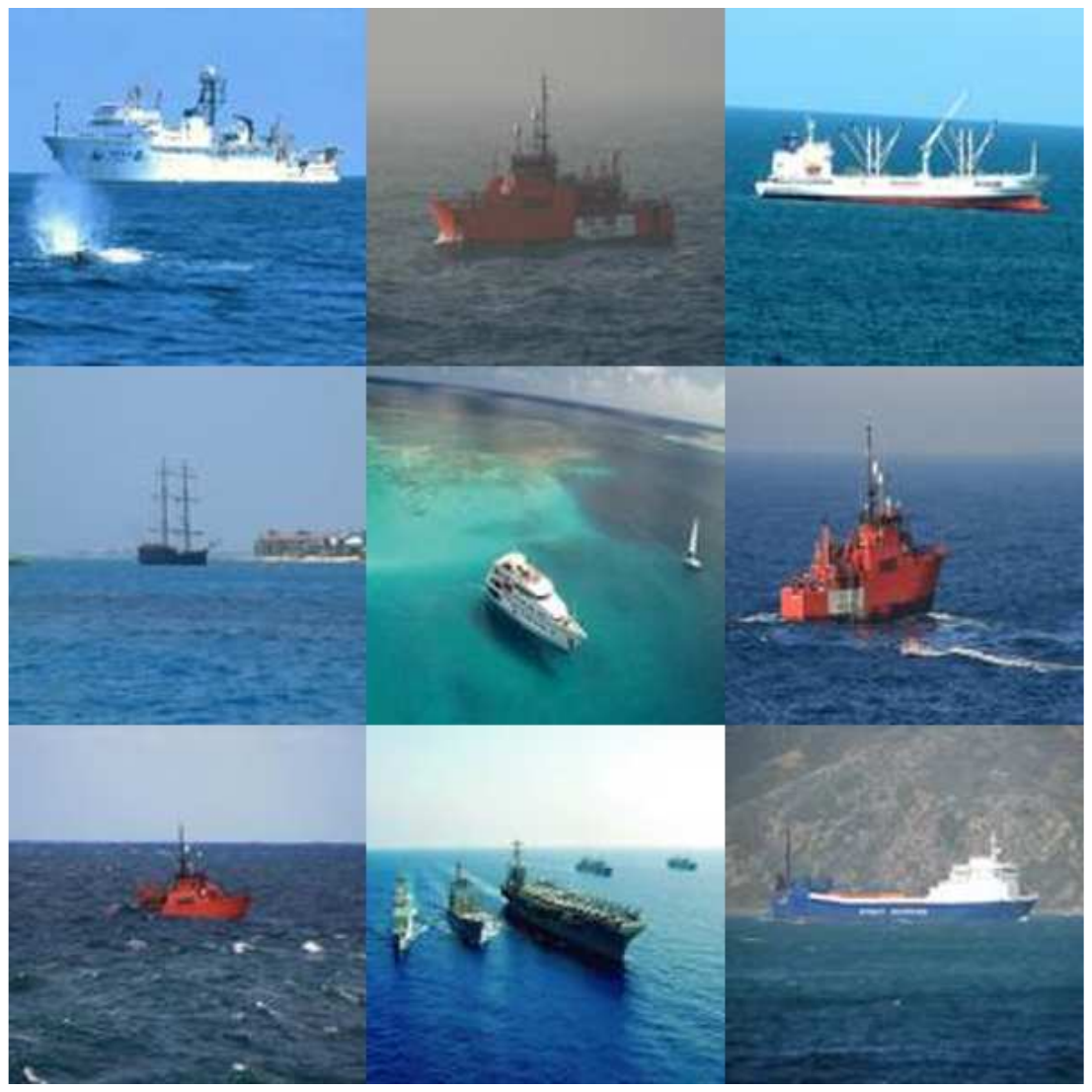}}
   \caption{Examples of tag-based image search on the Flickr-CIFAR dataset.}
   \label{key}
\end{figure*}

Next, we consider our supervised three-view model, CCA (V+T+K), where the third view is given by the search keywords used to retrieve the Flickr images. Even though this supervisory information is noisy (not all images retrieved by Flickr search for ``truck'' will actually contain trucks), we can see that incorporating it as a third view improves the precision of all three of our target retrieval tasks. The unsupervised version of our three-view model, CCA (V+T+C), where the third view is given by 20 NC clusters, performs almost identically to CCA (V+T+K) on I2I and K2I, and has slightly lower precision for T2I. This is a very encouraging result, since it shows that semantic information that is automatically recovered from noisy tags still provides a very powerful form of supervision.

Table \ref{main} also lists the performance of two-view models CCA (V+K) and CCA (V+C) given by replacing the lower-level tag-based view T by the higher-level but lower-dimensional semantic view (K or C). Compared to CCA (V+T), both models have significantly higher I2I precision (though it is a bit lower than that of the respective three-view models). Thus, replacing noisy tags with the cleaner semantic views can help to improve performance. 
However, the two-view V+K and V+C models are not suitable for tag-to-image search, while the three-view models can be used for all the tasks we care about.

The last two lines of Table \ref{main} report baseline comparisons to structural learning~\citep{Ando05,Quattoni07} and Wsabie~\citep{Weston11}. Both perform better than CCA (V+T) but worse than all our other multi-view CCA models. One reason for this is because these models were designed for discrimination, not retrieval. In the case of Wsabie, it is possible that a batch learning approach (for example, second-order batch optimization) can give better performance than first-order SGD. However, a batch implementation of Wsabie is beyond the scope of our baseline comparisons (as described in the original paper, Wsabie is a sampling method specifically designed for SGD). Furthermore, neither structural learning nor Wsabie produces an embedding for tags, so unlike CCA (V+T) and our three-view models, these baselines are not suitable for tag-to-image retrieval.


\begin{figure*}[] 
   \centering
   \subfigure[deer]{
   \includegraphics[width=1.5in, trim=50mm 85mm 50mm
85mm]{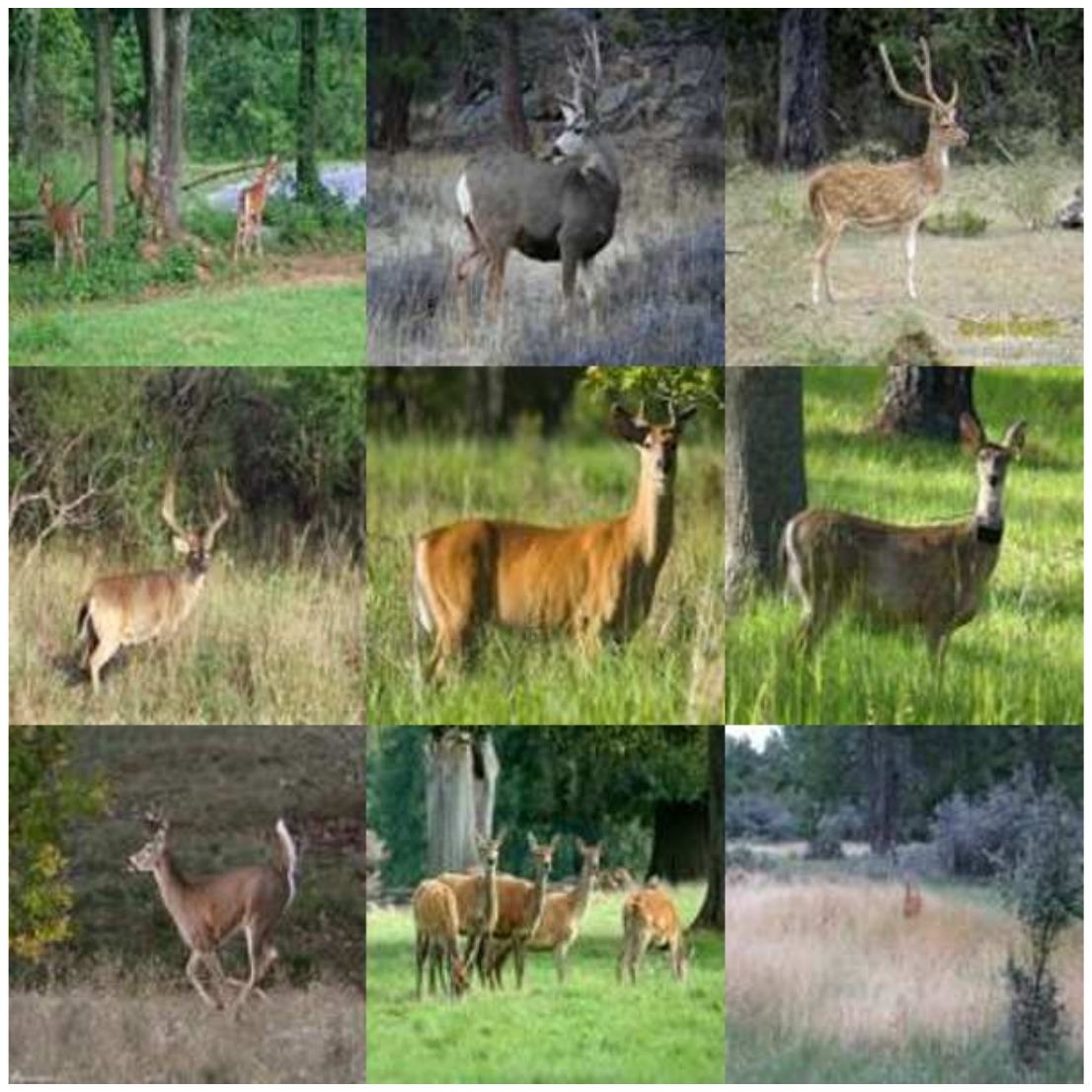}}
\subfigure[deer, snow$\times$2]{
   \includegraphics[width=1.5in, trim= 50mm 85mm 50mm
85mm]{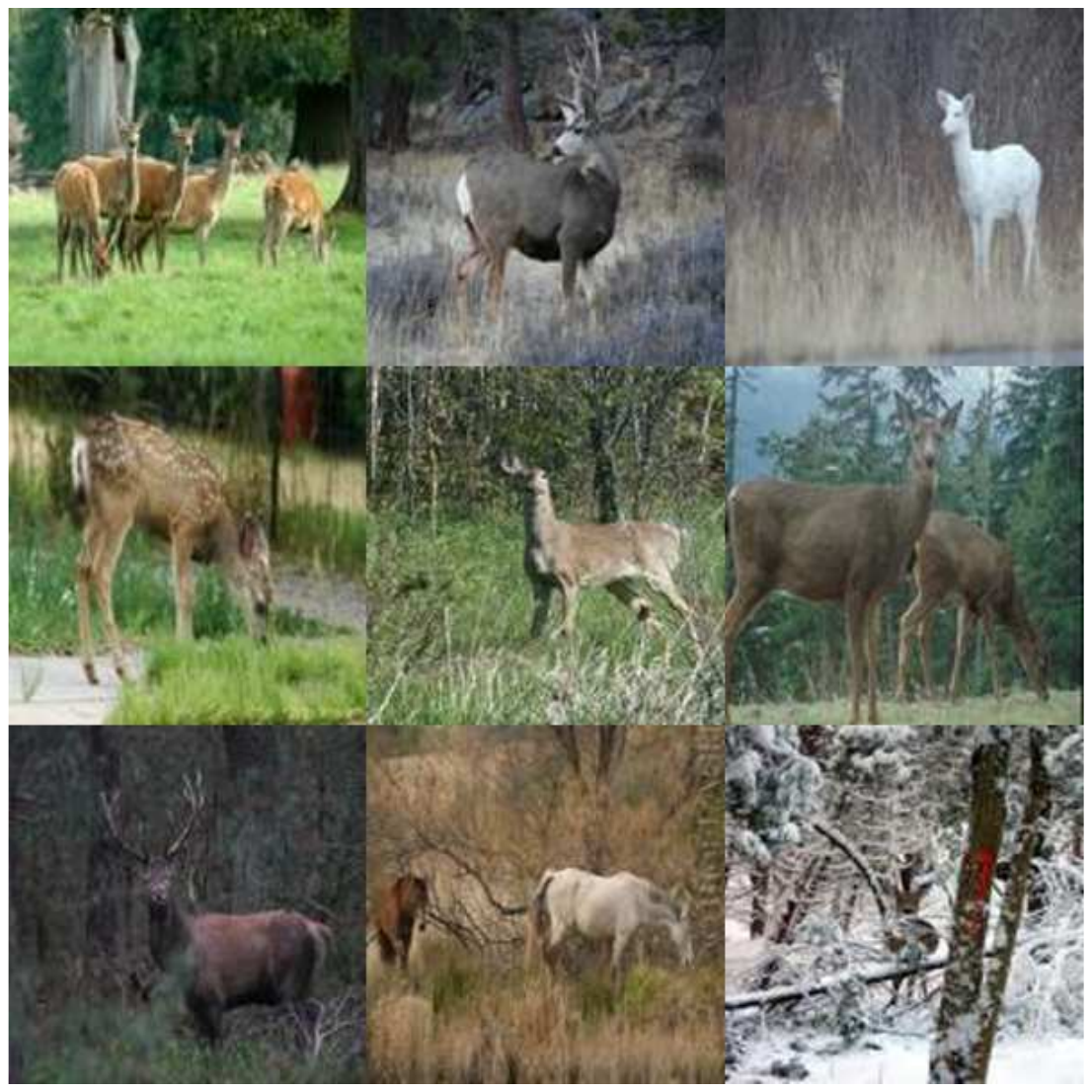}}
\subfigure[deer, snow$\times$6]{
   \includegraphics[width=1.5in, trim= 50mm 85mm 50mm
85mm]{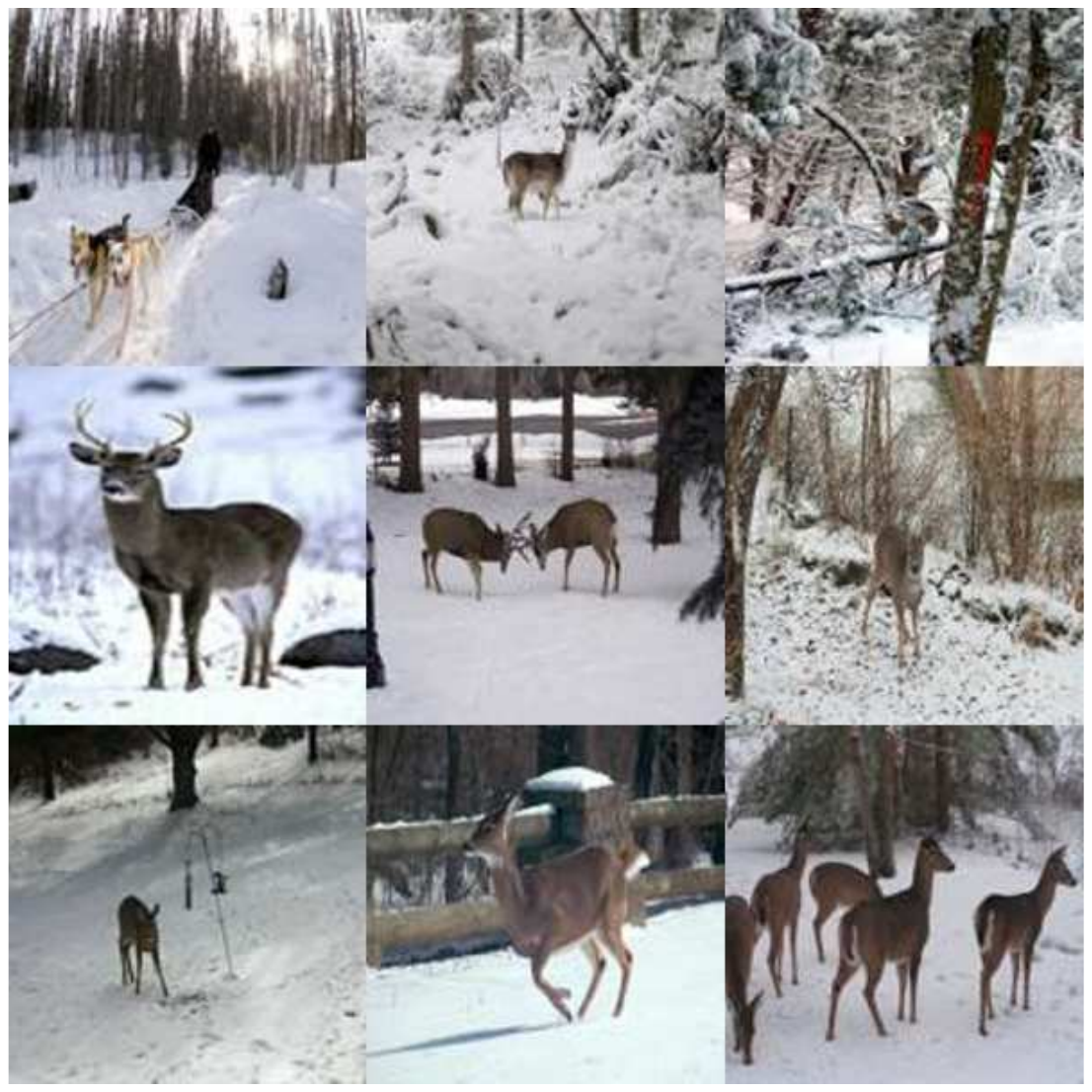}}
\subfigure[snow]{
   \includegraphics[width=1.5in, trim= 50mm 85mm 50mm
85mm]{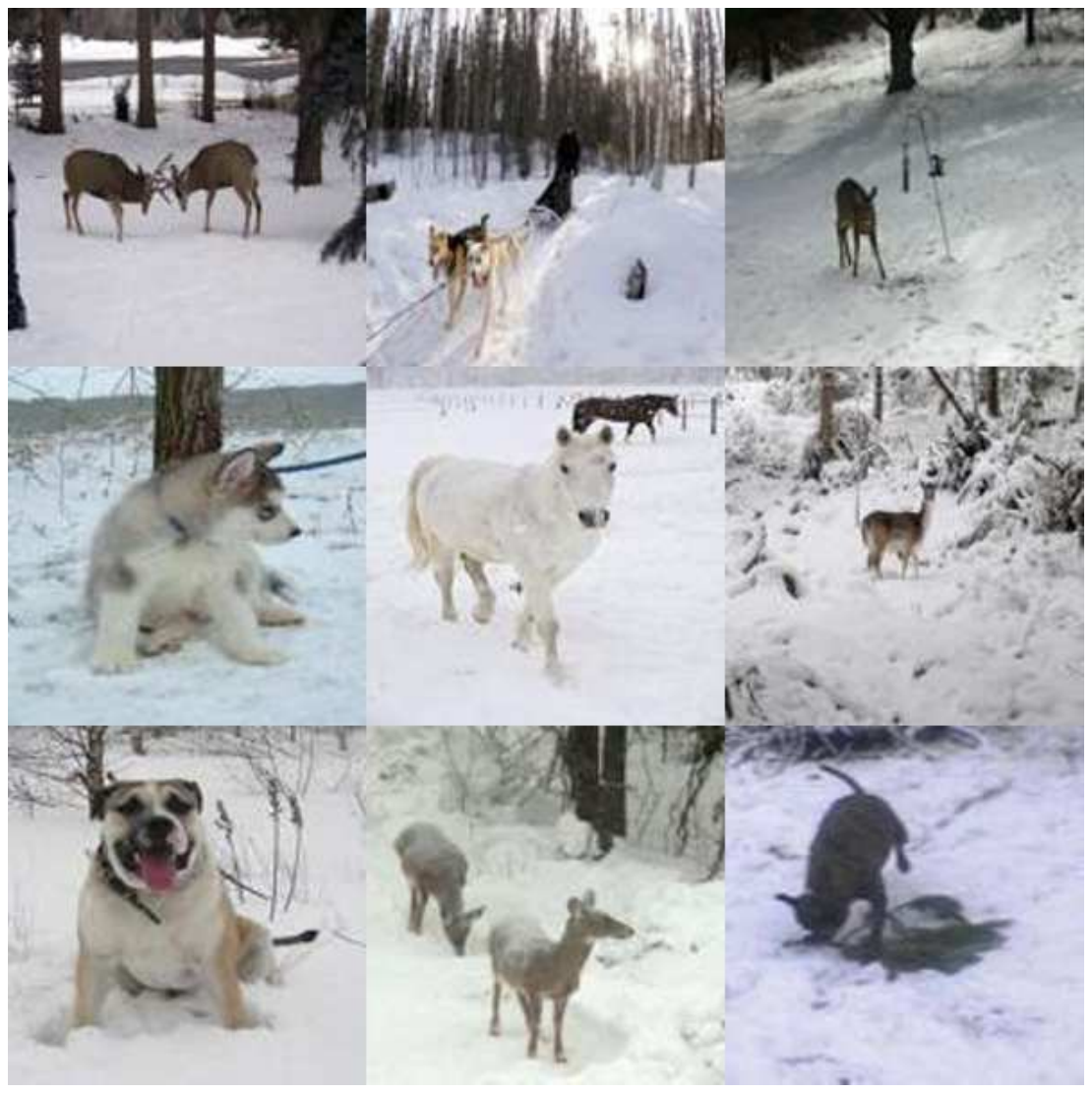}}
\vspace{2ex}
   \subfigure[ship]{
   \includegraphics[width=1.5in, trim= 50mm 85mm 50mm
85mm]{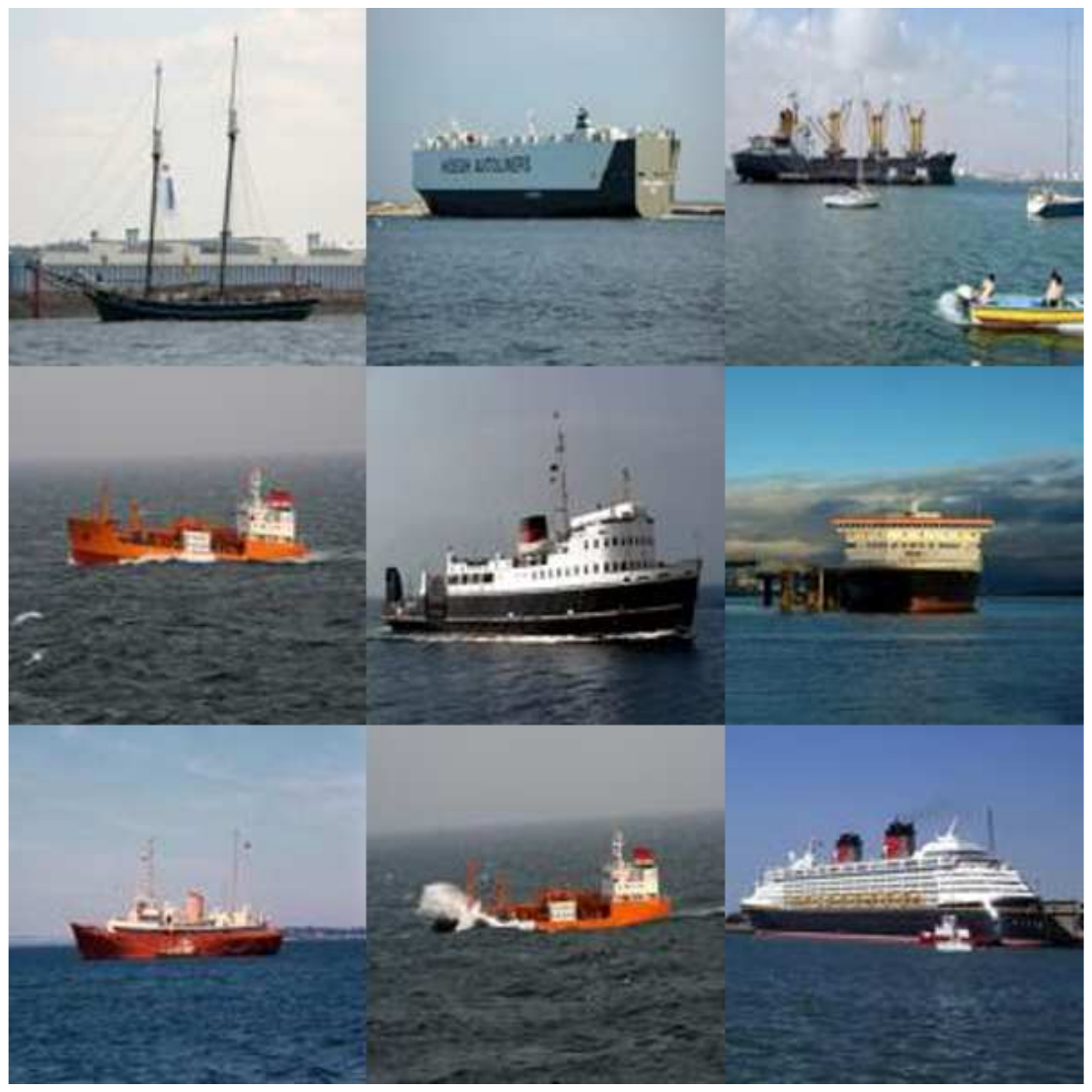}}
\subfigure[ship, sunset$\times$2]{
   \includegraphics[width=1.5in, trim= 50mm 85mm 50mm
85mm]{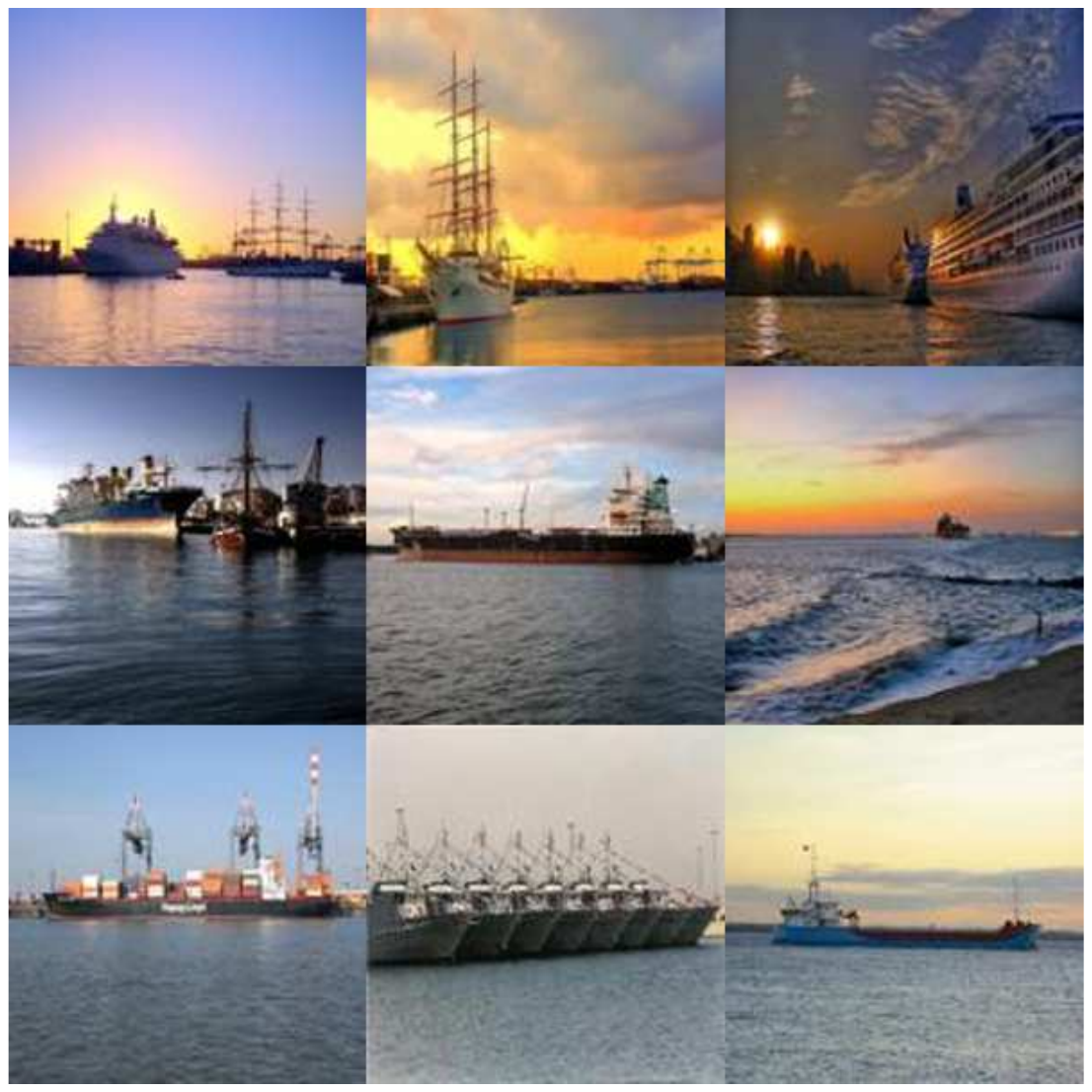}}
\subfigure[ship, sunset$\times$6]{
   \includegraphics[width=1.5in, trim= 50mm 85mm 50mm
85mm]{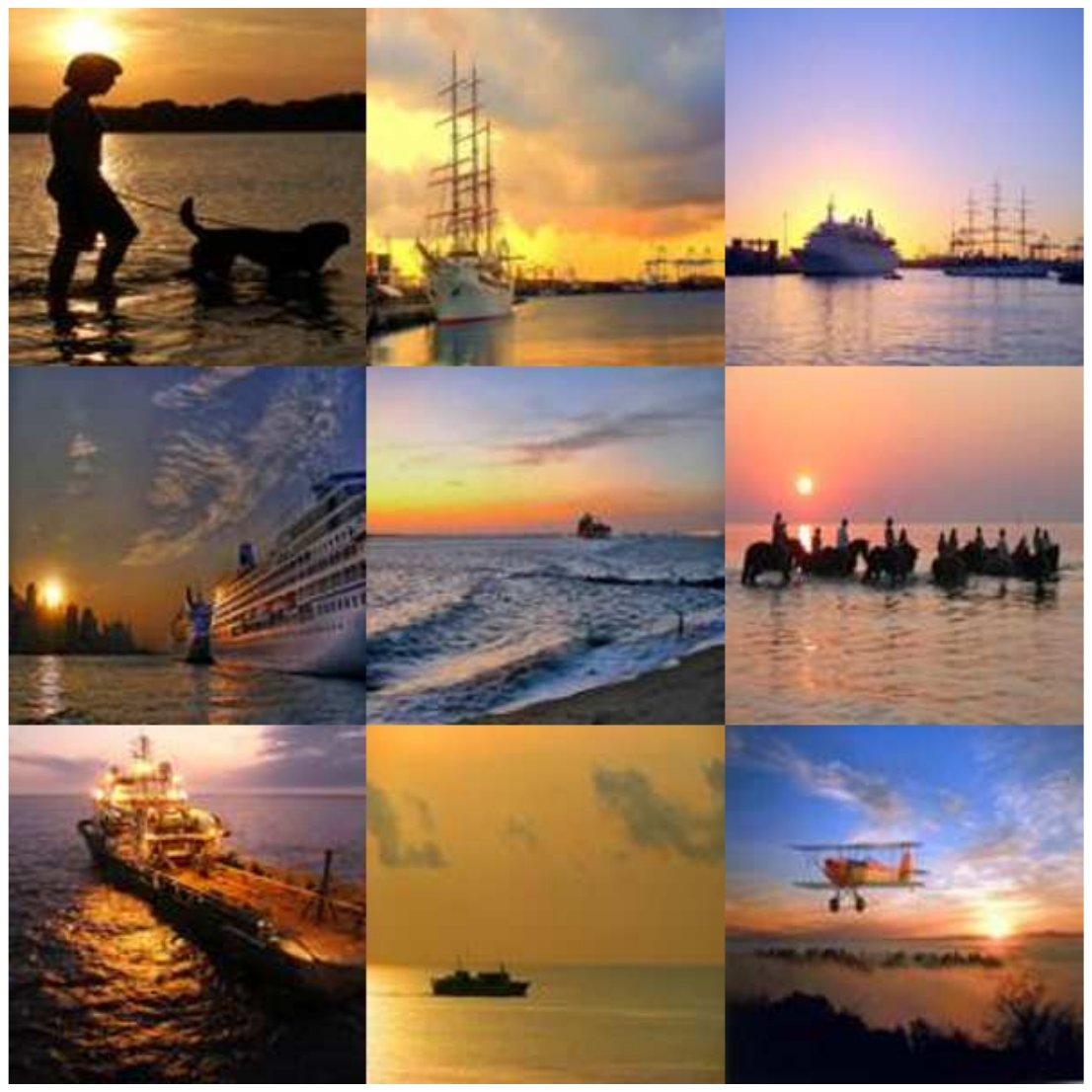}}
\subfigure[sunset]{
   \includegraphics[width=1.5in, trim= 50mm 85mm 50mm
85mm]{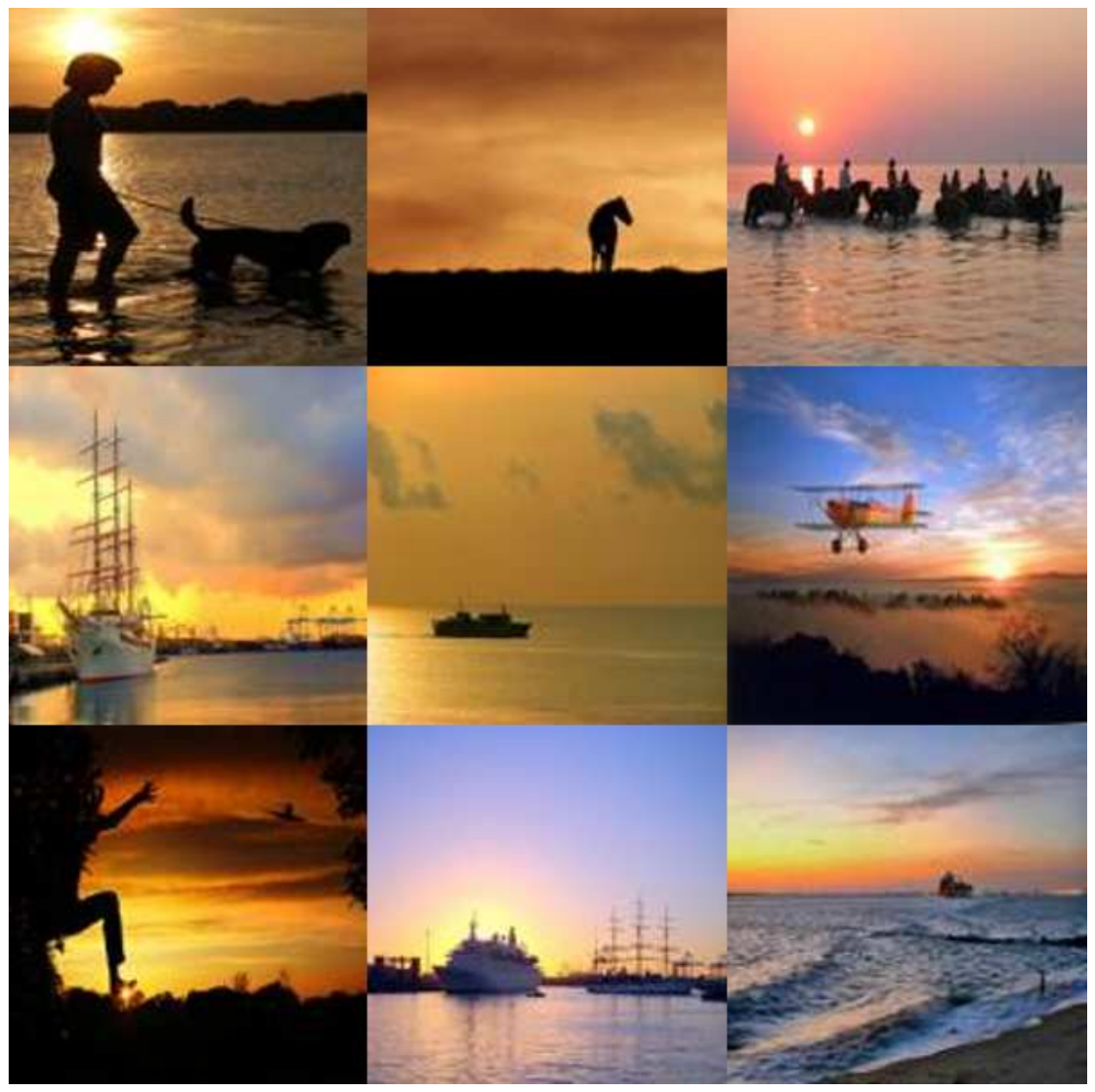}}
   \caption{Examples of tag-to-image search on Flickr-CIFAR with multiple query tags and adjustable weights (see text).}
   \label{weight}
\end{figure*}

\subsection{Qualitative results}

Figure \ref{sample_i2i2} shows image-to-image search results for two example queries, and Figures \ref{key} and \ref{weight} show examples of tag-to-image search results. As noted earlier, one advantage of our system over traditional tag-based search approaches is that once our multi-view embedding is learned, we can use it to perform tag-to-image search on databases of images \emph{without any accompanying text}. In fact, for the Flickr-CIFAR dataset, recall that we are using an embedding trained on tagged Flickr images to embed and search ImageNet images that lack tags. Figure \ref{key} shows top retrieved images for four tags that do not correspond to the main ten keywords that were used to download the dataset. In particular, we are able to learn colors, common background classes like ``ocean,'' and sub-classes of the main keywords like ``sail.''

Figure \ref{weight} shows images retrieved for more complex queries consisting of multiple tags such as ``deer, snow.'' Note that ``deer'' is one of our ten main keywords, and ``snow'' is a much less common tag. To get good retrieval results in such cases, we have found that we need to give higher weights to the more minor concepts when forming the query tag vector. Intuitively, the tag projection matrix found by minimizing the CCA objective function (eq. \ref{eq:cca}) is much more influenced by the distortion due to the common tags rather than the rare ones. We have empirically observed that we can counteract this effect and obtain more accurate results for less frequent tags by increasing their weights in the tag vector at query time. To date, we have not designed a way to tune the weights automatically. However, in an interactive image search system, it would be very natural for users to adjust the weights on the fly to modulate the importance of different concepts in their query. For example, in Figure \ref{weight} (a)-(c), when we increase the weight for ``snow,'' snow becomes more and more prominent in the retrieved images.

\begin{figure}[] 
   \centering
   \includegraphics[width=0.8\columnwidth, trim=10mm 70mm 10mm
80mm]{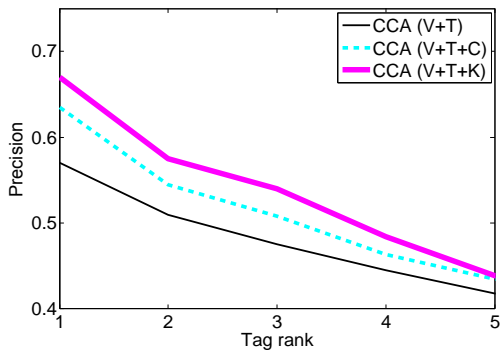}
   \caption{Tagging results on the Flickr-CIFAR dataset: Average precision of retrieved tags vs. tag rank based on manual evaluation (see text).}
   \label{taggingcifar}
\end{figure}

\begin{figure*}[] 
   \centering
   \includegraphics[width=0.7\textwidth, trim=25mm 145mm 25mm 30mm]{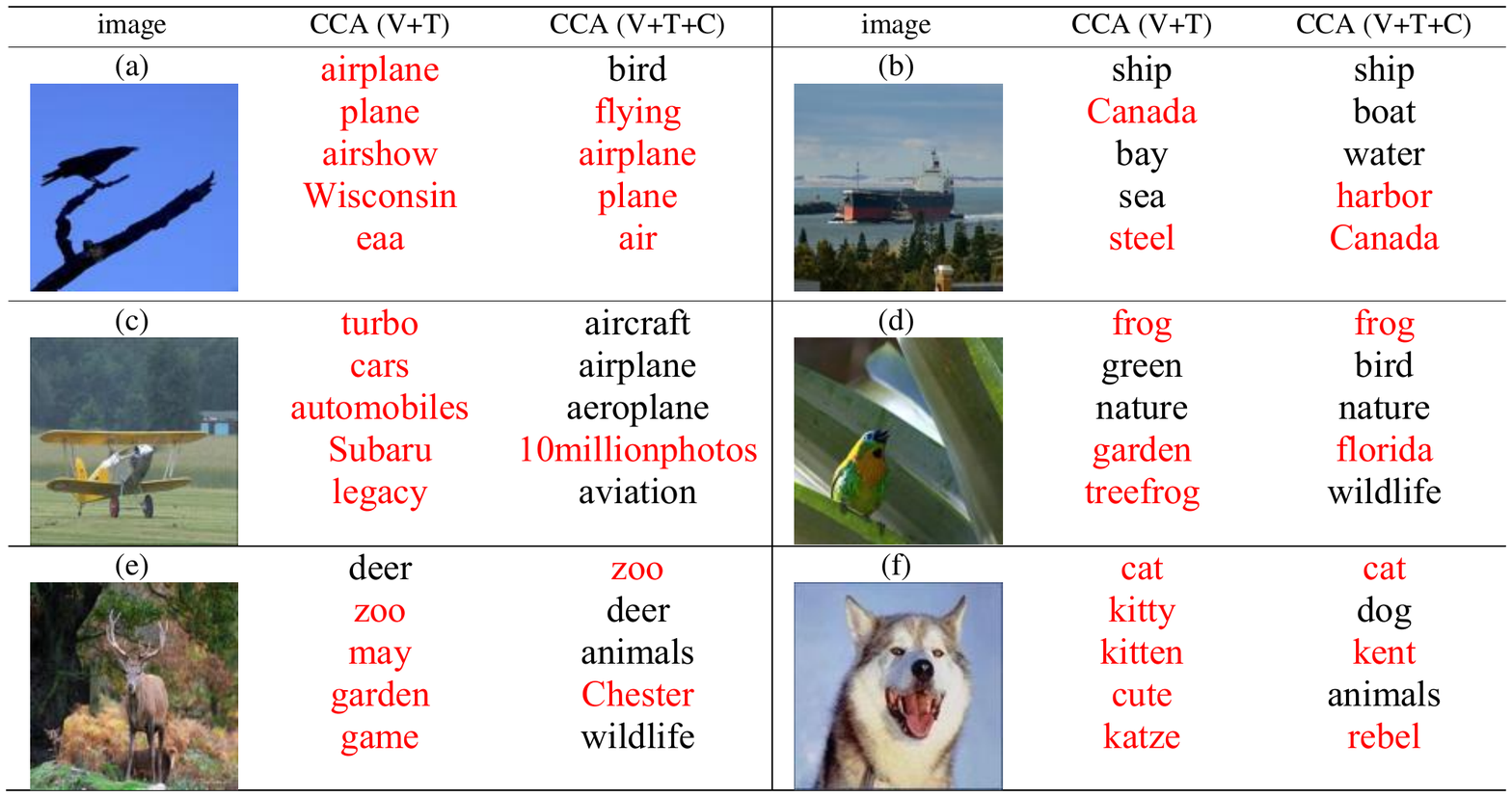}
   \caption{Example image tagging results on Flickr-CIFAR dataset for CCA (V+T) and CCA (V+T+C). Tags in red have been marked as irrelevant by human annotators.}
   \label{tagging}
\end{figure*}

\subsection{Tagging results} \label{sec:tagging}

This section presents a quantitative evaluation of our method for image tagging or annotation. As described in Section \ref{dataset}, we use the data-driven annotation scheme of \cite{Makadia08}, where tags are transferred from top fifty neighbors to the query in the latent space. We randomly sample 200 query images from our ImageNet test set and use CCA (V+T), CCA (V+T+C), and CCA (V+T+K) spaces to transfer tags. To evaluate the results, we ask four individuals (members of the research group not directly involved with this project) to verify the tags suggested by different methods, that is, mark each tag as either relevant or irrelevant to the image. To avoid bias, our evaluation interface does not tell the evaluators which set of tags was produced by which method, and presents the sets of tags corresponding to different methods in random order for each test image. Our reasons for using human evaluation are twofold: first, our test images do not have any ground truth annotations; second, it is hard to provide ground truth consisting of a complete set of tags that could be relevant to an image. We combine the results of the human evaluators by voting: each tag that gets marked as relevant by three or more evaluators is considered correct.

Figure \ref{taggingcifar} reports average precision as a function of tag rank (which is determined by frequency of the tag in the top fifty closest images to the query in the CCA space). We can find that our proposed three-view models, CCA (V+T+K) and CCA (V+T+C), lead to better accuracy than the baseline CCA (V+T) method.
Figure \ref{tagging} shows the tagging results for CCA (V+T) vs. CCA (V+T+C) on a few example test images.

\section{Results on the NUS-WIDE Dataset\label{sec:evaluation2}}

In this section, we compare different multi-view embeddings on the NUS-WIDE dataset. We randomly split the dataset into 219,648 training and 50,000 test images. As before, we learn the joint embedding using the training images, and test retrieval accuracy on testing dataset. In the test set, we randomly sample and fix 1,000 images as the queries, 1,000 images as the validation set, and retrieve the remaining images. The validation set is used to find the number of clusters for NC. For this dataset, this number ends up being 100, vs. 20 for Flickr-CIFAR. The larger number of clusters for NUS-WIDE is not surprising, since this dataset has a much larger number of underlying semantic concepts than Flickr-CIFAR (81 vs. ten). Since there are relatively fewer images per class, we report Precision@20 instead of Precision@50. Also, since the images in this dataset may contain multiple ground truth keywords, we compute {\em average per-keyword} precision. That is, if $q$ is the number of keywords for a given query image and $a$ is the number of relevant keywords retrieved in the top $p$ images, we define Precision@$p$ as $\frac{a}{pq}$.

\begin{table}[]
{
\hfill{}
\begin{tabular}{l||ccc}
\hline
method   &    I2I        &   T2I      	&	K2I \\
\hline
V-full   &       25.25	       &       --   	&	--      \\
V   &          	32.23         &       --       &	-- \\
\hline
 CCA (V+T) &      42.44  	       &    42.37   &	 60.87 	        \\
 CCA (V+K)   &	 	48.53	  	      &   -- 		& 74.39   \\
CCA (V+T+K)   &	 	{48.06}  	 	     &    {50.46}   	&	 {68.25} 	  	       	\\
CCA (V+C)  &     	41.72  	          &   --    &	--  \\
CCA (V+T+C)  &    	 {44.03}	        &  	  {43.11}   	&	  {64.02} 	         \\
\hline
Structural learning   &    41.21   	&	--		         &       --          \\
Wsabie & 	43.65 		 & 	 --     &    --  \\
\hline
\end{tabular}}
\hfill{}
\caption{Comparison of multi-view models and baselines on the NUS-WIDE dataset. For K2I, since images may have multiple ground truth keywords, we do not generate the keyword queries directly but use the keyword vectors of the 1,000 query images used for I2I. The performance metric is Precision@20 averaged over the number of keywords per query, as described in the text. {\bf Structural learning} refers to the method of~\cite{Ando05,Quattoni07} and {\bf Wsabie} refers to the method of~\cite{Weston11}.  We have obtained standard deviations from five random database/query splits, and they are around 0.66\% - 1.06\%.
}
\label{nus1}
\end{table}

\begin{figure*}[] 
   \centering
   \centering
\includegraphics[width=0.9in, trim=73mm 105mm 74mm
85mm]{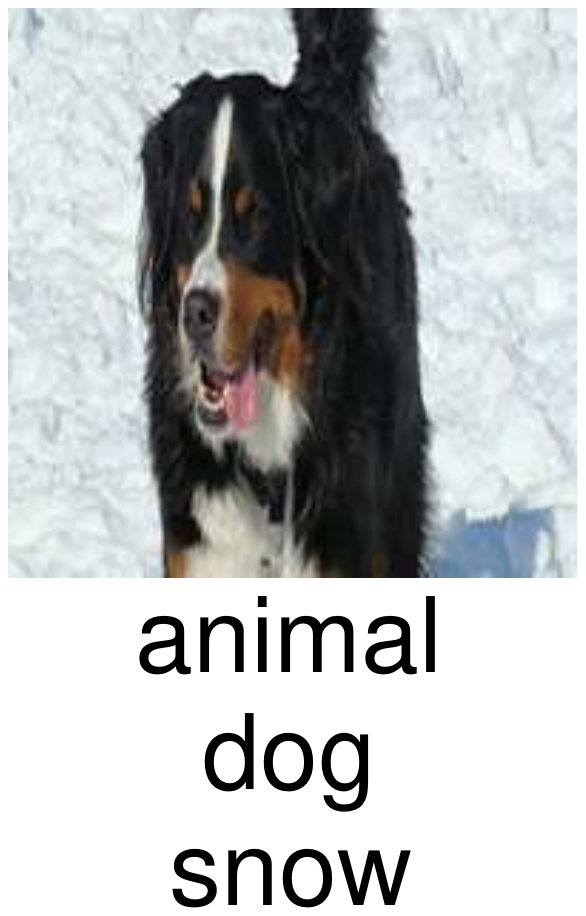}
  \includegraphics[width=1.95in, trim= 22mm 90mm 7mm
80mm]{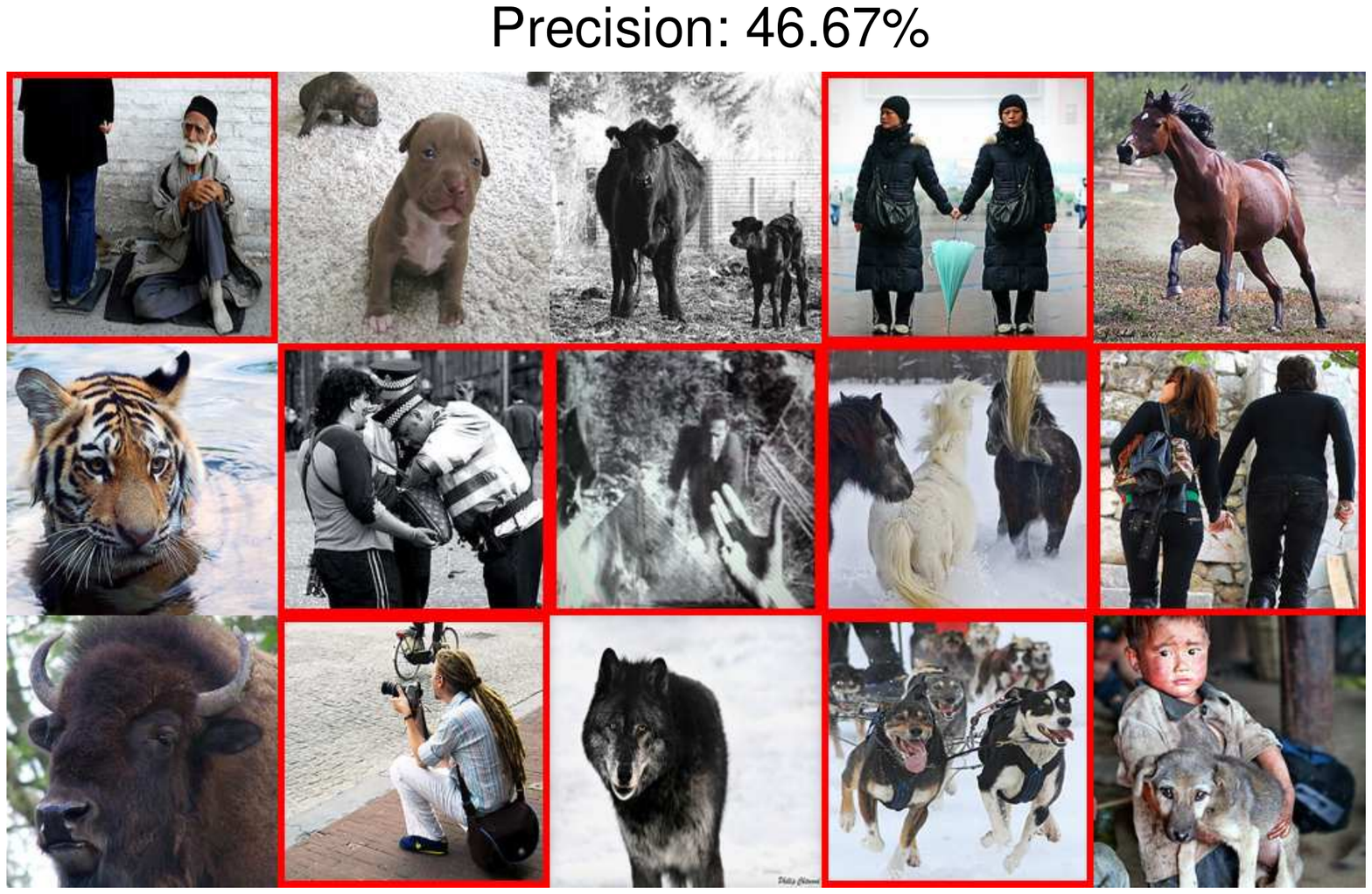}
   \includegraphics[width=1.95in, trim= 22mm 90mm 7mm
80mm]{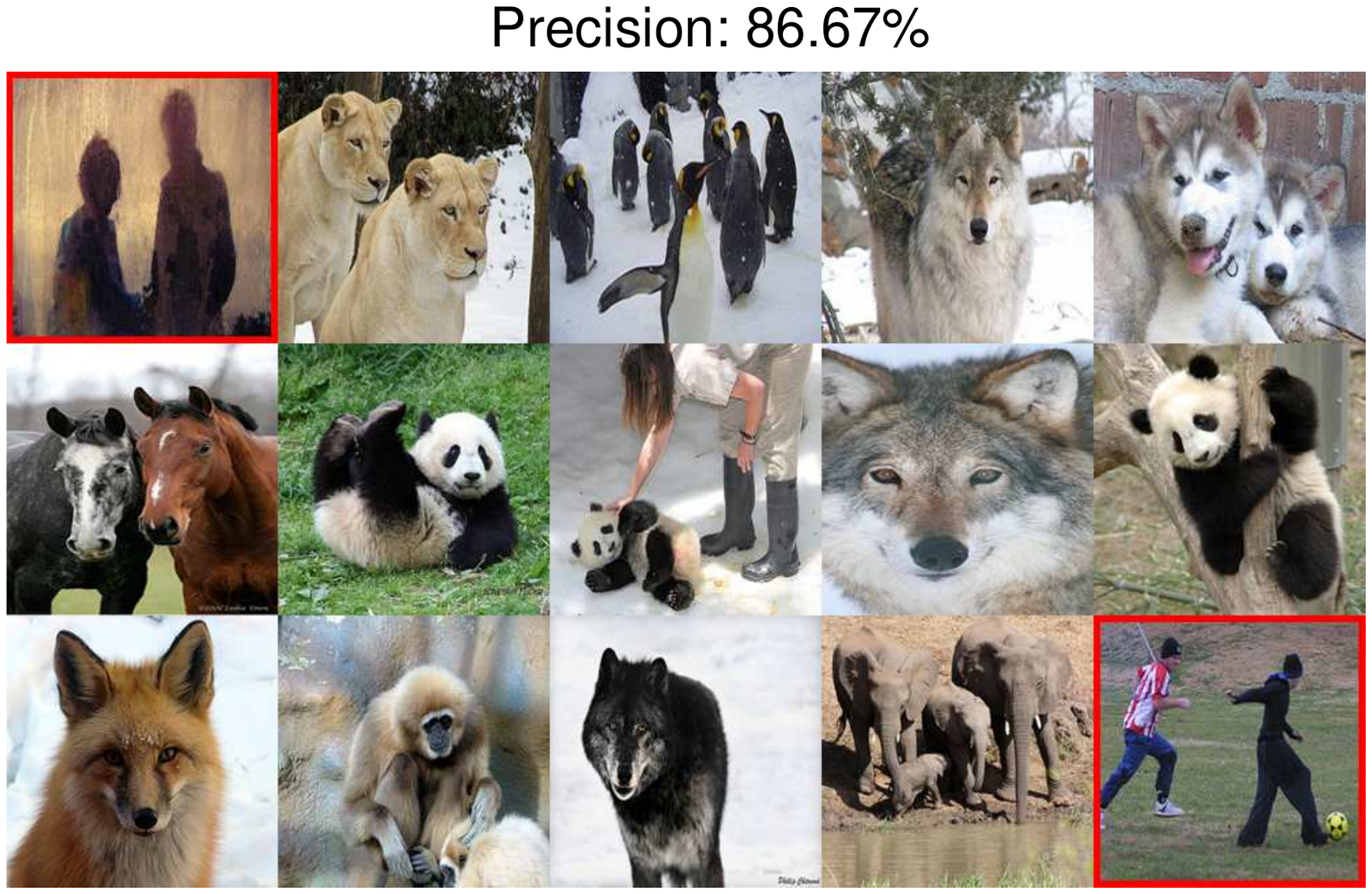}
   \includegraphics[width=1.95in, trim= 22mm 90mm 7mm
80mm]{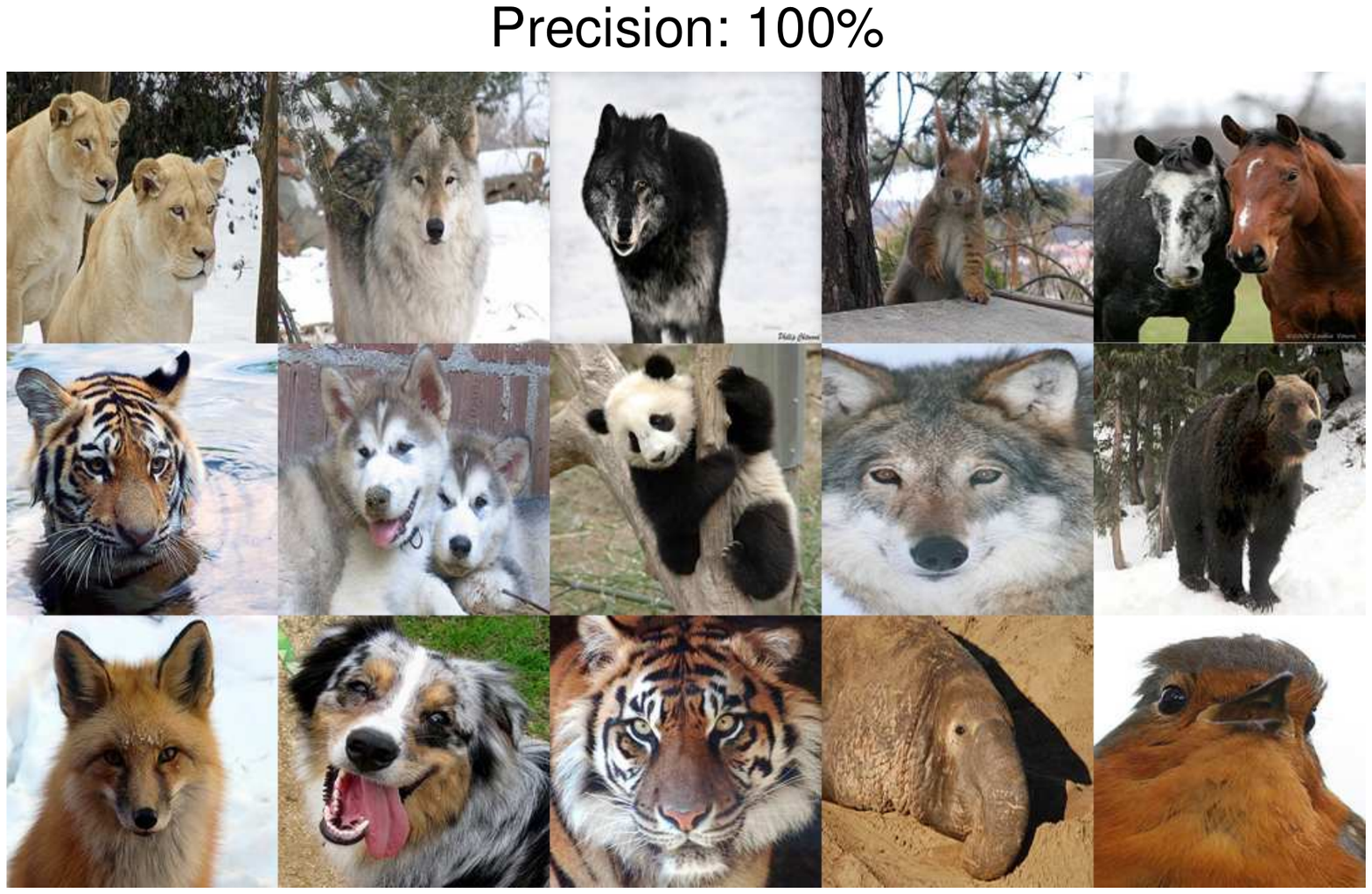}
\includegraphics[width=0.9in, trim=73mm 105mm 74mm
65mm]{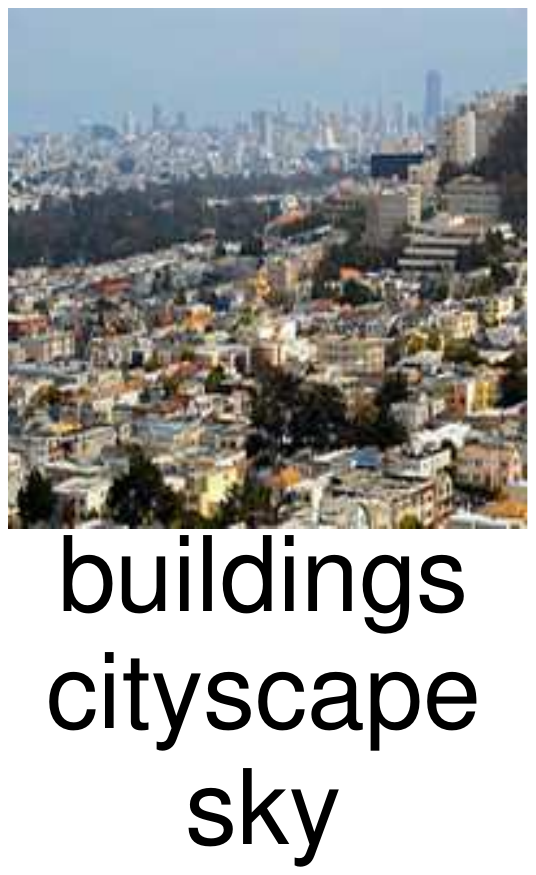}
  \includegraphics[width=1.95in, trim= 28mm 90mm 15mm
80mm]{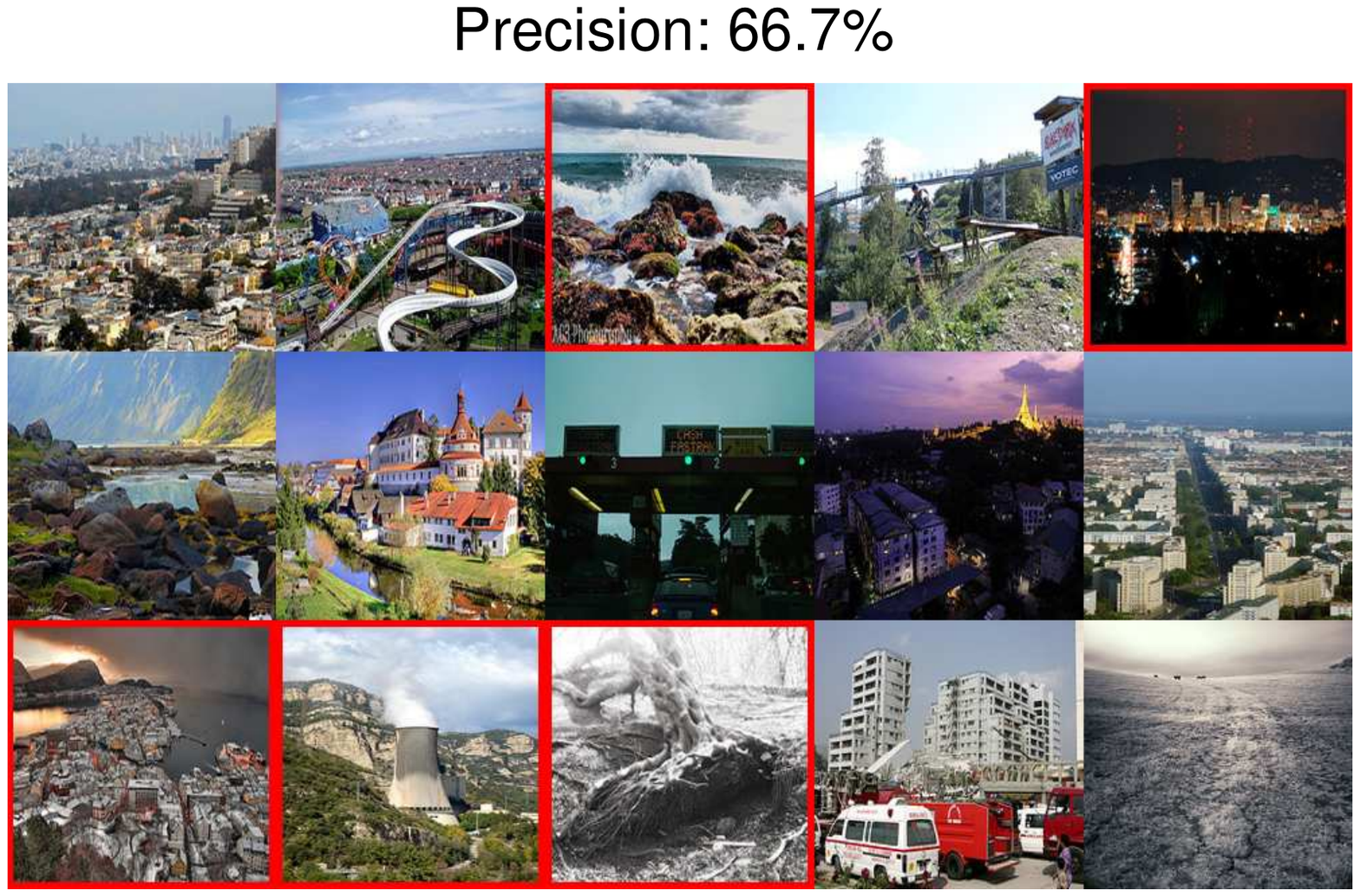}
   \includegraphics[width=1.95in, trim= 28mm 90mm 15mm
80mm]{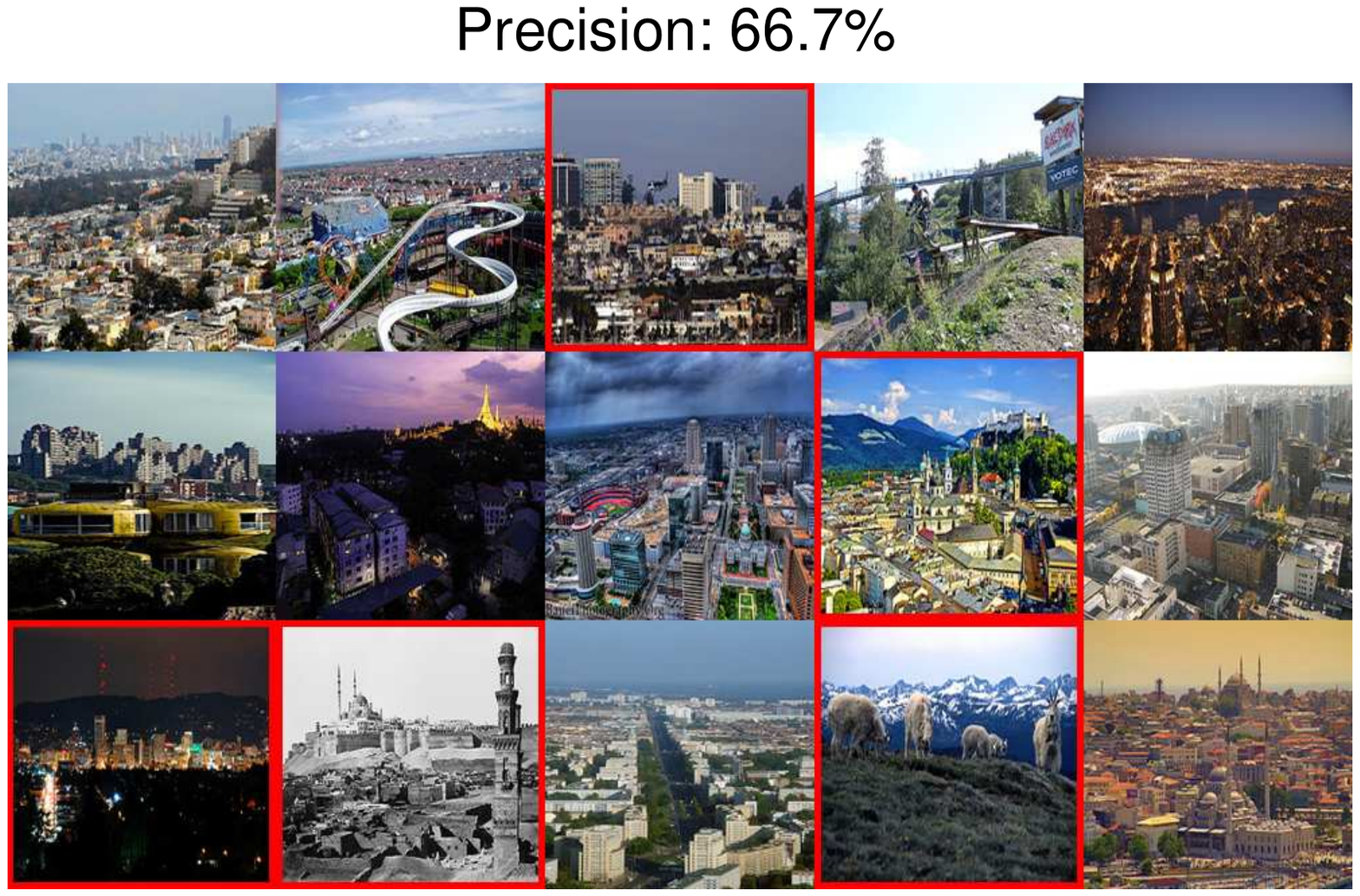}
   \includegraphics[width=1.95in, trim= 28mm 90mm 15mm
80mm]{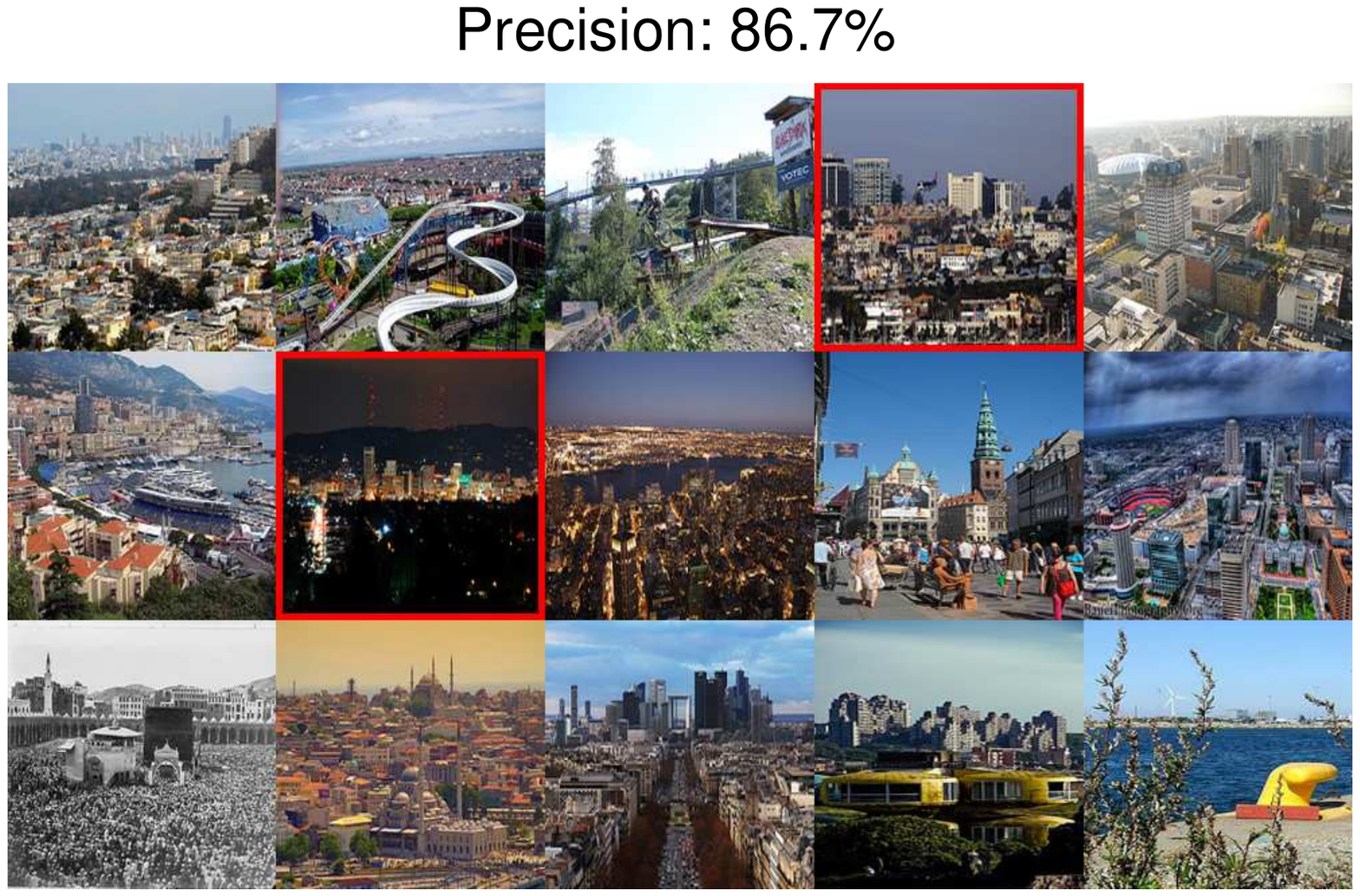}
   \includegraphics[width=0.7in, trim= 80mm 95mm 80mm
78mm]{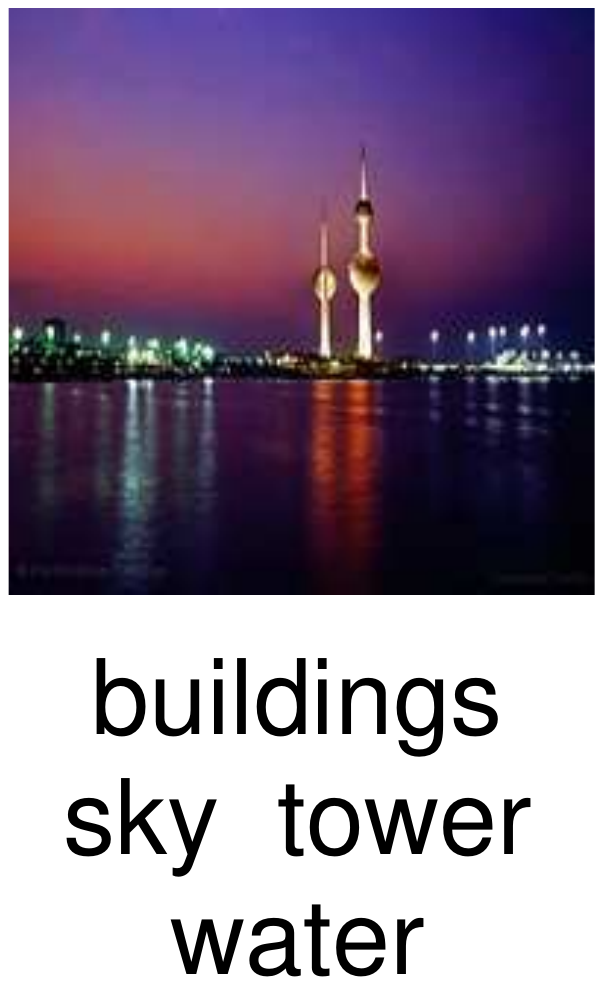}
\subfigure[Original visual feature.]{
   \includegraphics[width=1.9in, trim= 15mm 90mm 15mm
85mm]{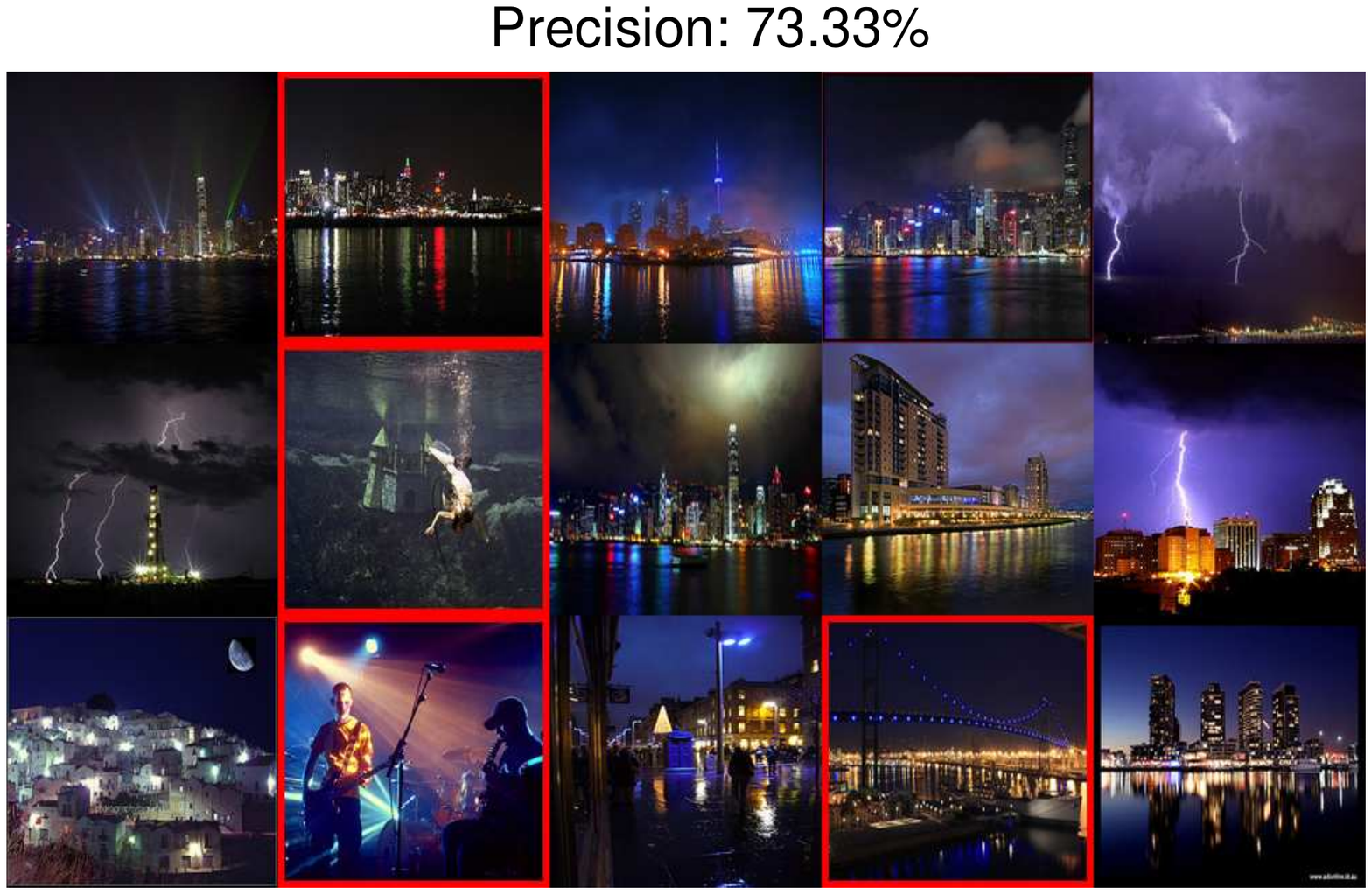}}
\subfigure[CCA (V+T).]{
   \includegraphics[width=1.9in, trim= 15mm 90mm 15mm
85mm]{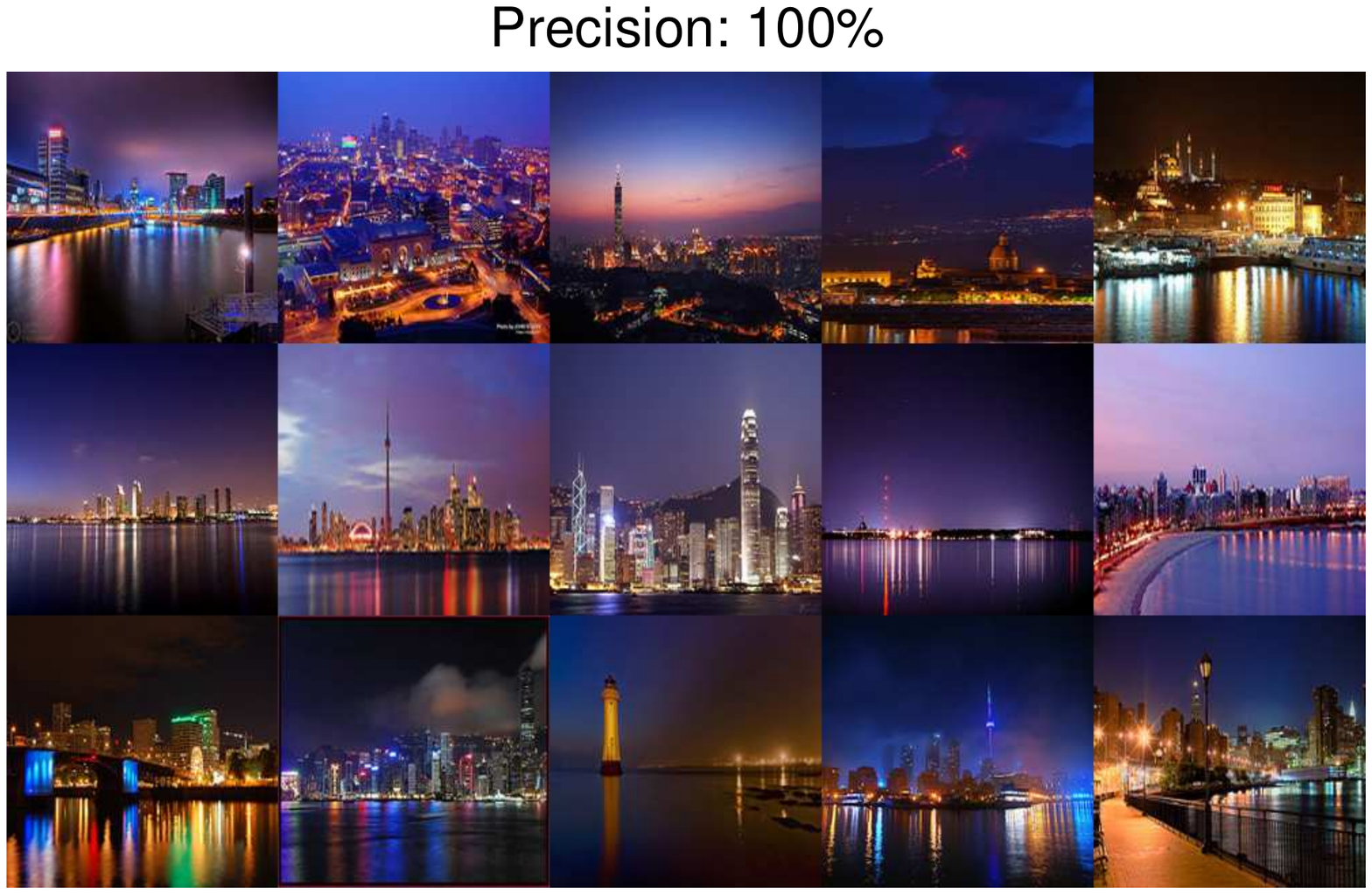}}
\subfigure[CCA (V+T+C).]{
   \includegraphics[width=1.9in, trim= 15mm 90mm 15mm
85mm]{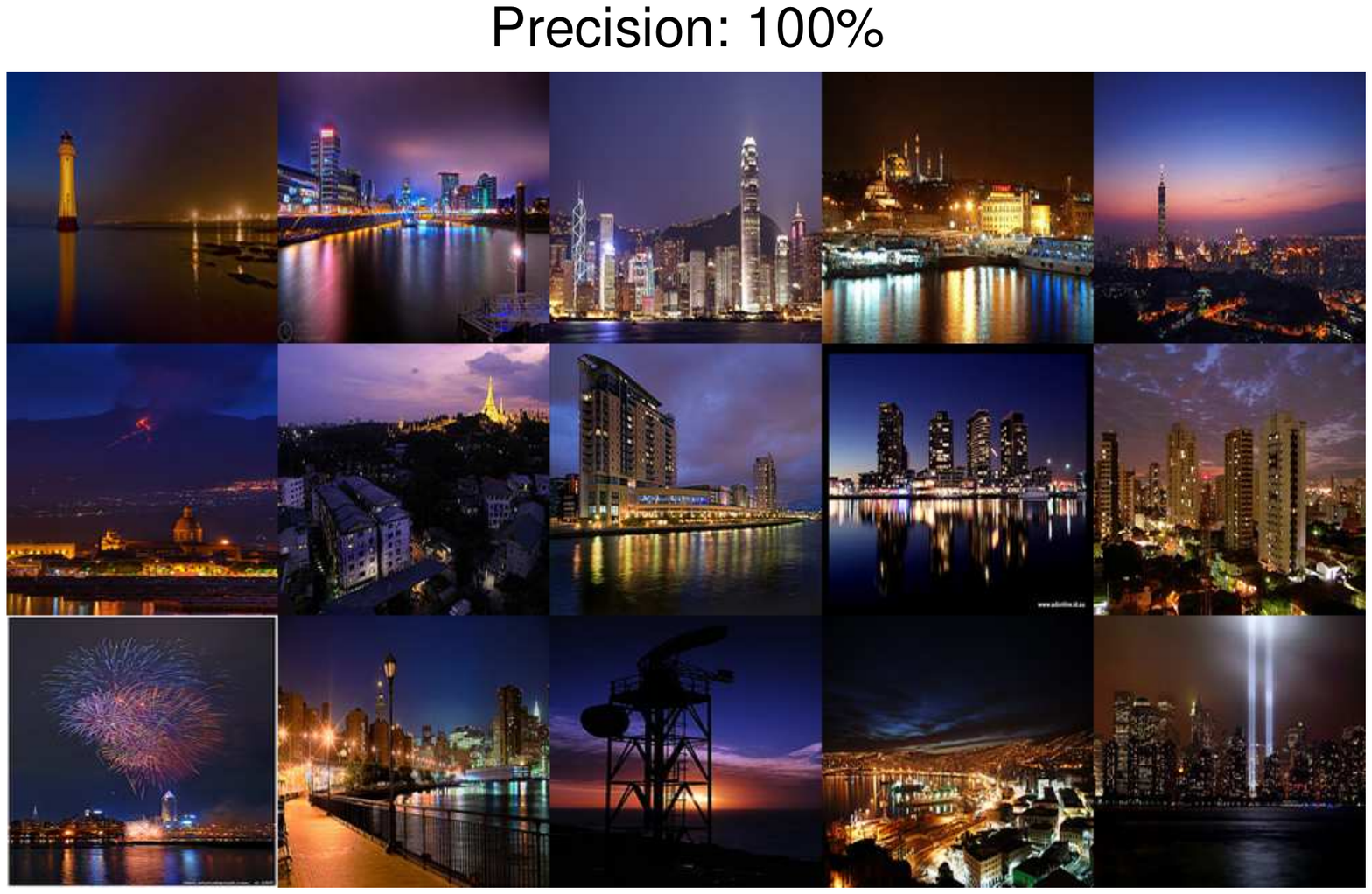}}
   \caption{Image-to-image retrieval results  on the NUS-WIDE dataset. The query image is shown on the left, together with its ground truth concepts. Red borders indicate false positive retrieval results. We consider an image to be a false positive if its ground truth annotation does not share any concepts with the query. Please note, however, that the ground truth is noisy, so some false (resp. true) positives are labeled inaccurately.}
      \label{sample_i2i2nus}
\end{figure*}

Table \ref{nus1} reports results for different multi-view models on I2I, T2I, and K2I search. For the supervised K view, we directly use the ground truth annotations (which may contain multiple nonzero entries per image). On this dataset, the best performance is achieved by the supervised CCA (V+T+K) and CCA (V+K) models. The unsupervised three-view model CCA (V+T+C) still improves over CCA (V+T) for all three tasks, but not as much as CCA (V+T+K).  By contrast, on the Flickr-CIFAR dataset (Table \ref{main}), we found that CCA (V+T+C) and CCA (V+T+K) were very close together. The weaker performance of the unsupervised three-view model on NUS-WIDE is not entirely surprising, however, since the tag clusters for NUS-WIDE are likely much more mixed than for Flickr-CIFAR, whose concepts were fewer and better separated. Intuitively, for richer and more diverse datasets, ground truth annotations are likely to be the strongest source of semantic information.
Also, unlike in Table \ref{main}, the two-view supervised model CCA (V+K) appears to have stronger results than the three-view CCA (V+T+K) for I2I and especially K2I. This may be due to the T view adding noise to the K view. Despite this, the two-view CCA (V+K) model is not as useful or flexible as the three-view CCA (V+T+K) one -- in particular, the former is not suitable for T2I retrieval.

\begin{figure*}[]
   \centering
\subfigure[\emph{mountain} river grass]{
   \includegraphics[width=1.5in, trim= 40mm 80mm 40mm
45mm]{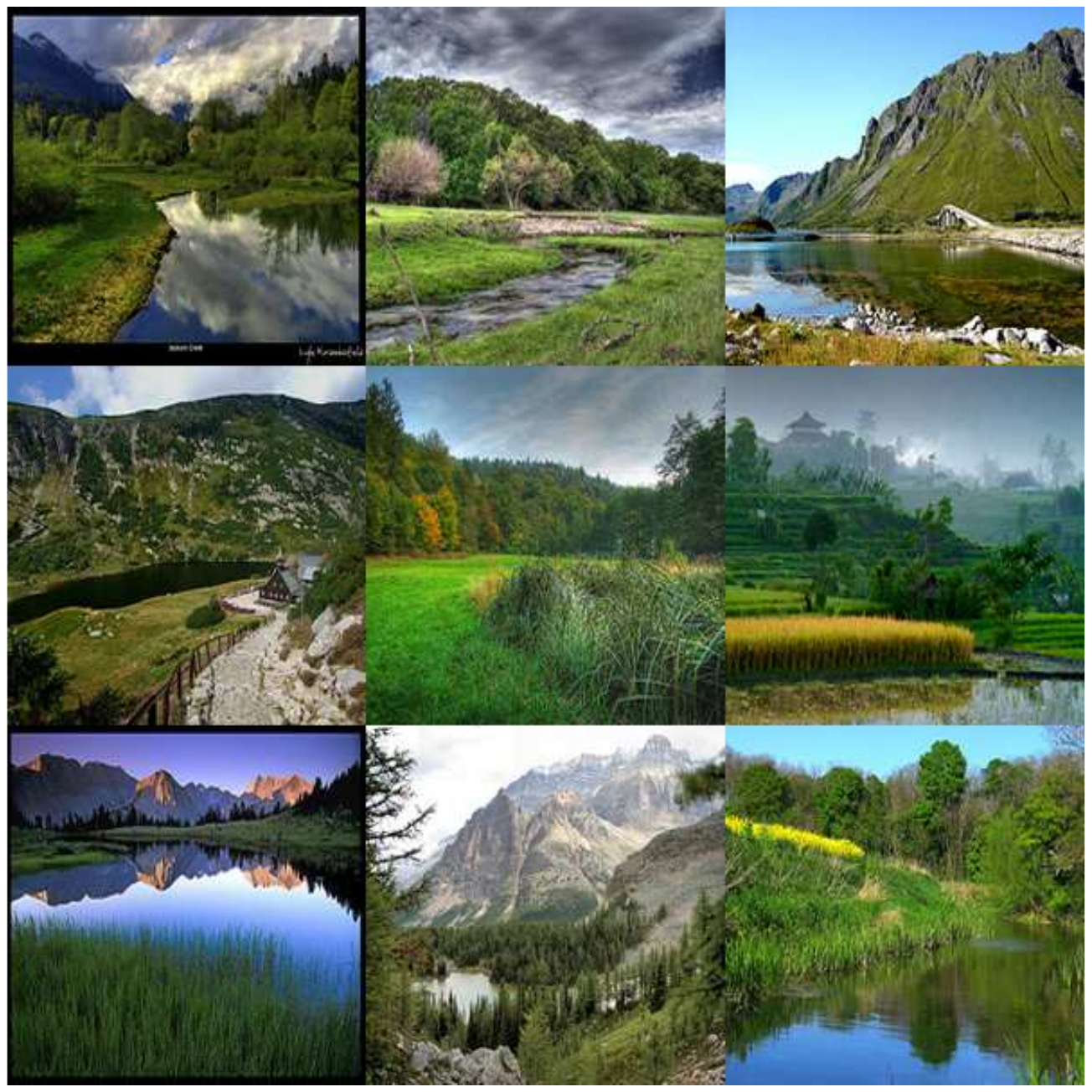}}
\subfigure[\emph{mountain} river \emph{tree}]{
   \includegraphics[width=1.5in, trim= 40mm 80mm 40mm
45mm]{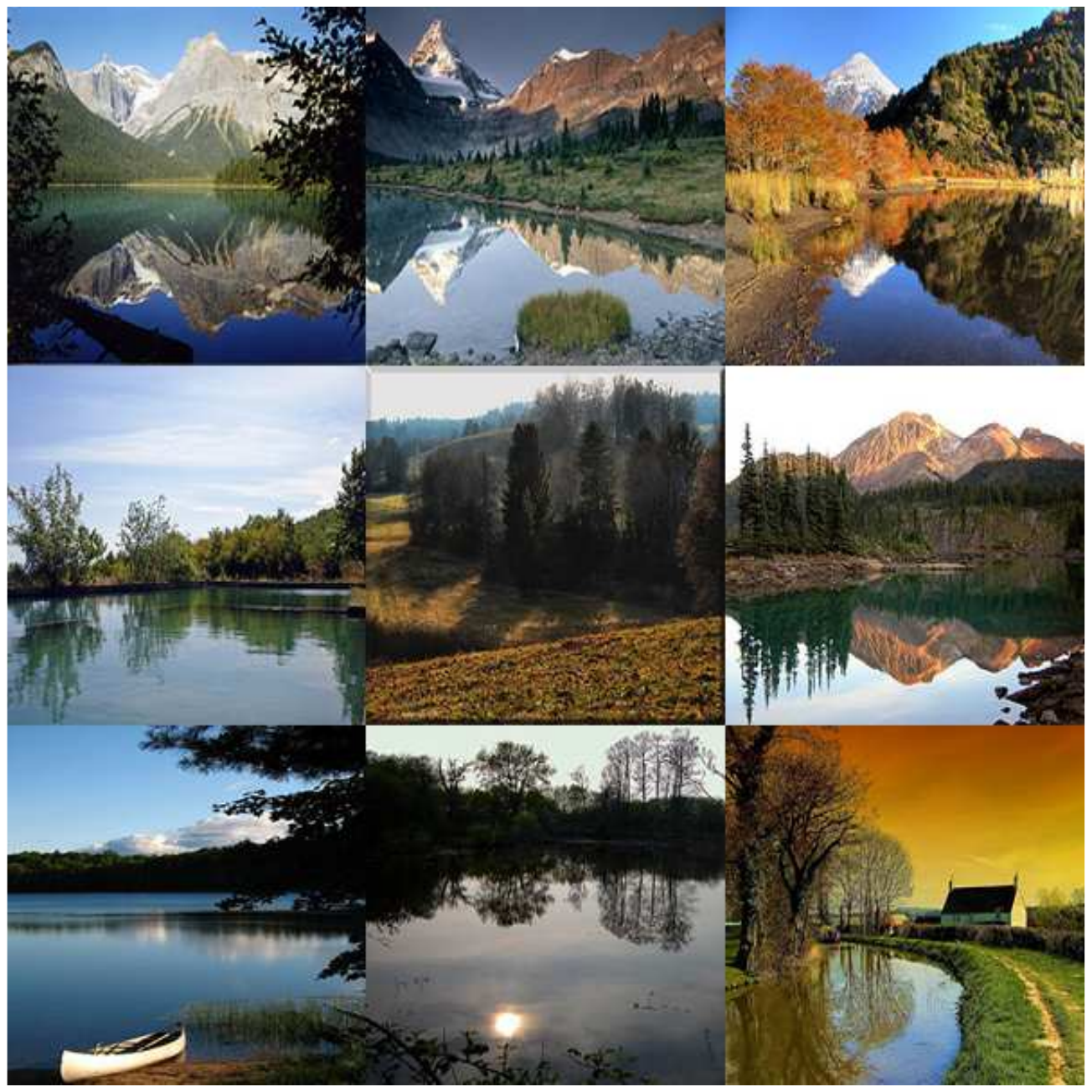}}
\subfigure[\emph{mountain} river \emph{sky}]{
   \includegraphics[width=1.5in, trim= 40mm 80mm 40mm
45mm]{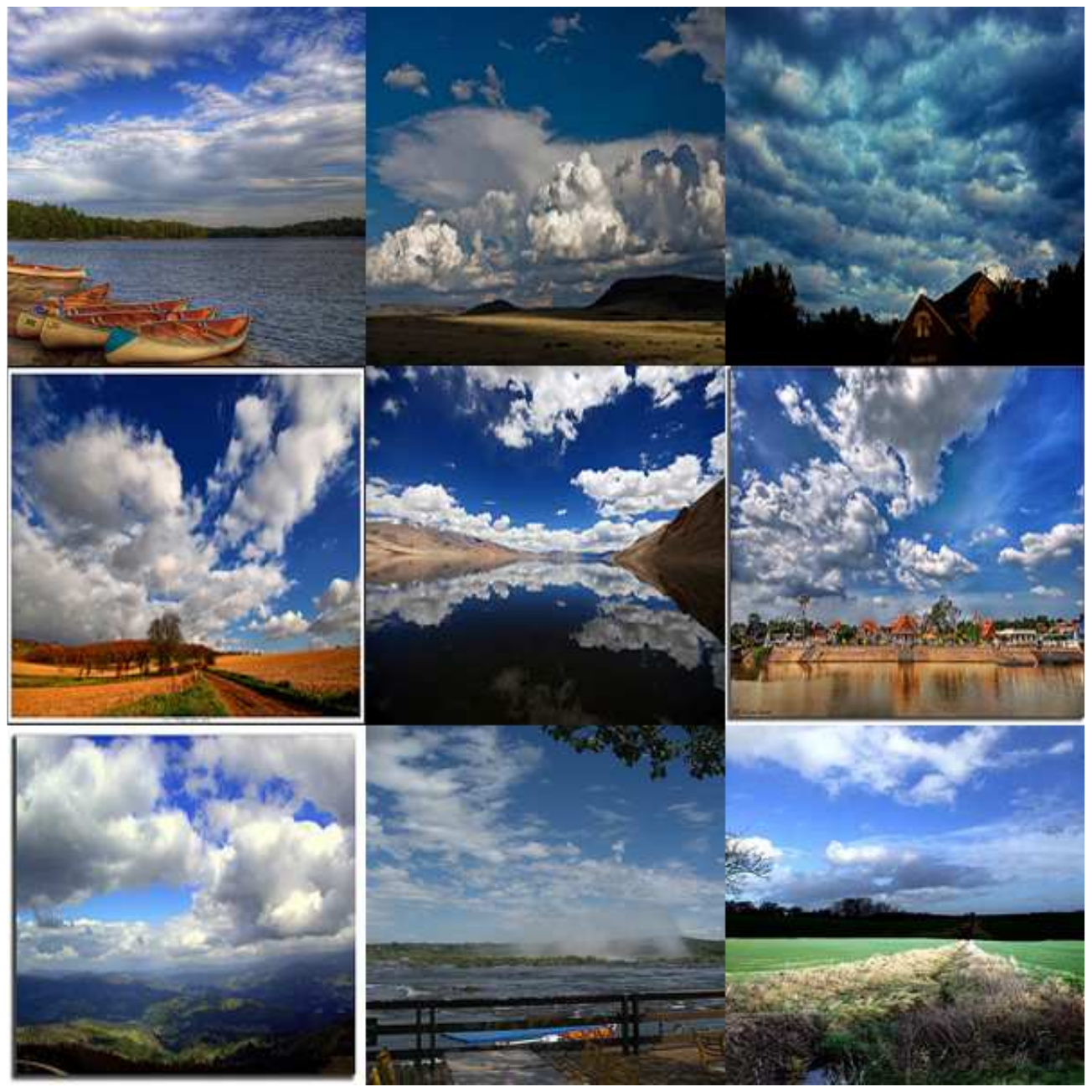}}
\subfigure[\emph{mountain} river \emph{waterfalls}]{
   \includegraphics[width=1.5in, trim= 40mm 80mm 40mm
45mm]{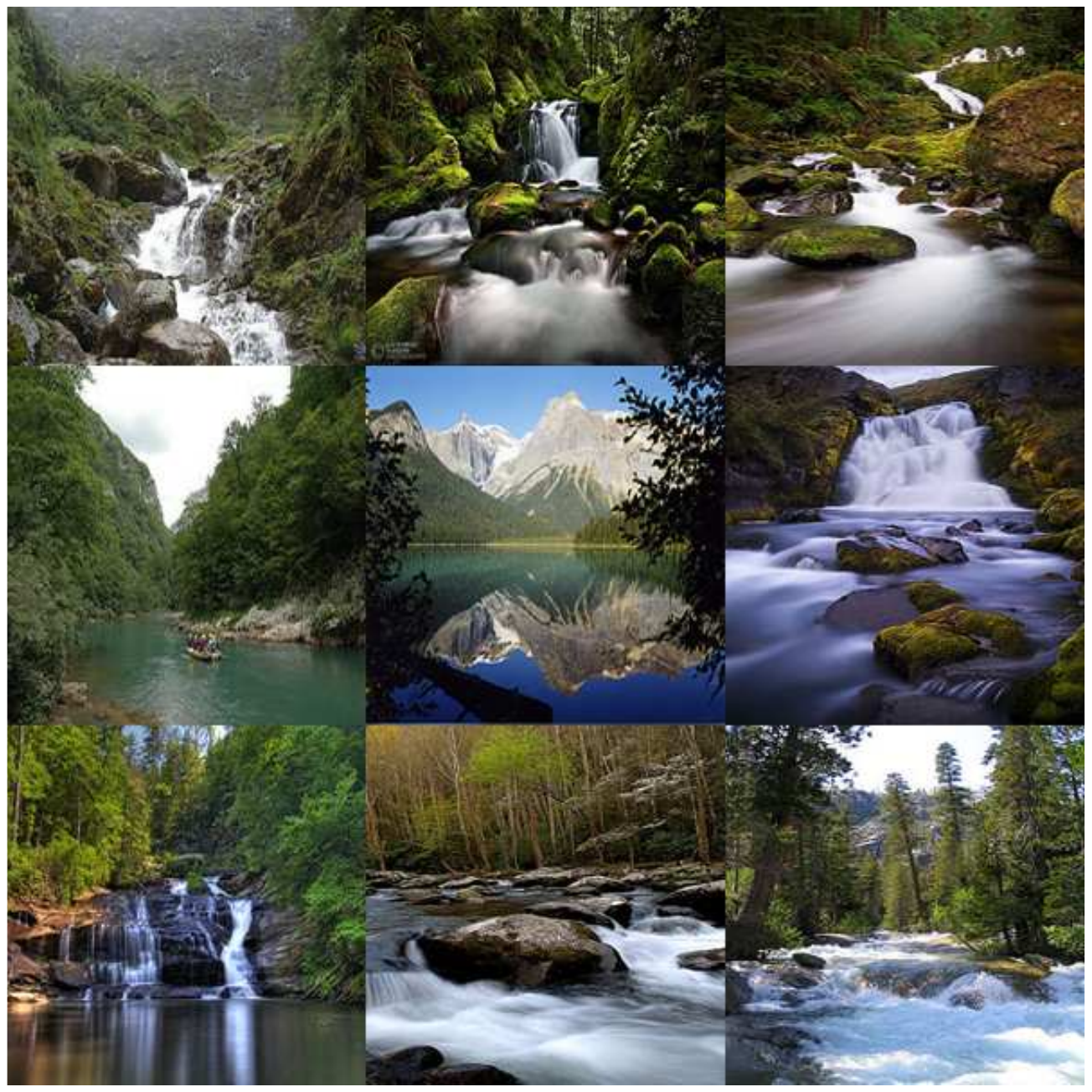}}
\subfigure[\emph{mountain} river]{
   \includegraphics[width=1.5in, trim=44mm 85mm 44mm
75mm]{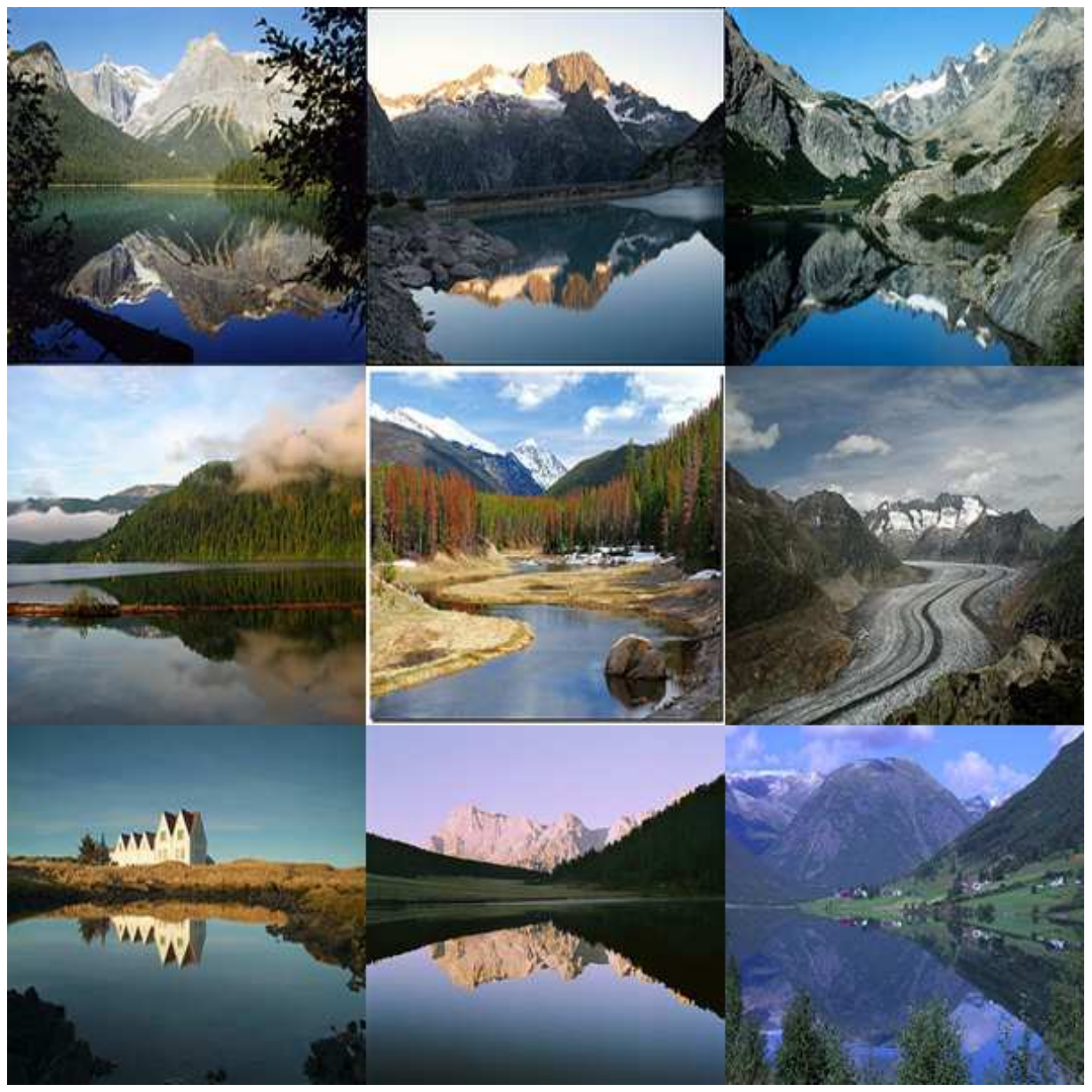}}
\subfigure[river]{
   \includegraphics[width=1.5in, trim= 44mm 85mm 44mm
75mm]{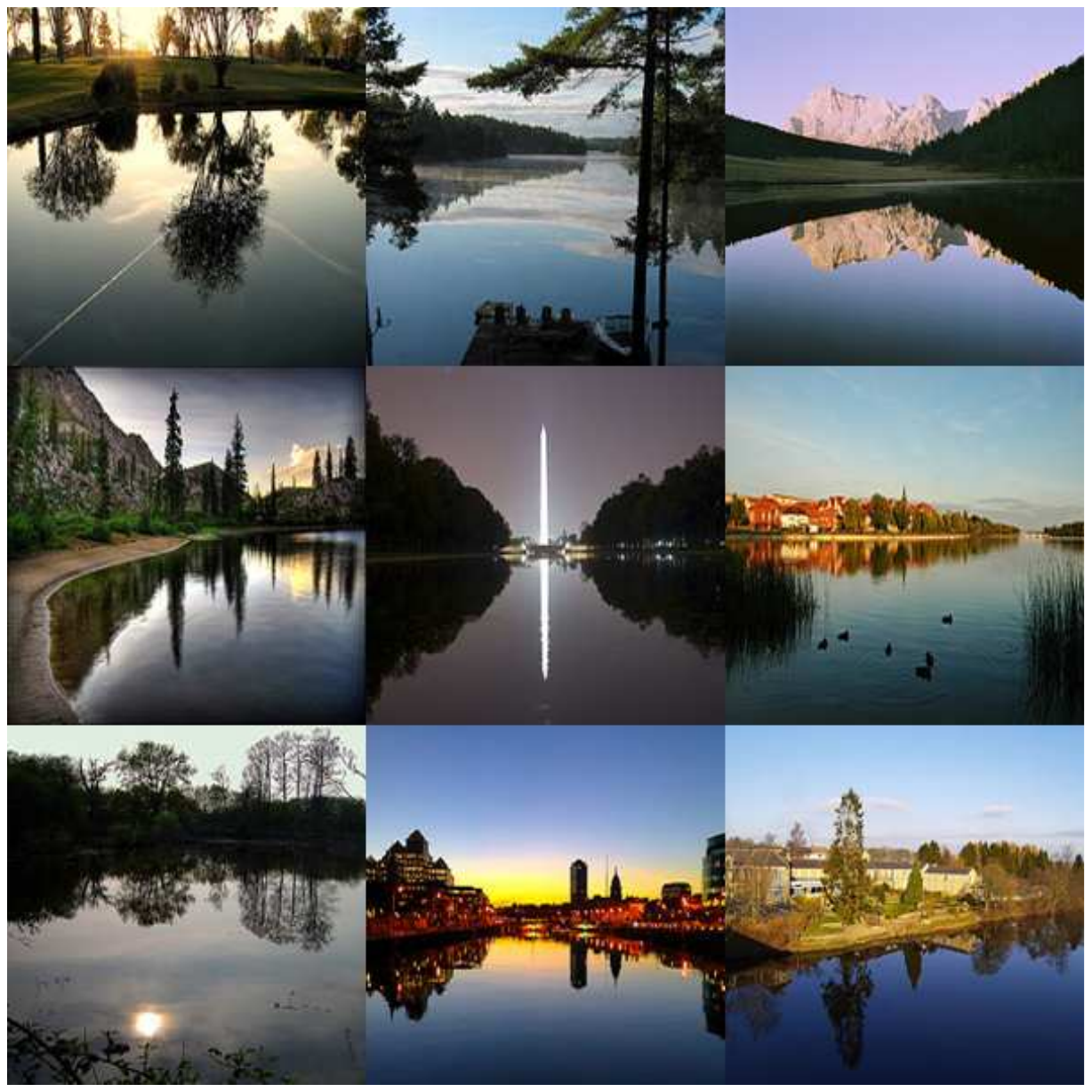}}
\subfigure[\emph{reflection city} river]{
   \includegraphics[width=1.5in, trim= 40mm 80mm 40mm
75mm]{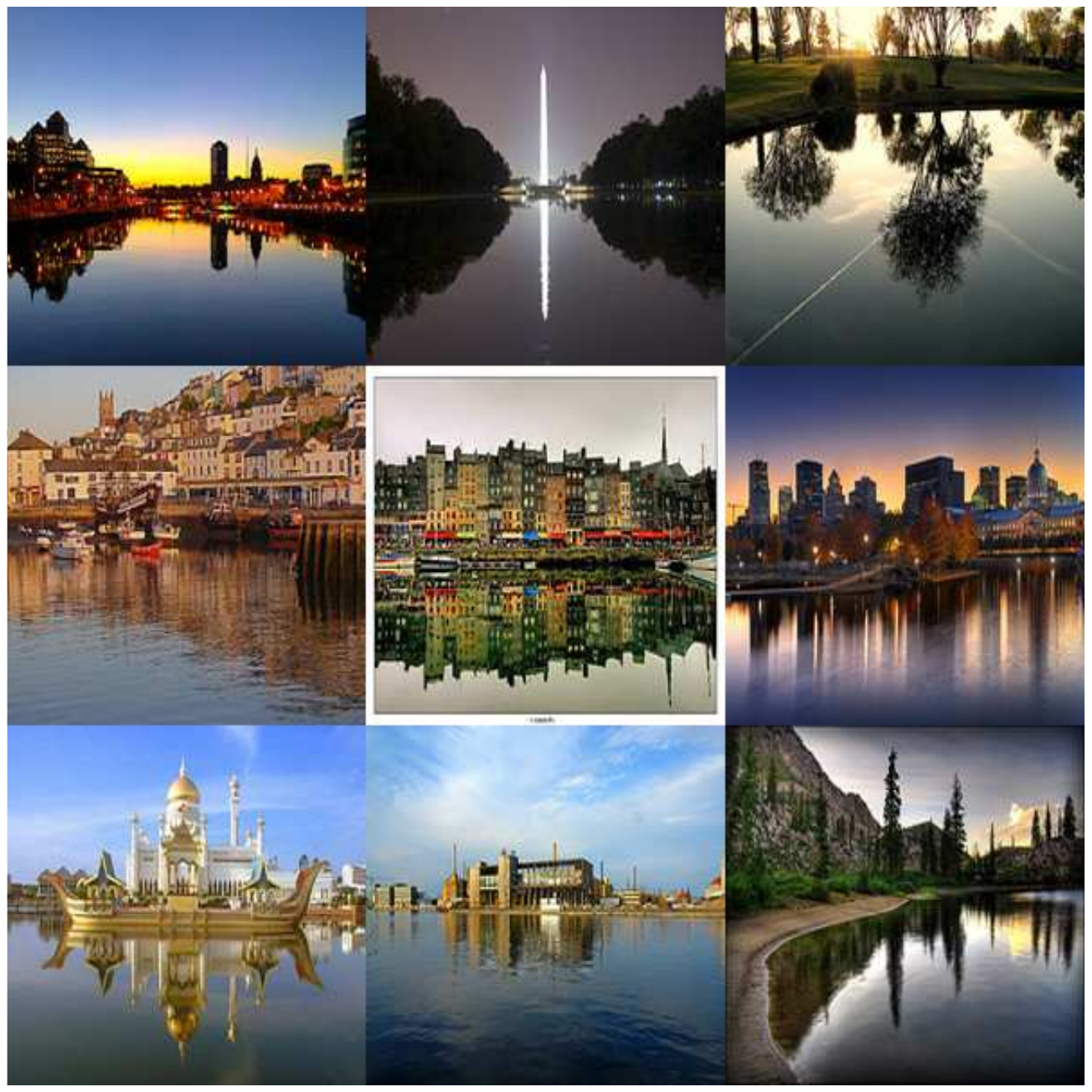}}
\subfigure[\emph{reflection city}]{
 \includegraphics[width=1.5in, trim= 40mm 80mm 40mm
75mm]{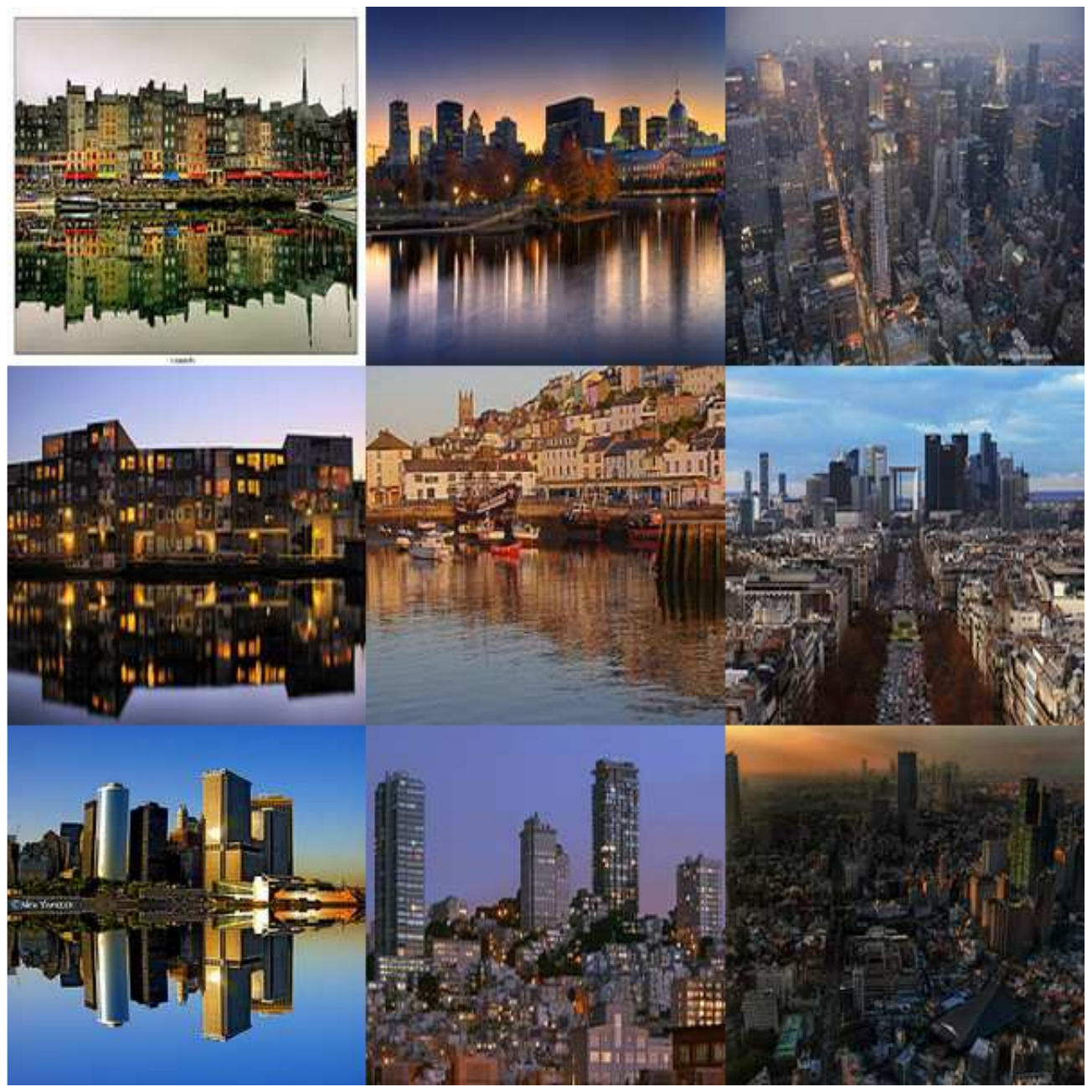}}
\subfigure[\emph{city}]{
   \includegraphics[width=1.5in, trim= 44mm 85mm 44mm
75mm]{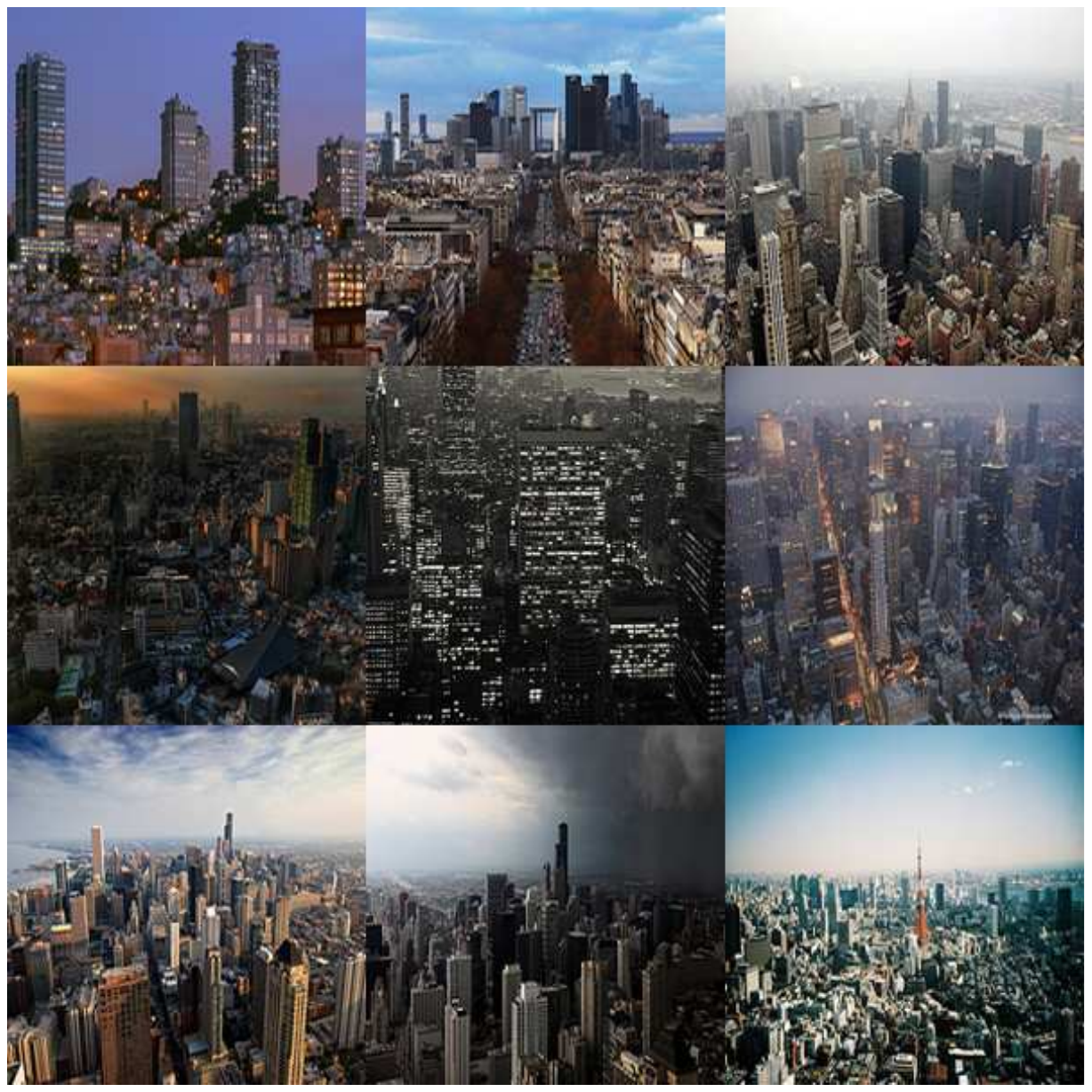}}
\subfigure[night \emph{city}]{
   \includegraphics[width=1.5in, trim= 44mm 85mm 44mm
75mm]{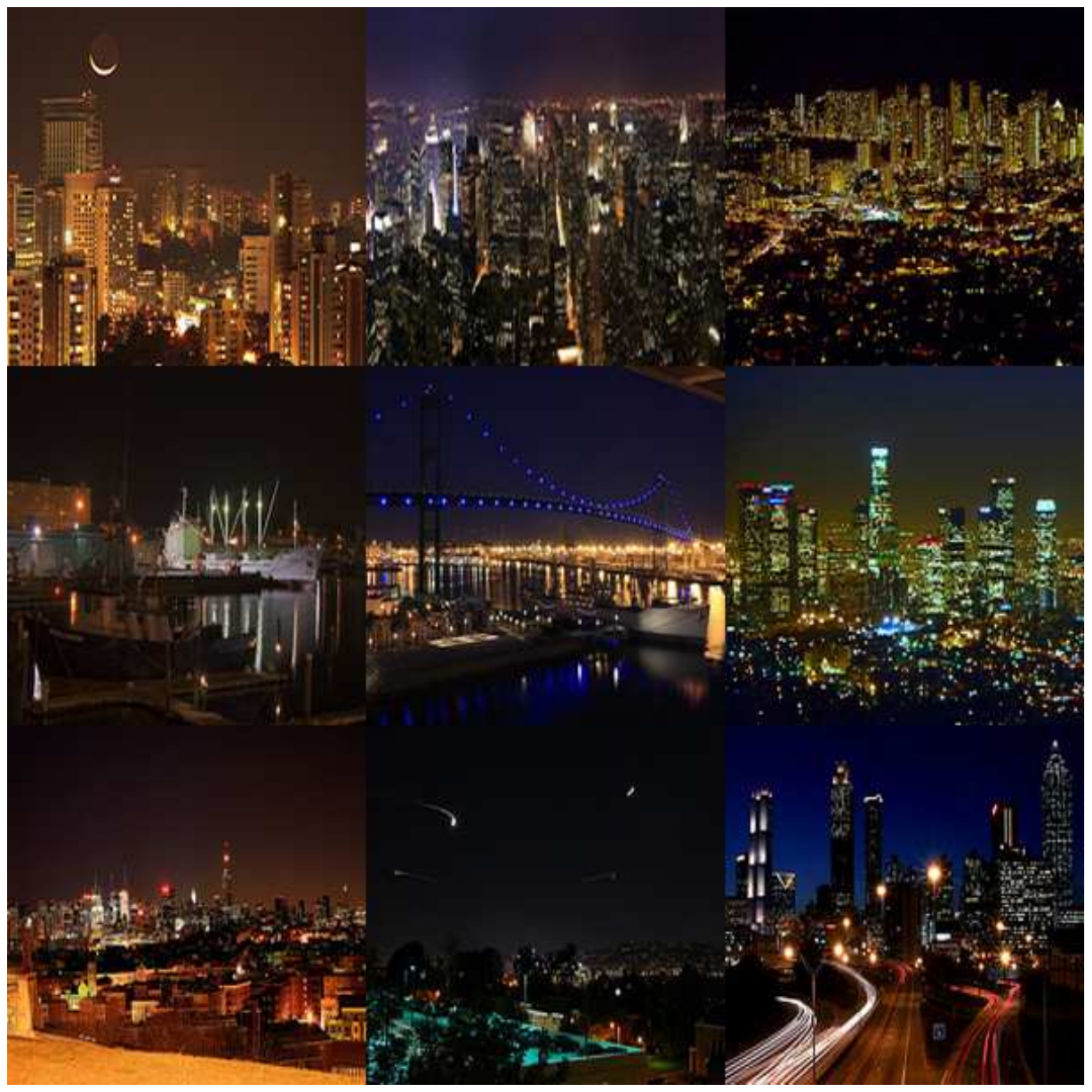}}
\subfigure[\emph{city} fog]{
 \includegraphics[width=1.5in,  trim= 44mm 85mm 44mm
75mm]{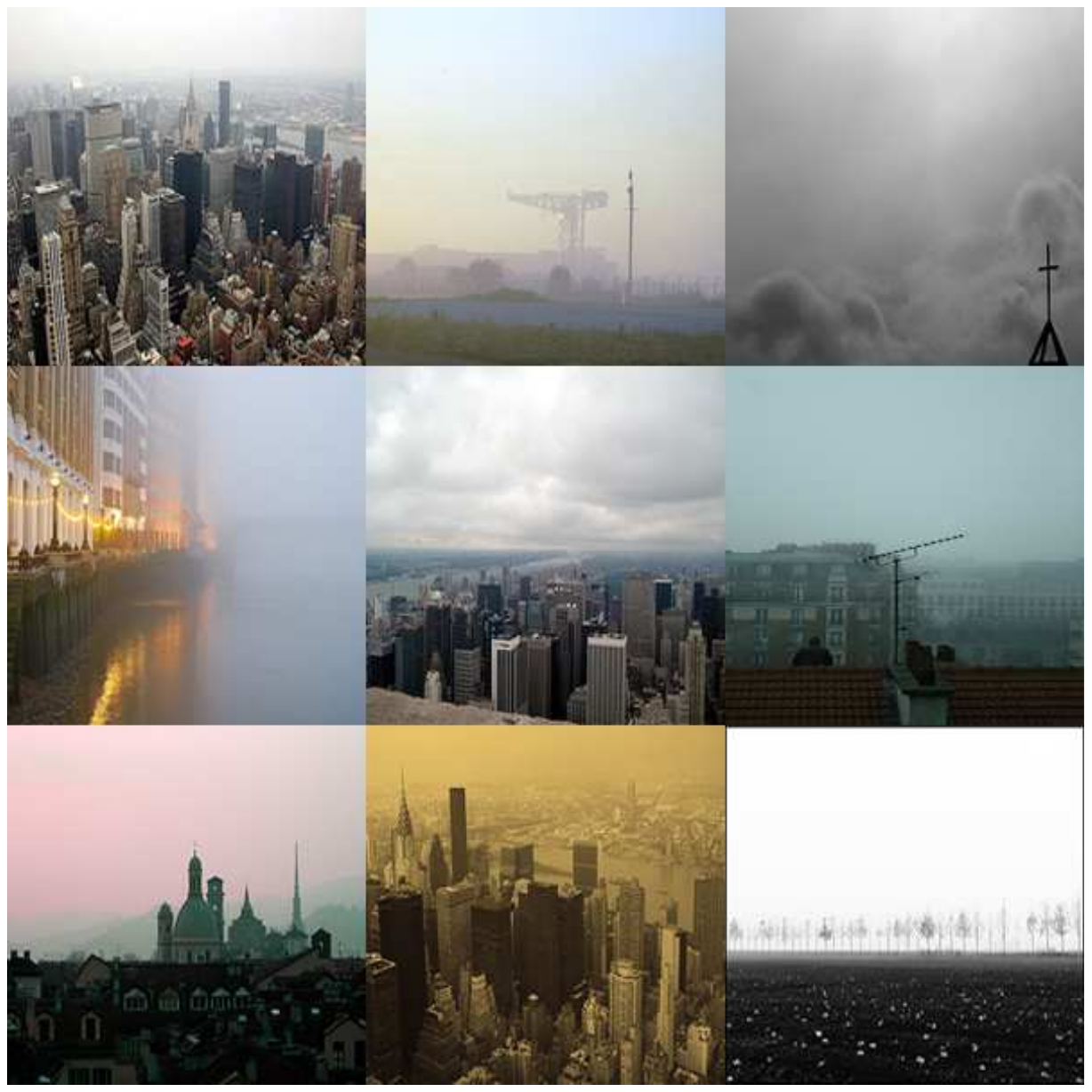}}
\subfigure[fog]{
   \includegraphics[width=1.5in,  trim= 44mm 85mm 44mm
75mm]{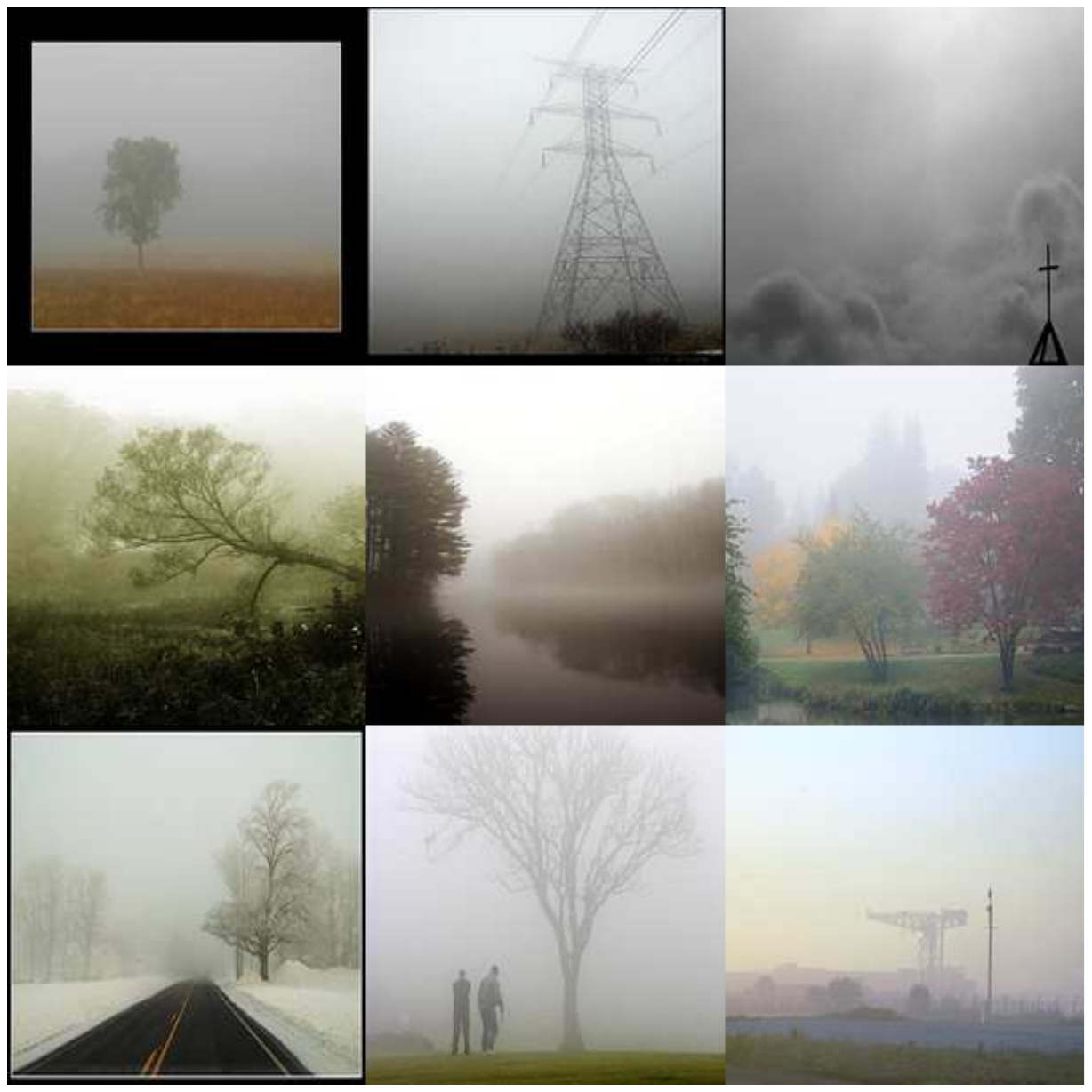}}
\subfigure[\emph{cloud}]{
   \includegraphics[width=1.5in, trim= 44mm 85mm 44mm
75mm]{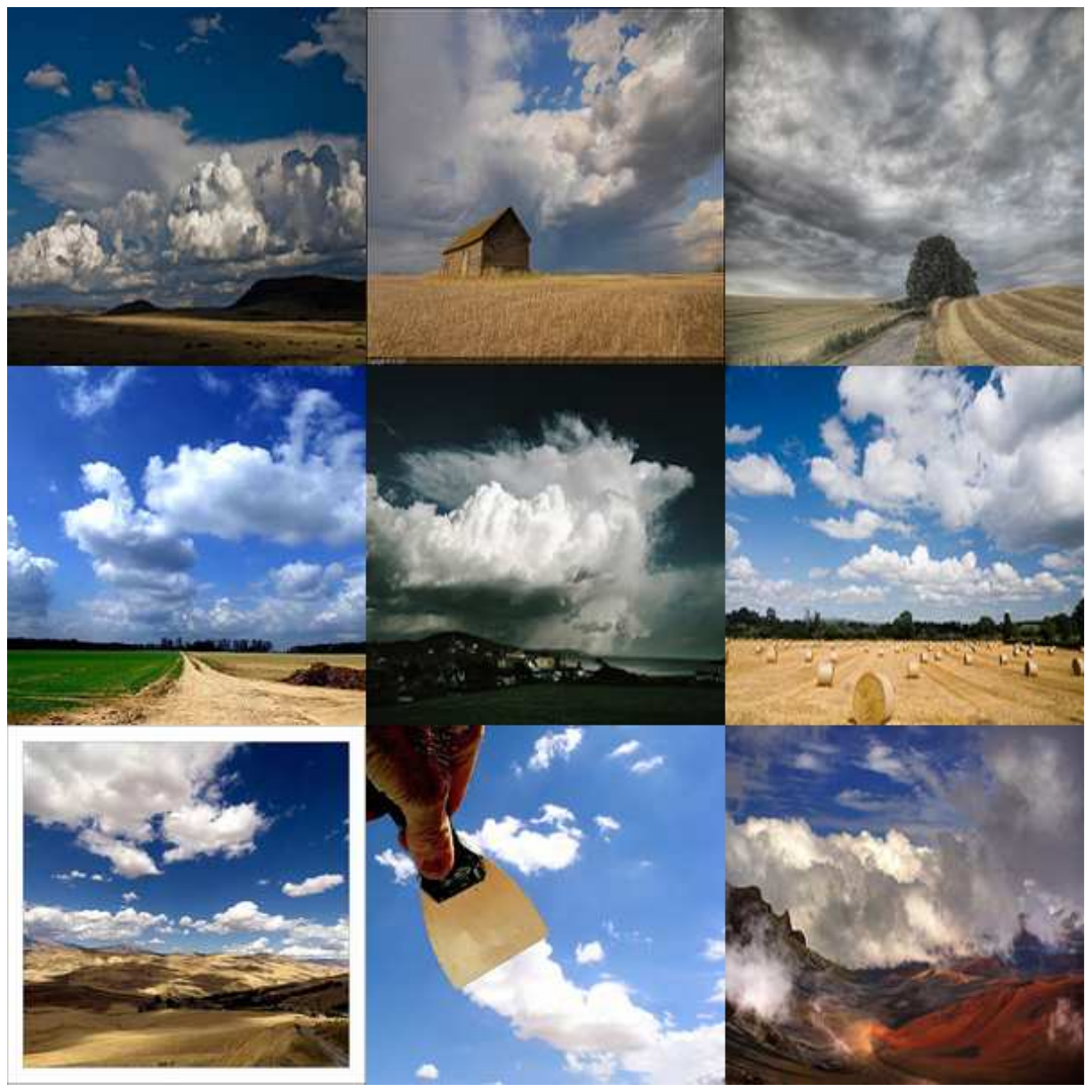}}
\subfigure[golden \emph{cloud}]{
   \includegraphics[width=1.5in, trim= 44mm 85mm 44mm
75mm]{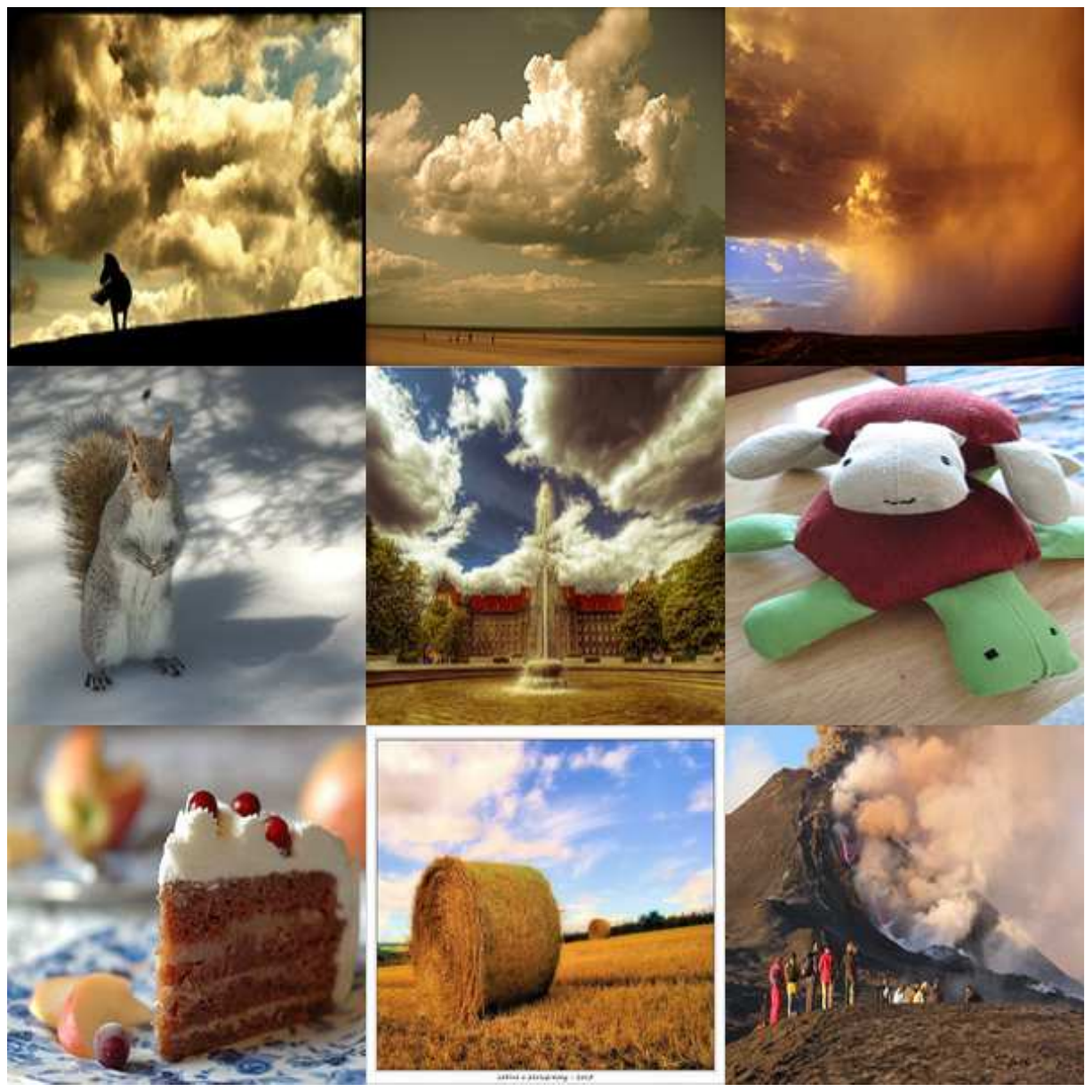}}
\subfigure[storm]{
  \includegraphics[width=1.5in, trim= 44mm 85mm 44mm
75mm]{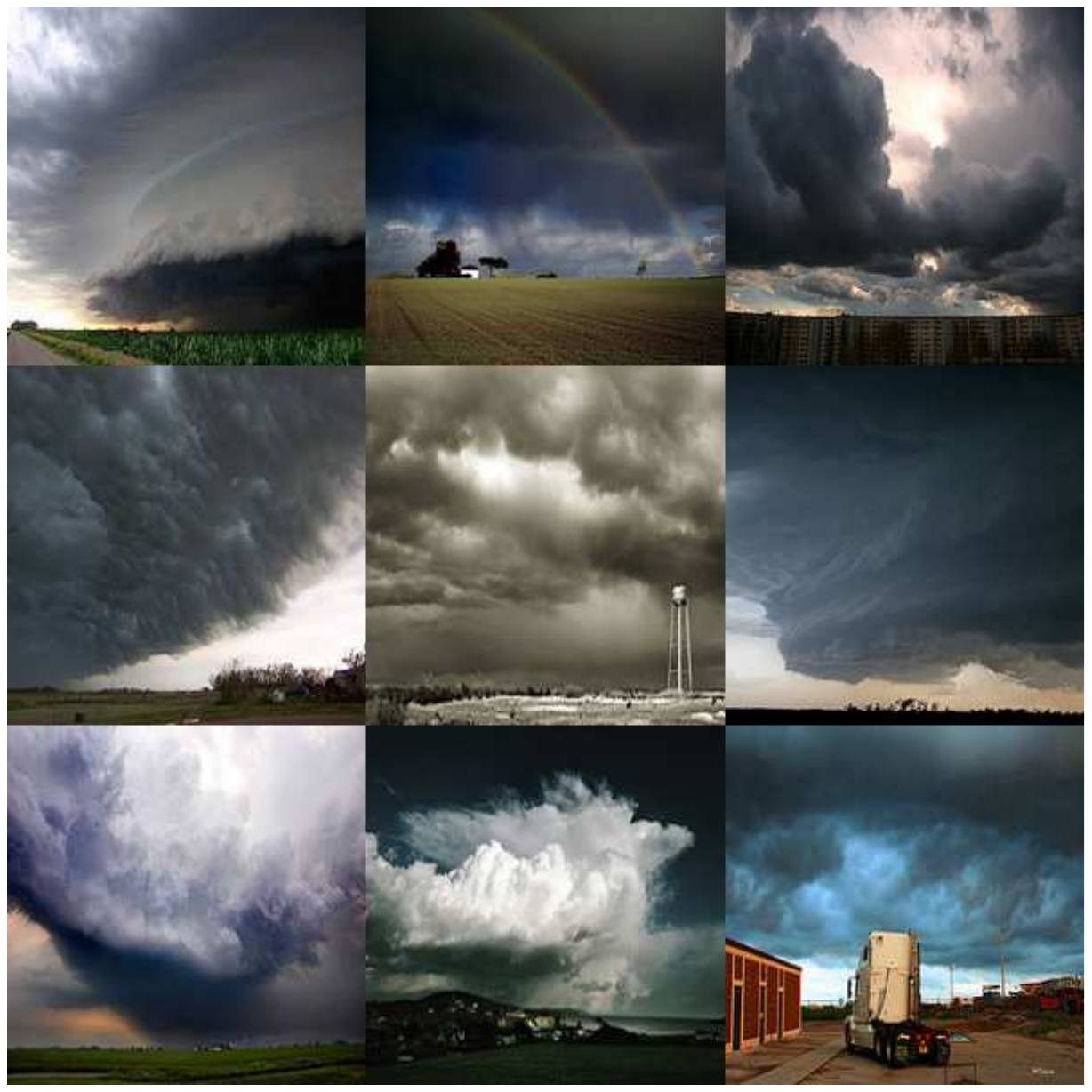}}
\subfigure[storm \emph{beach}]{
 \includegraphics[width=1.5in, trim= 44mm 85mm 44mm
78mm]{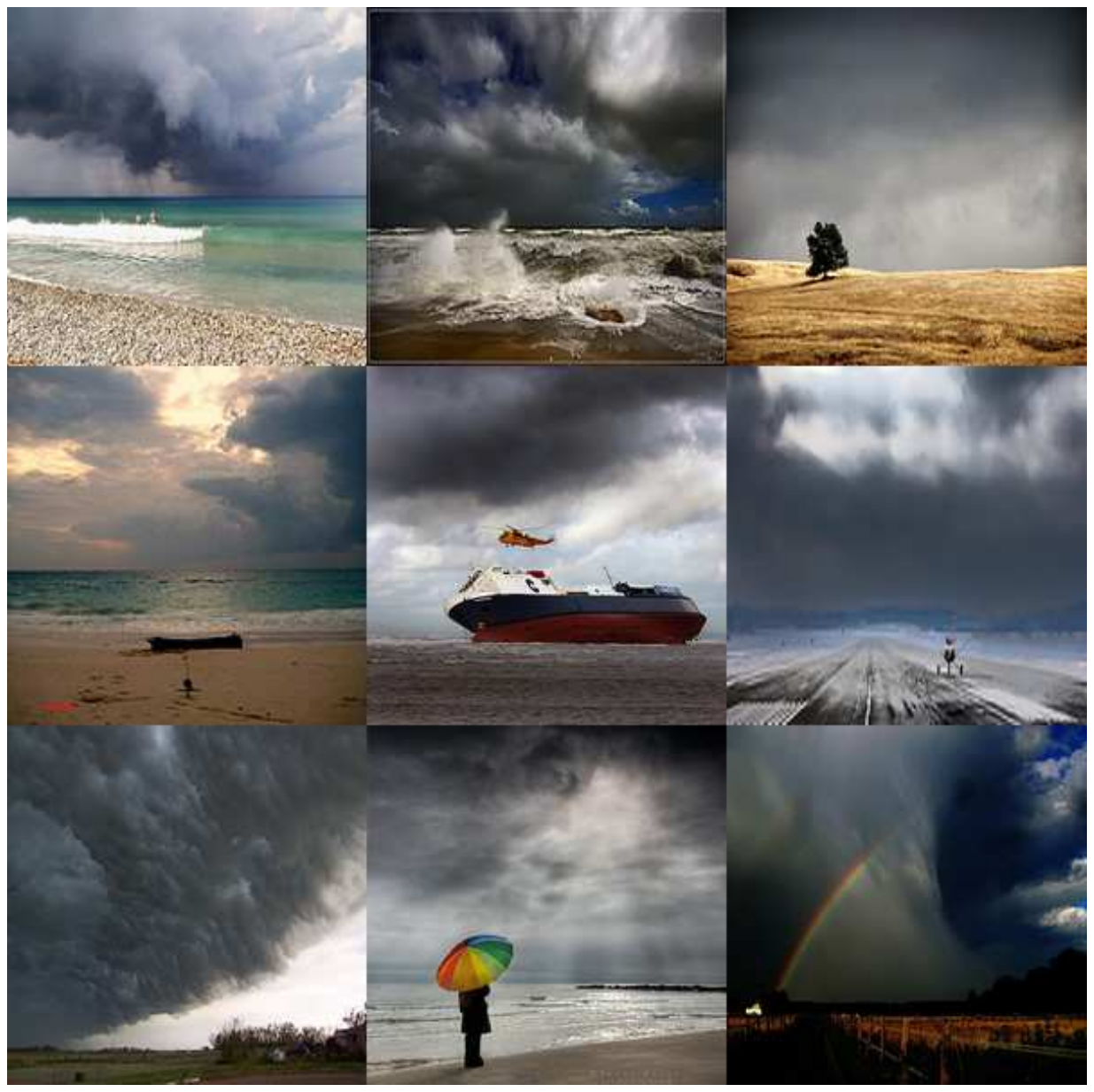}}
\subfigure[\emph{beach}]{
   \includegraphics[width=1.5in, trim= 44mm 85mm 44mm
75mm]{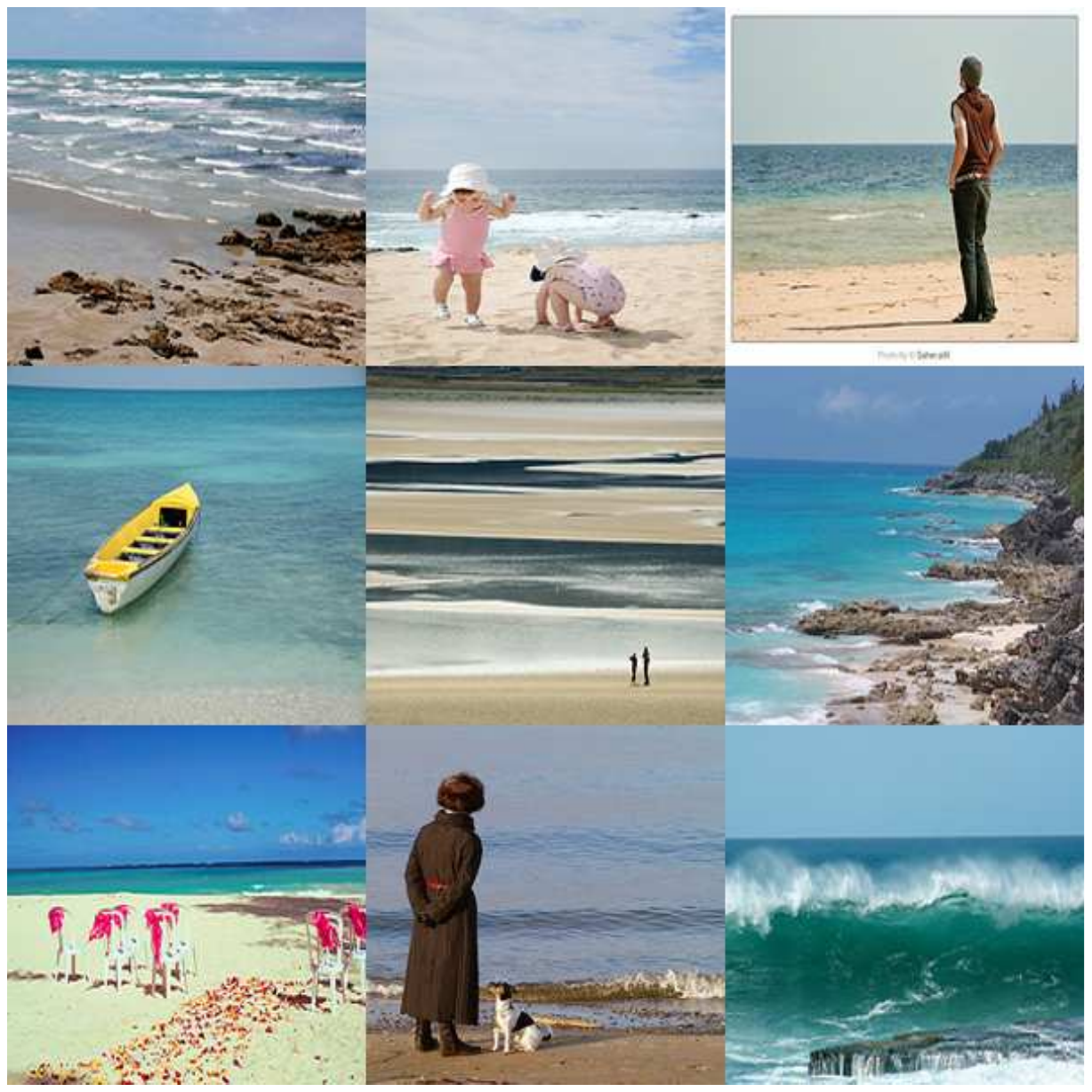}}
\subfigure[\emph{beach} people]{
   \includegraphics[width=1.5in, trim= 44mm 85mm 44mm
75mm]{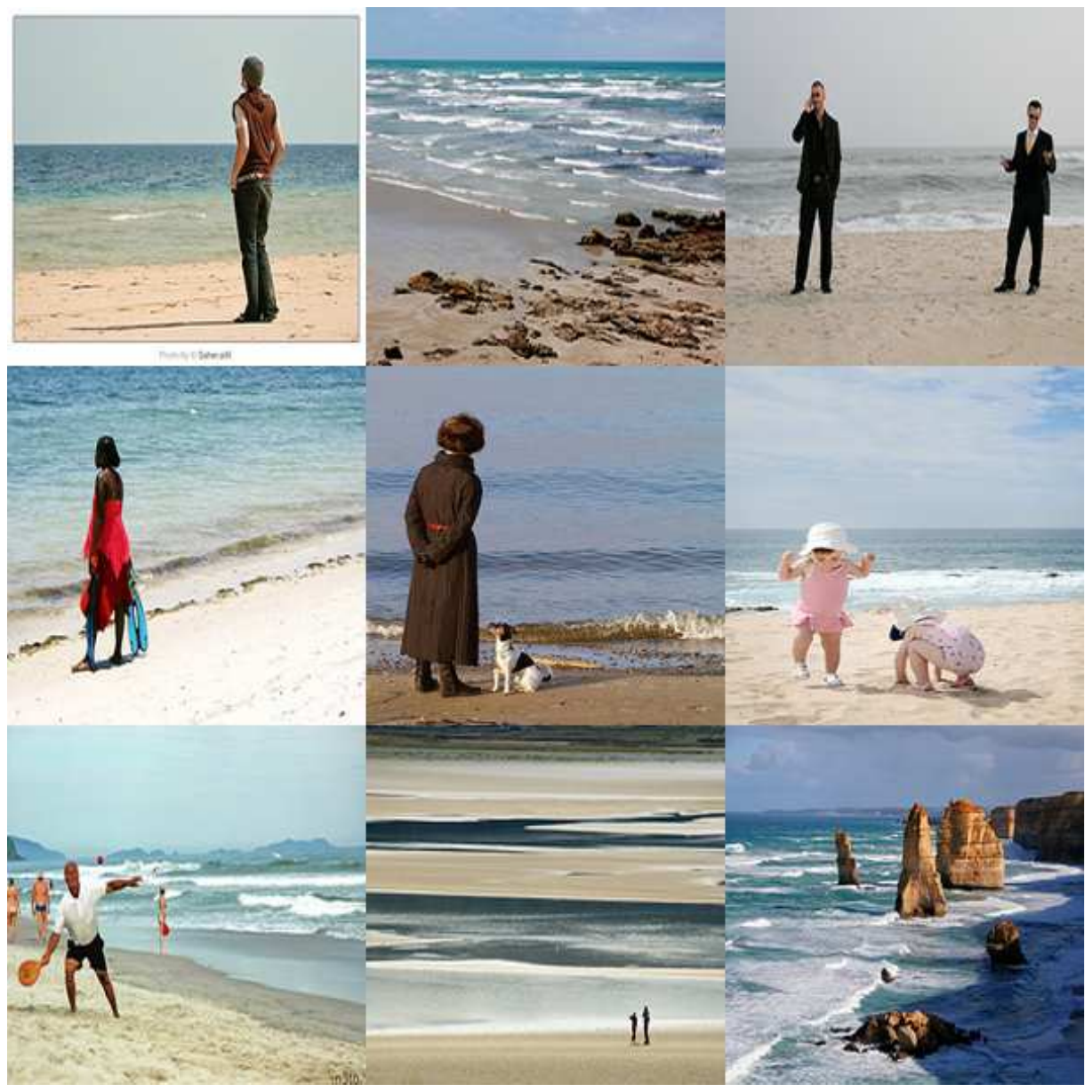}}
\subfigure[\emph{beach} people red]{
  \includegraphics[width=1.5in, trim= 40mm 80mm 40mm
75mm]{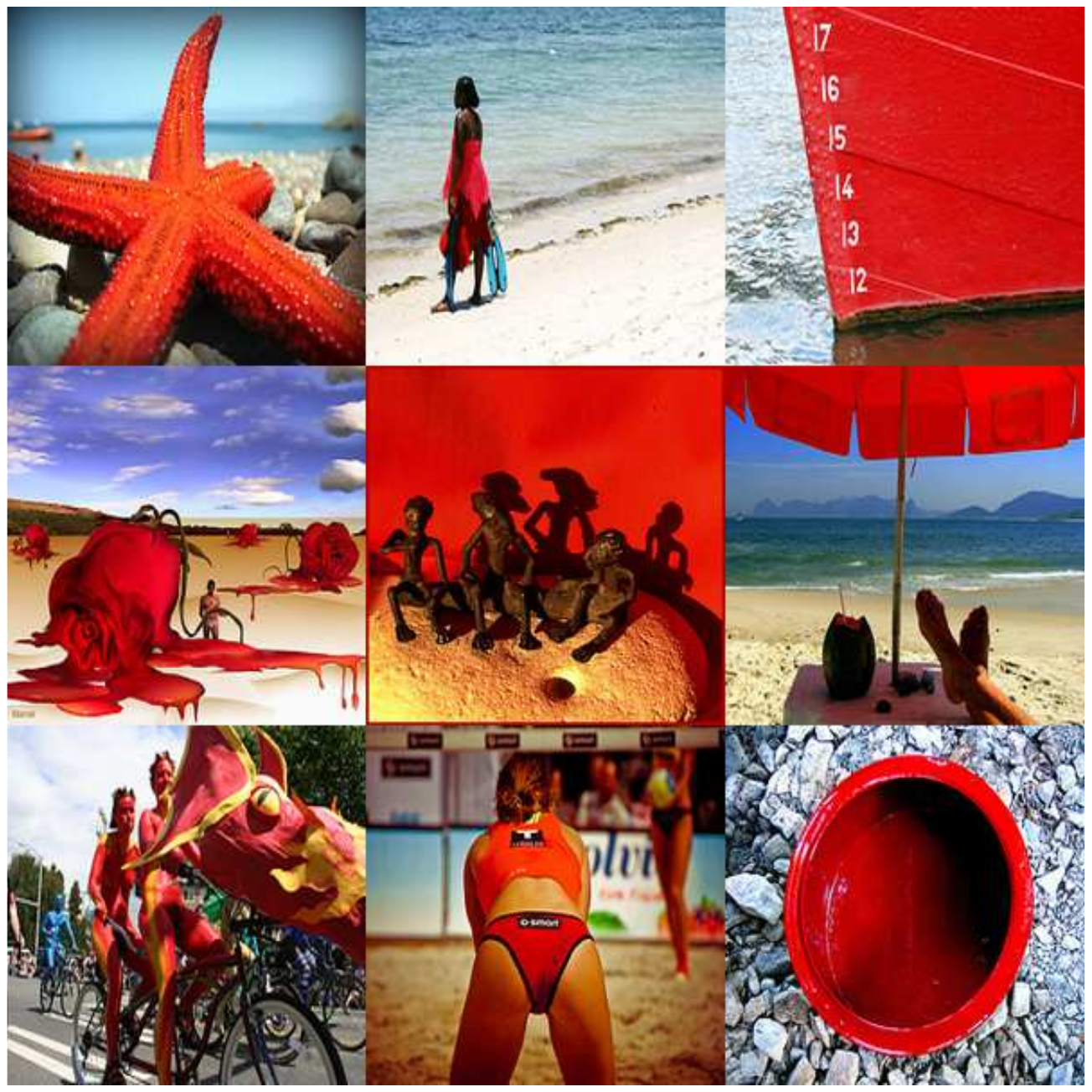}}
\subfigure[red]{
 \includegraphics[width=1.5in, trim= 40mm 79mm 40mm
78mm]{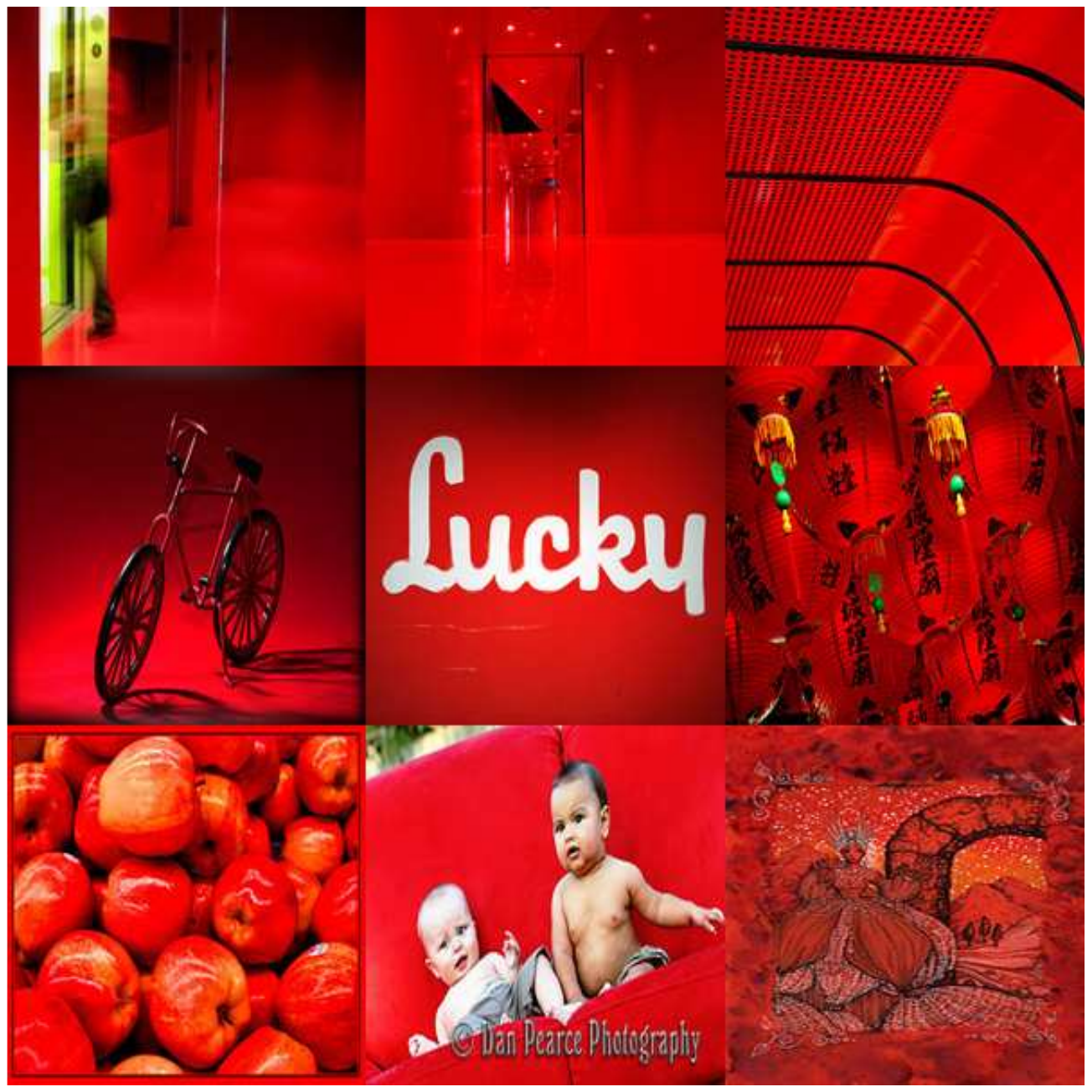}}
   \caption{Examples of tag-to-image search on the NUS-WIDE dataset with CCA (V+T+C). Tags in italic are also part of the 81-member semantic concept vocabulary. Notice that the three-view model can return appropriate images for combinations of up to three query tags.}
   \label{key2}
\end{figure*}

\begin{figure*}[t]
   \centering
\subfigure[CCA (V+T).]{
   \includegraphics[width=1.5in, trim= 43mm 80mm 37mm
75mm]{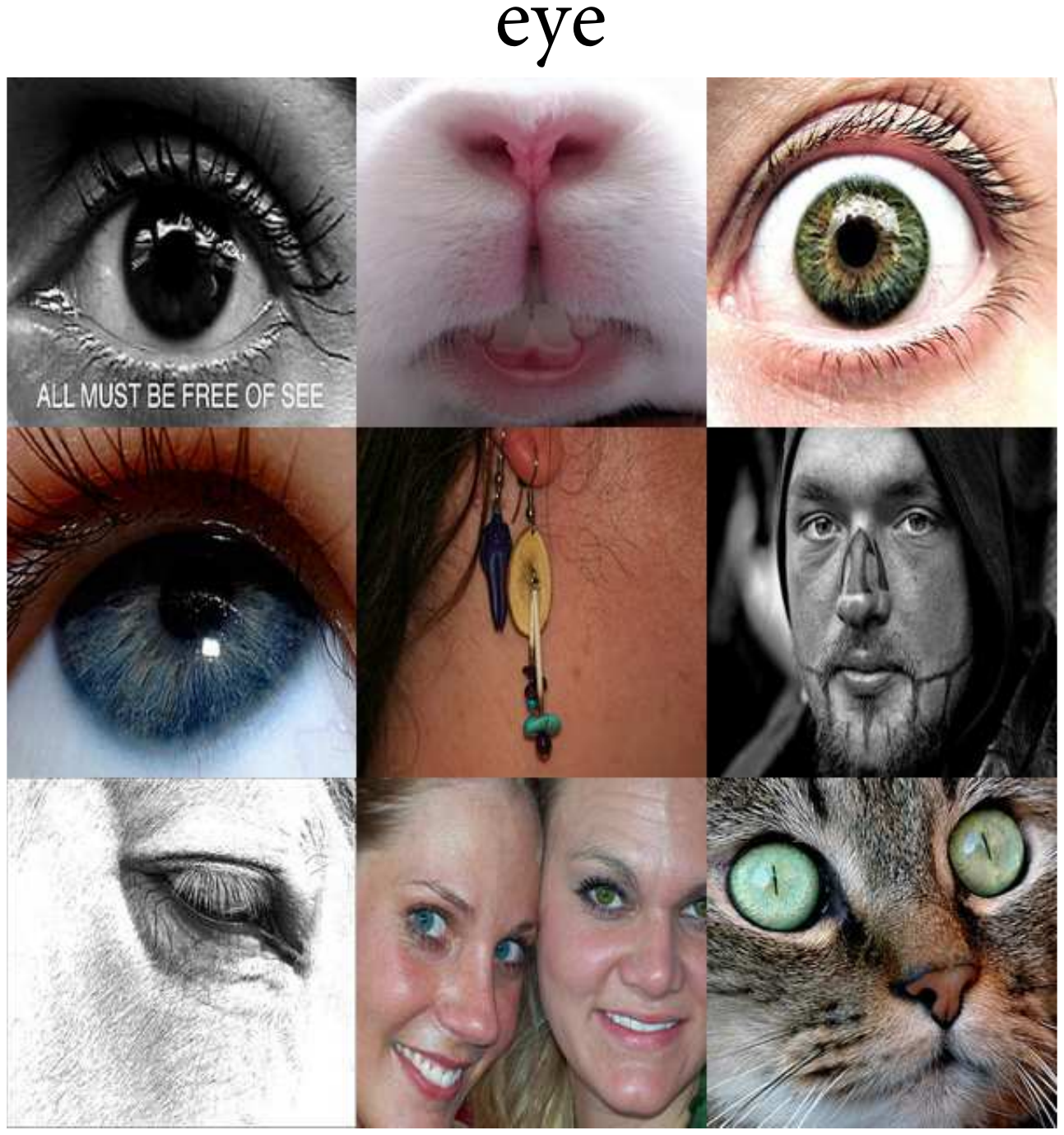}}
   \subfigure[CCA (V+T+C).]{
  \includegraphics[width=1.5in, trim= 43mm 80mm 37mm
75mm]{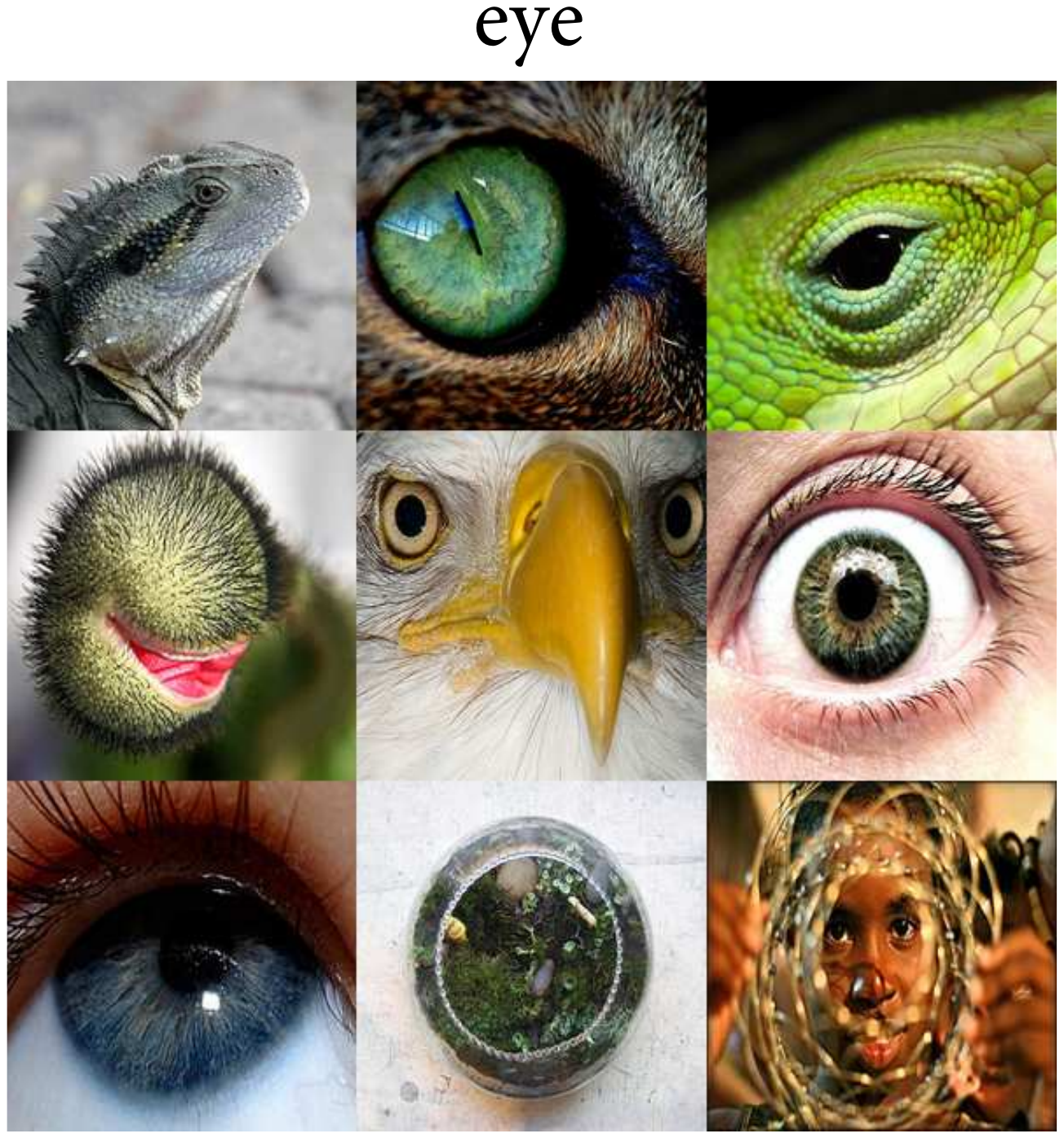}}
 \includegraphics[width=0.15in, trim=-1mm 5mm 1mm
0mm]{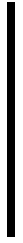}
\subfigure[CCA (V+T).]{
   \includegraphics[width=1.5in, trim= 43mm 80mm 37mm
75mm]{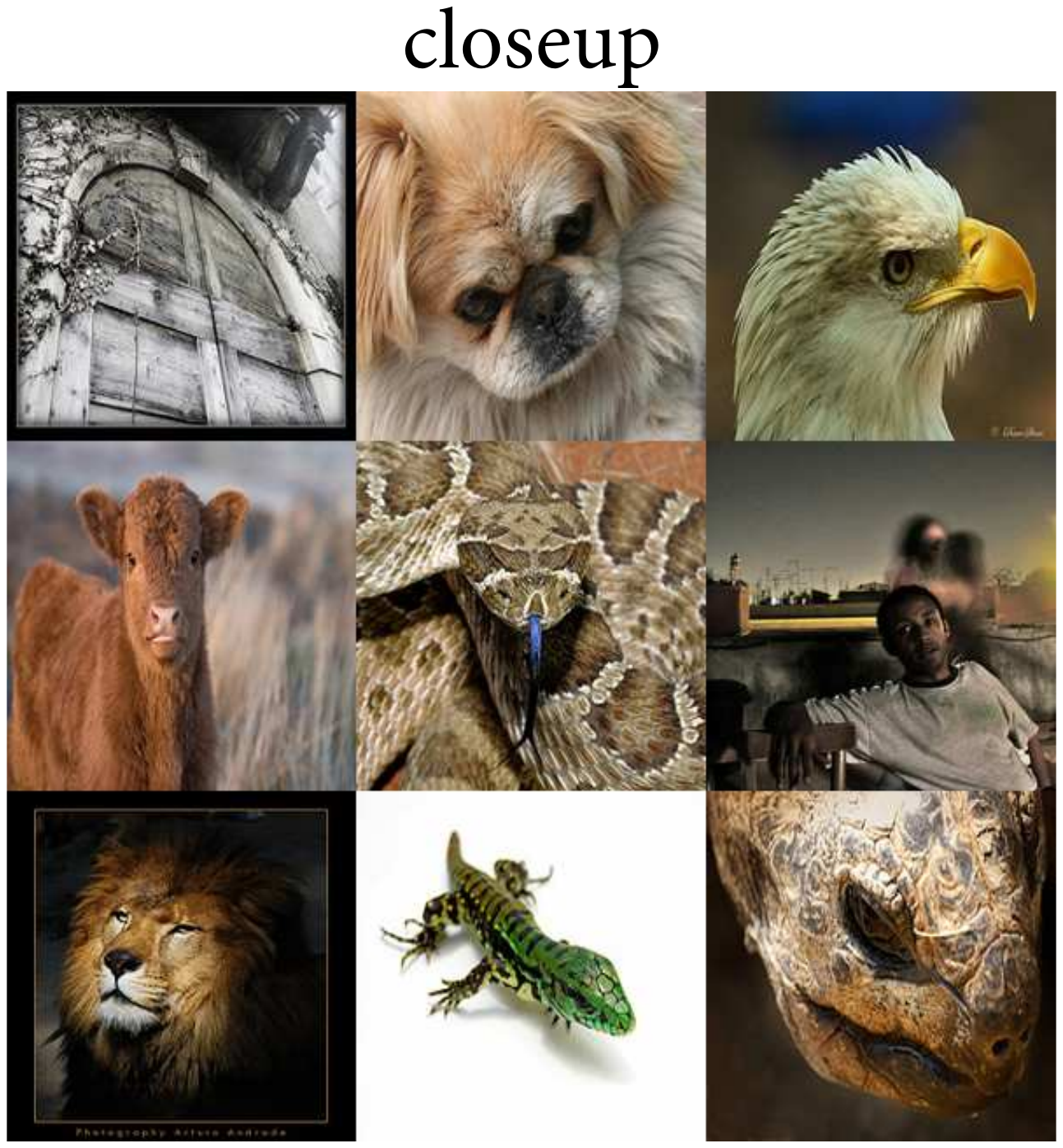}}
   \subfigure[CCA (V+T+C).]{
 \includegraphics[width=1.5in, trim= 45mm 82mm 39mm
75mm]{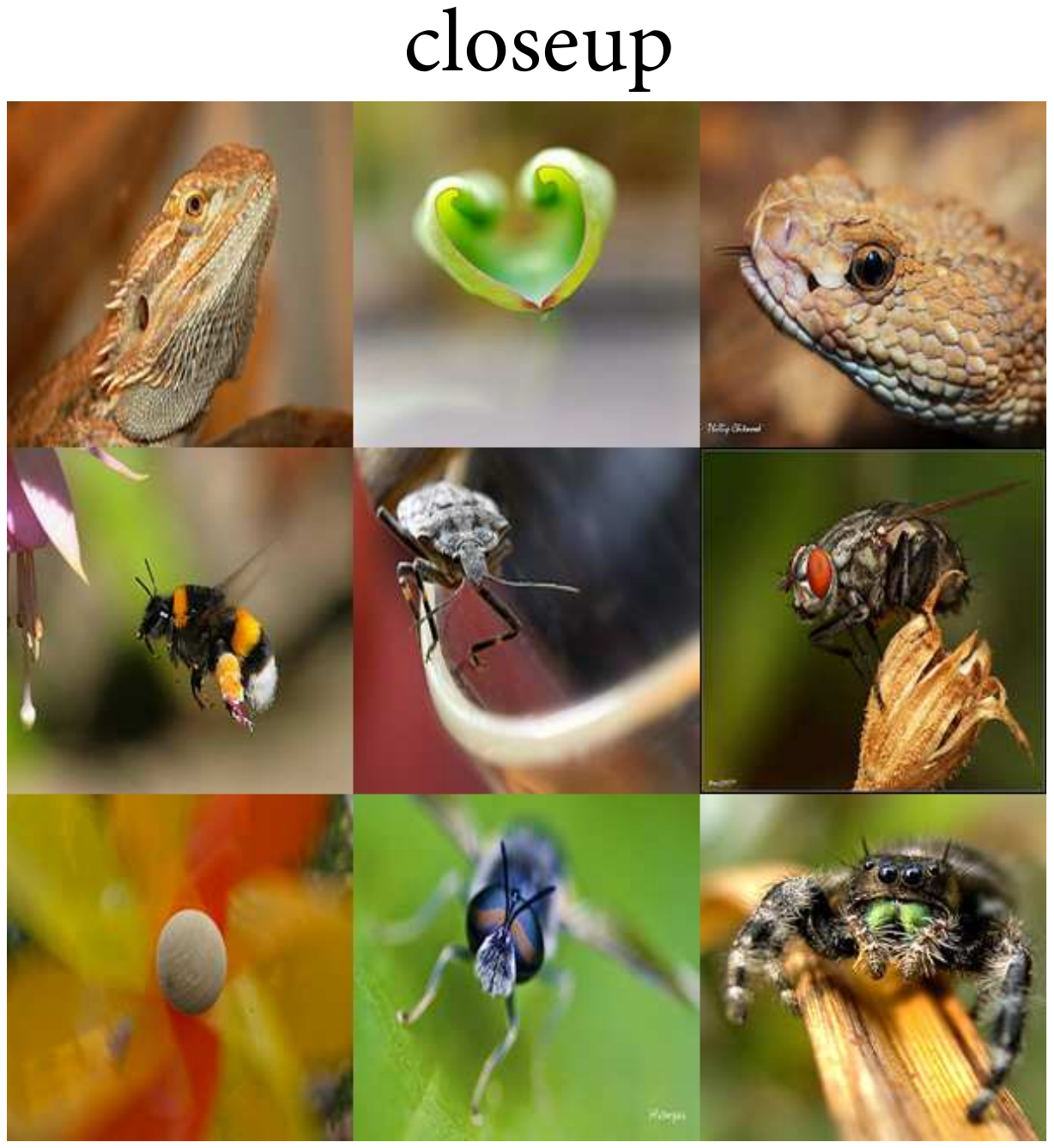}}
\subfigure[CCA (V+T).]{
 \includegraphics[width=1.5in, trim= 43mm 80mm 37mm
55mm]{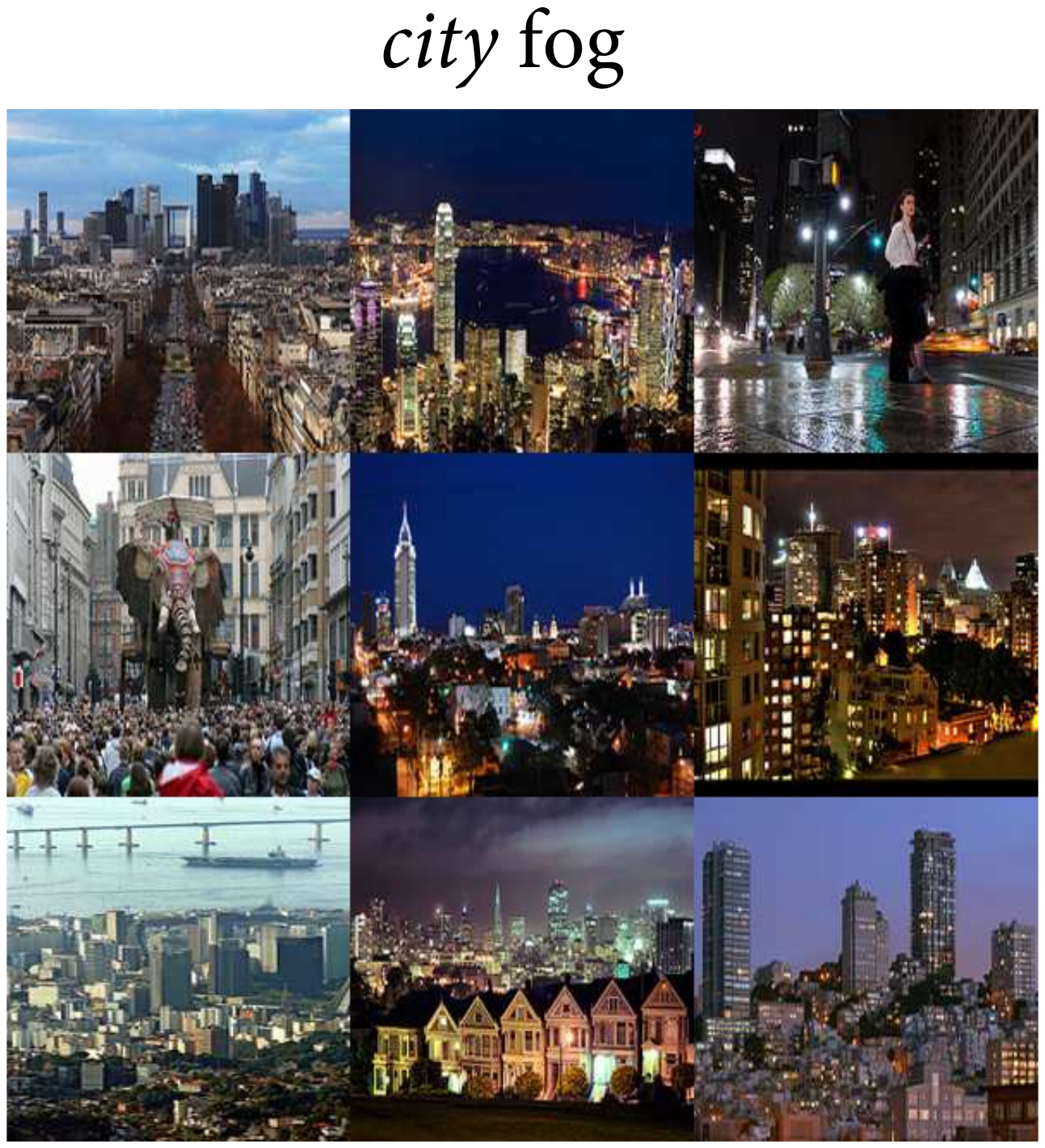}}
\subfigure[CCA (V+T+C).]{
   \includegraphics[width=1.5in, trim= 45mm 83mm 40mm
55mm]{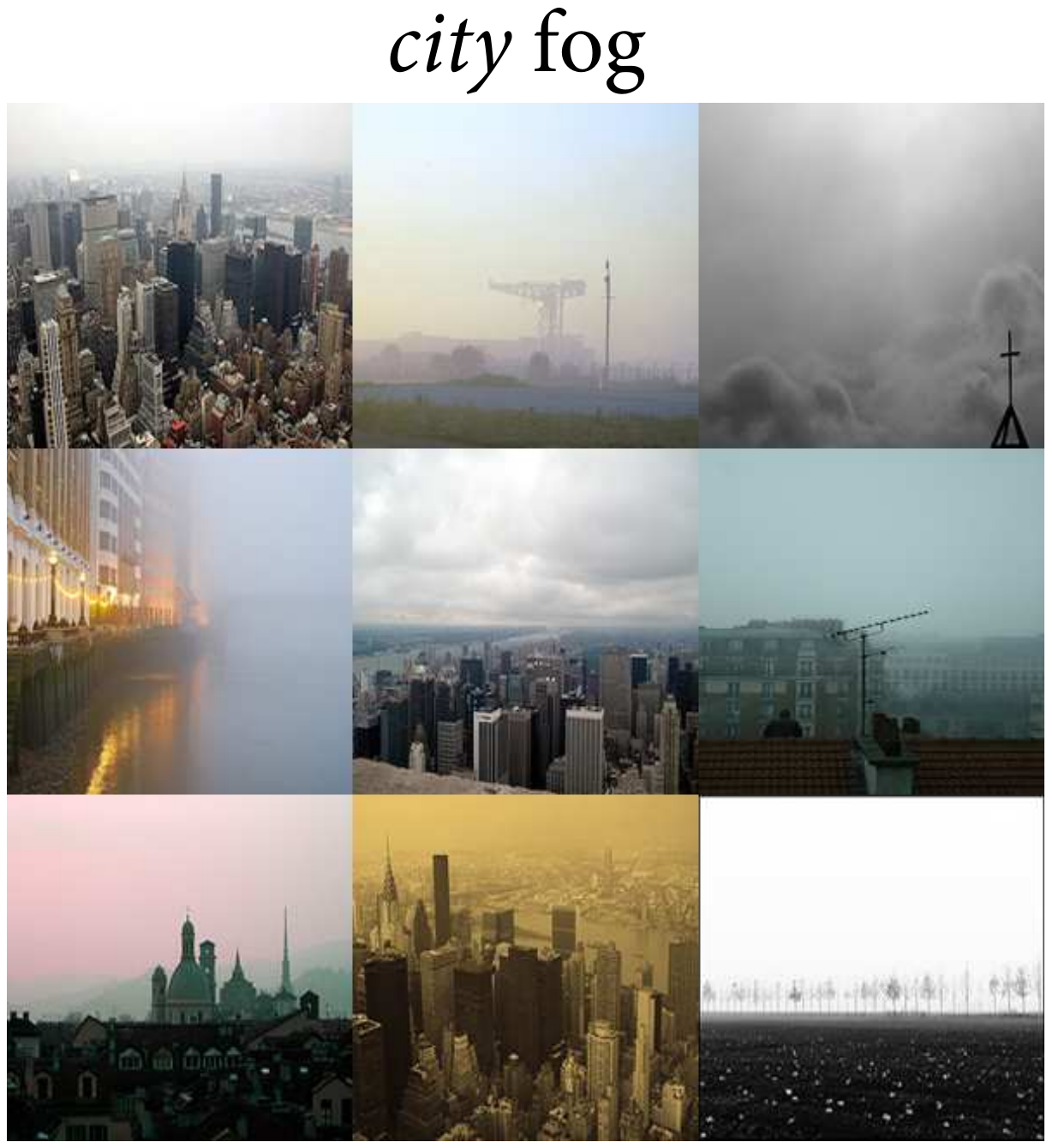}}
 \includegraphics[width=0.15in, trim=-1mm 5mm 1mm
0mm]{sample/line.png}
\subfigure[CCA (V+T).]{
   \includegraphics[width=1.5in, trim= 43mm 80mm 37mm
55mm]{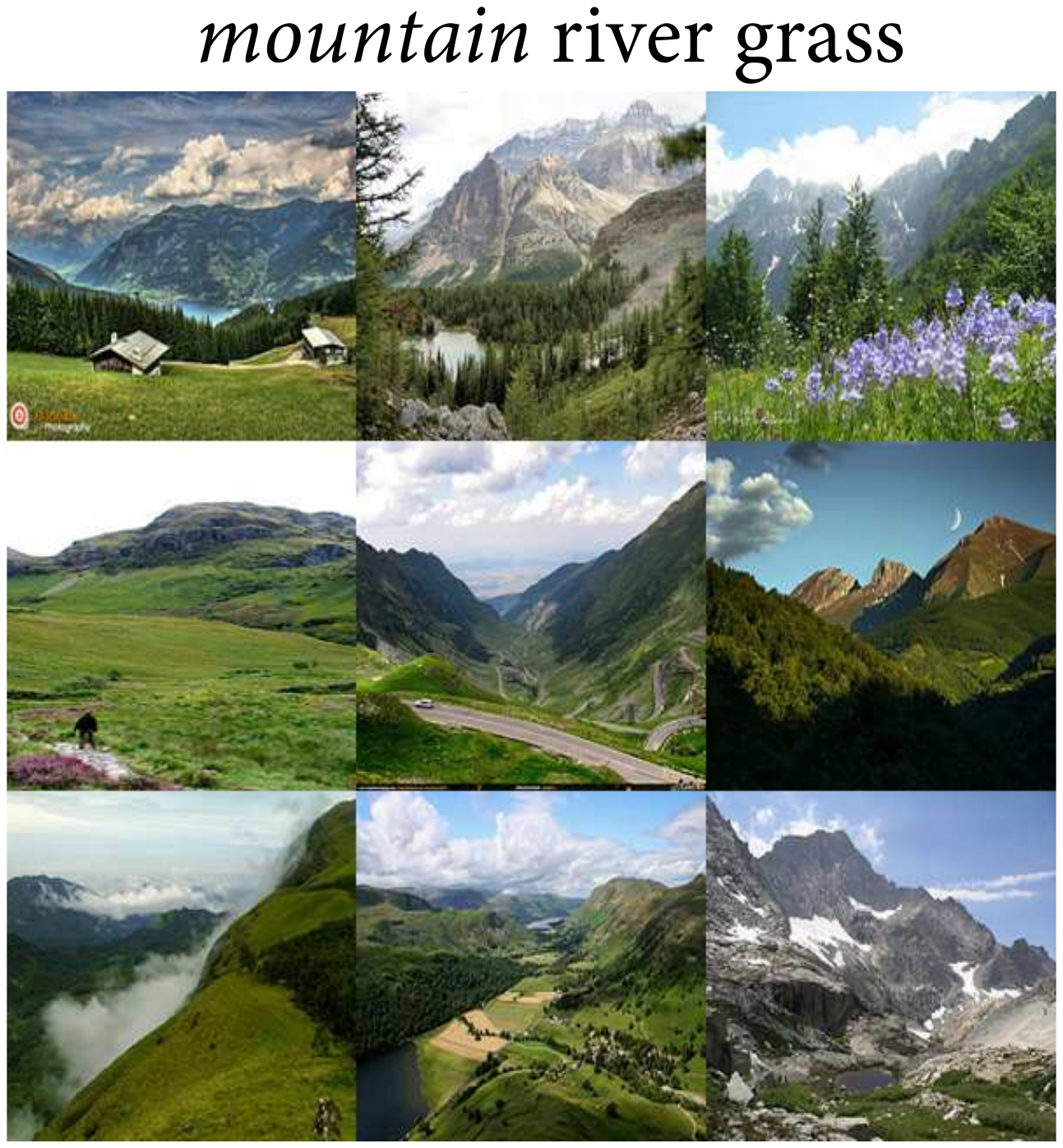}}
\subfigure[CCA (V+T+C).]{
 \includegraphics[width=1.5in, trim= 43mm 80mm 37mm
55mm]{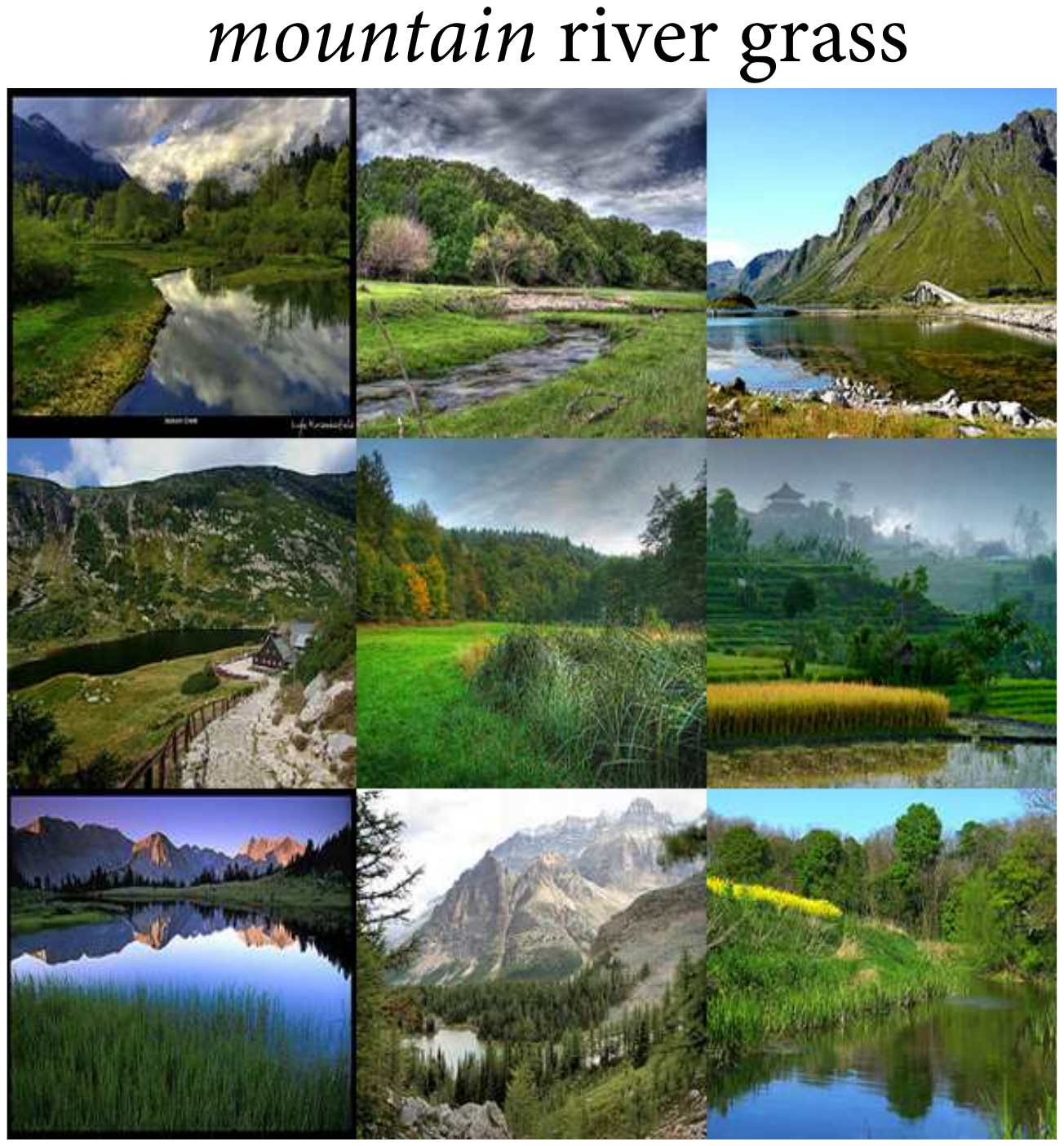}}
   \caption{A qualitative comparison of tag-to-image search for CCA (V+T) and CCA (V+T+C) on the NUS-WIDE dataset. Qualitatively, CCA (V+T+C) works better. For ``city, fog,'' the three-view model successfully finds city images with fog, while CCA (V+T) only finds city images. For ``mountain, river, grass,'' almost all images found by the three-view model contain some river, while the images found by CCA (V+T) do not contain river.}
   \label{key3}
\end{figure*}

Figure \ref{sample_i2i2nus} shows example image-to-image search results and Figure \ref{key2} shows example tag-to-image search results for the CCA (V+T+C) model.
As can be seen from the latter figure, our system can return appropriate images for compound queries consisting of combinations of as many as three tags, e.g., ``mountain, river, waterfalls'' or ``beach, people, red.'' Figure \ref{key3} compares tag-to-image retrieval results for the two-view model, CCA (V+T), and the three-view one, CCA (V+T+C). The three-view model tends to retrieve more relevant images, especially for compound queries.

\begin{figure*}[] 
   \centering
   \includegraphics[width=0.7\textwidth, trim=25mm 110mm 25mm 30mm]{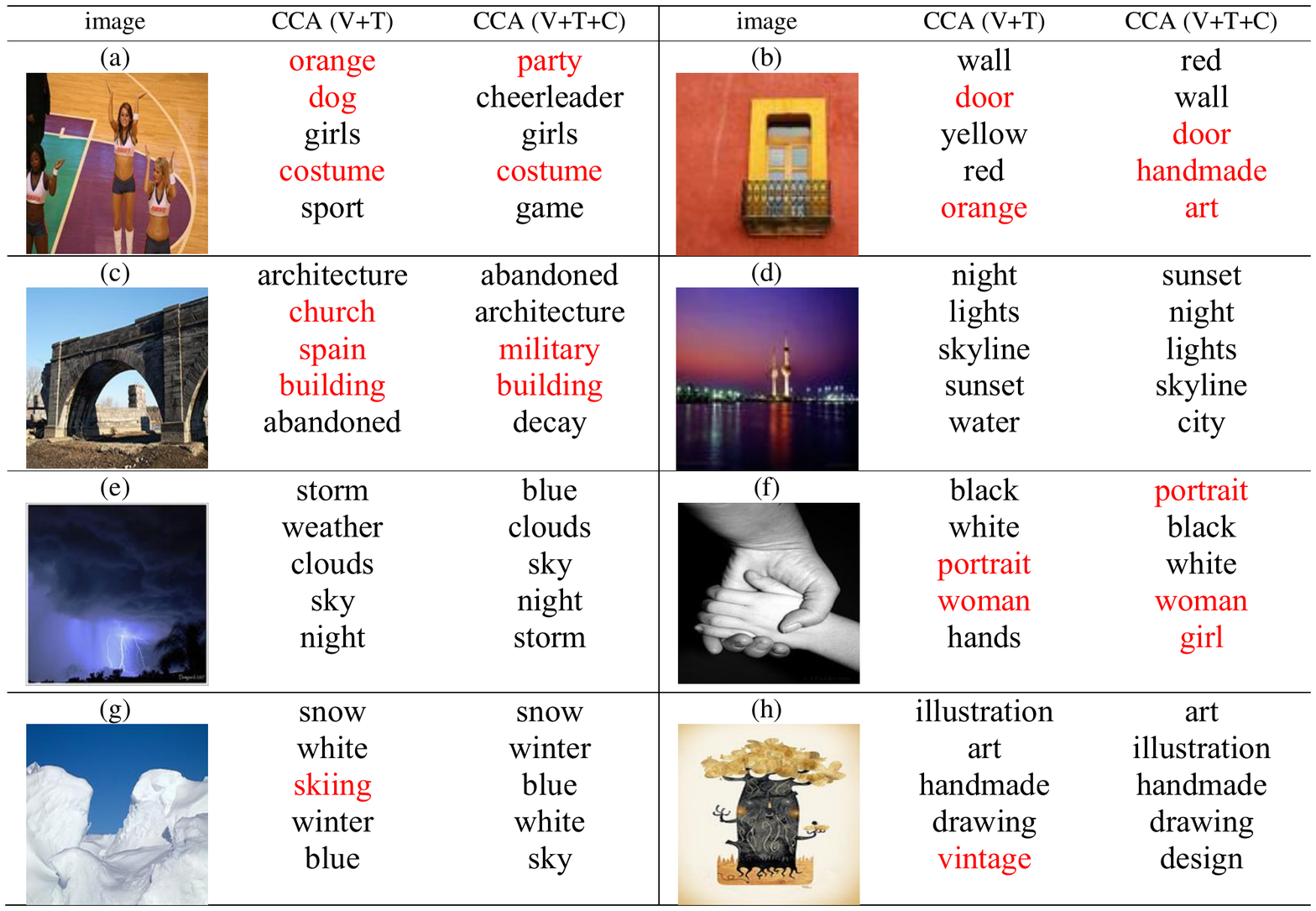}
   \caption{Example tagging results on the NUS-WIDE dataset (see text for discussion).}
   \label{taggingnus}
\end{figure*}

\begin{figure}[] 
   \centering
   \includegraphics[width=0.8\columnwidth, trim=10mm 70mm 10mm
80mm]{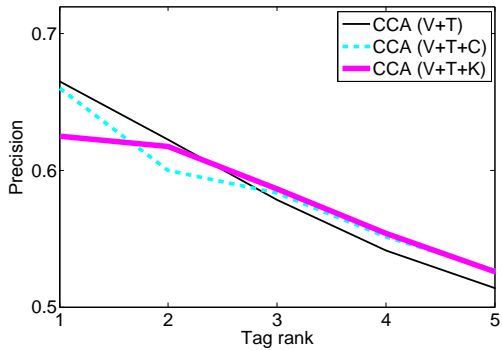}
   \caption{Tagging results on the NUS-WIDE dataset: average precision of retrieved tags vs. tag rank.}
   \label{taggingnusqual}
\end{figure}

Figure \ref{taggingnus} compares image annotation results for CCA (V+T), CCA (V+T+C), and CCA (V+T+K) using the same human evaluation protocol as in Section \ref{sec:tagging}. Unlike the Flickr-CIFAR results in Figure \ref{taggingcifar}, where the three-view models produced higher precision than CCA (V+T), all three models work comparably for image tagging on NUS-WIDE. The example results shown in Figure \ref{taggingnusqual} confirm that the subjective quality of the tags produced by two- and three-view models is similar. We believe that the explanation for this result has to do, at least in part, with the statistics of images and tags in NUS-WIDE. Specifically, many images in this dataset are either abstract or are natural landscape scenes with no distinctive objects. For such images, all our embeddings tend to suggest generic tags. Also, suggesting tags such as ``landscape,'' ``night,'' ``light,'' etc., appears to be somewhat easier than trying to suggest object-specific tags, which are much more important for Flickr-CIFAR -- indeed, in terms of absolute performance, the precision curves for NUS-WIDE (Figure \ref{taggingnusqual}) are higher than for Flickr-CIFAR (Figure \ref{taggingcifar}). Furthermore, as discussed in Section \ref{dataset}, our embedding does not provide a complete solution to the image annotation problem, as it does not include a decoding step exploiting multi-label constraints. Developing such a solution is an important subject for our future work.

\section{Results on the INRIA-Websearch Dataset \label{sec:evaluation3}}

Finally, we report results on the INRIA web search dataset. As explained in Section \ref{dataset}, ground-truth semantic information for each image in this dataset is in the form of a binary label saying whether or not that image is relevant to a particular query concept. This information directly gives us our third view for the supervised CCA (V+T+K) model. Since this dataset, just as NUS-WIDE, has relatively few images per concept, we evaluate performance using Precision@20. We randomly split the dataset into 51,478 training and 20,000 test images. In the test set, we use 18,000 images as the database, 1,000 images as validation queries, and 1,000 as test queries. Note that the database includes images marked as ``irrelevant,'' but not the validation or test queries. For CCA (V+T+C), we tune the number of NC clusters on the validation dataset to obtain 450 clusters.

\begin{figure*}[] 
   \centering
\includegraphics[width=0.65in, trim=85mm 80mm 65mm
75mm]{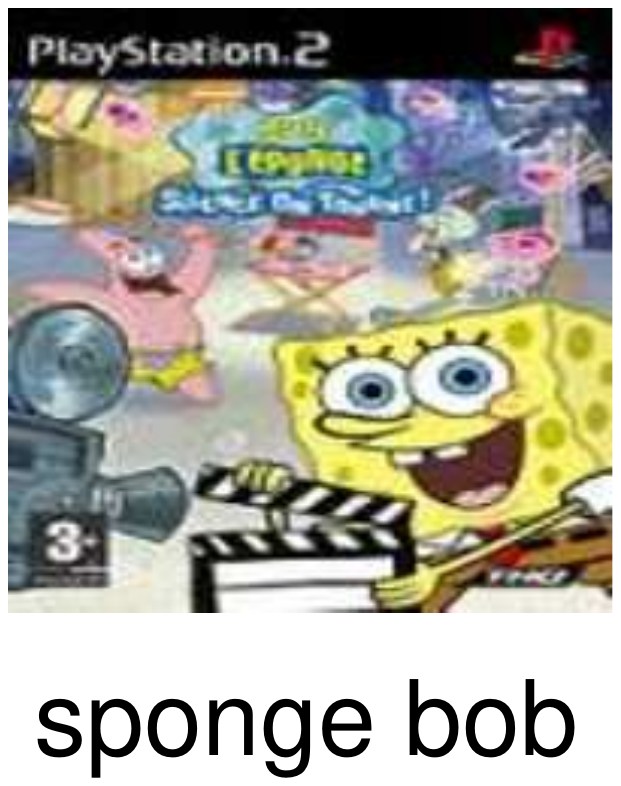}
  \includegraphics[width=1.9in, trim= 17mm 85mm 15mm
75mm]{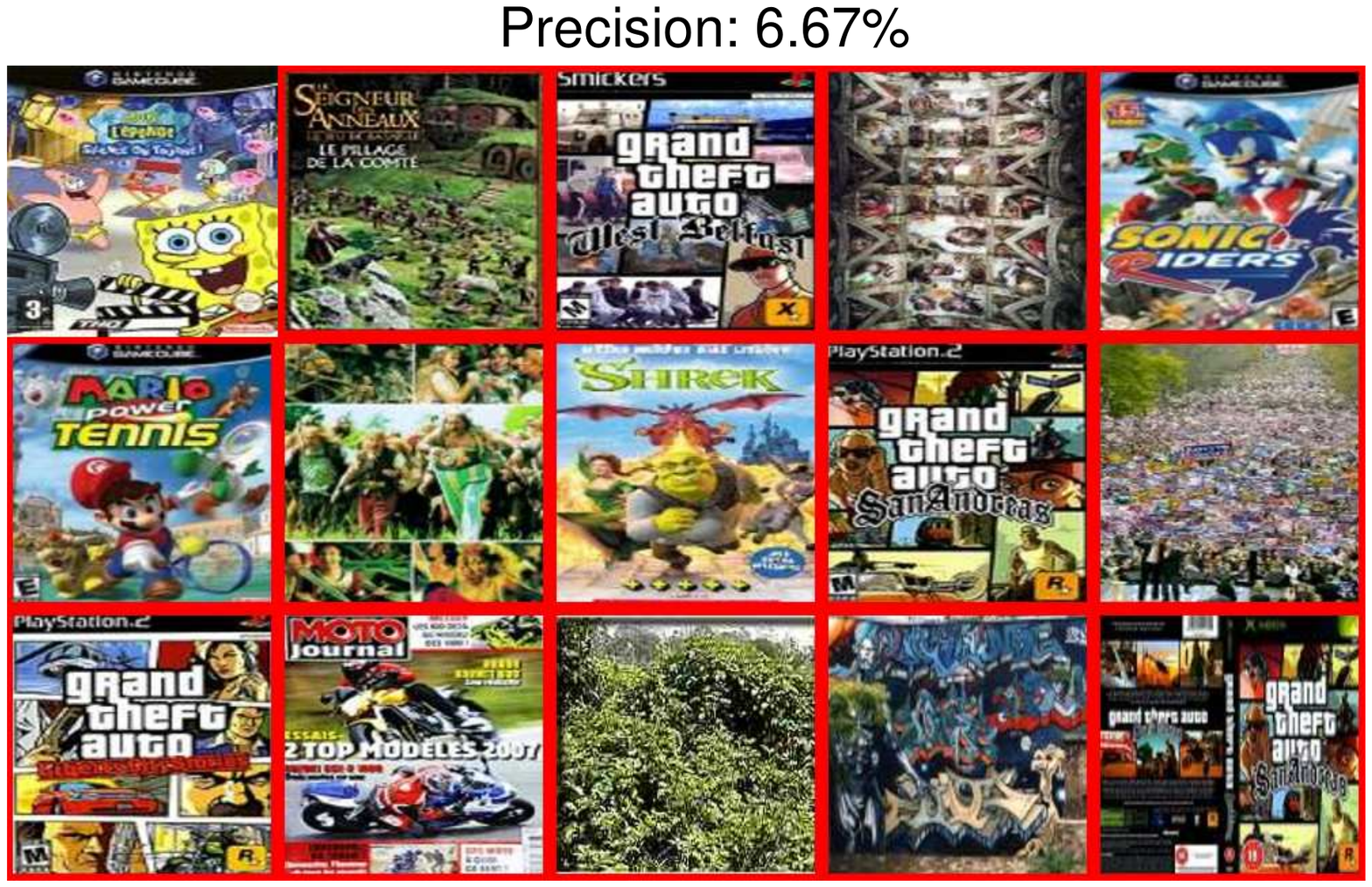}
   \includegraphics[width=1.9in, trim= 17mm 85mm 18mm
75mm]{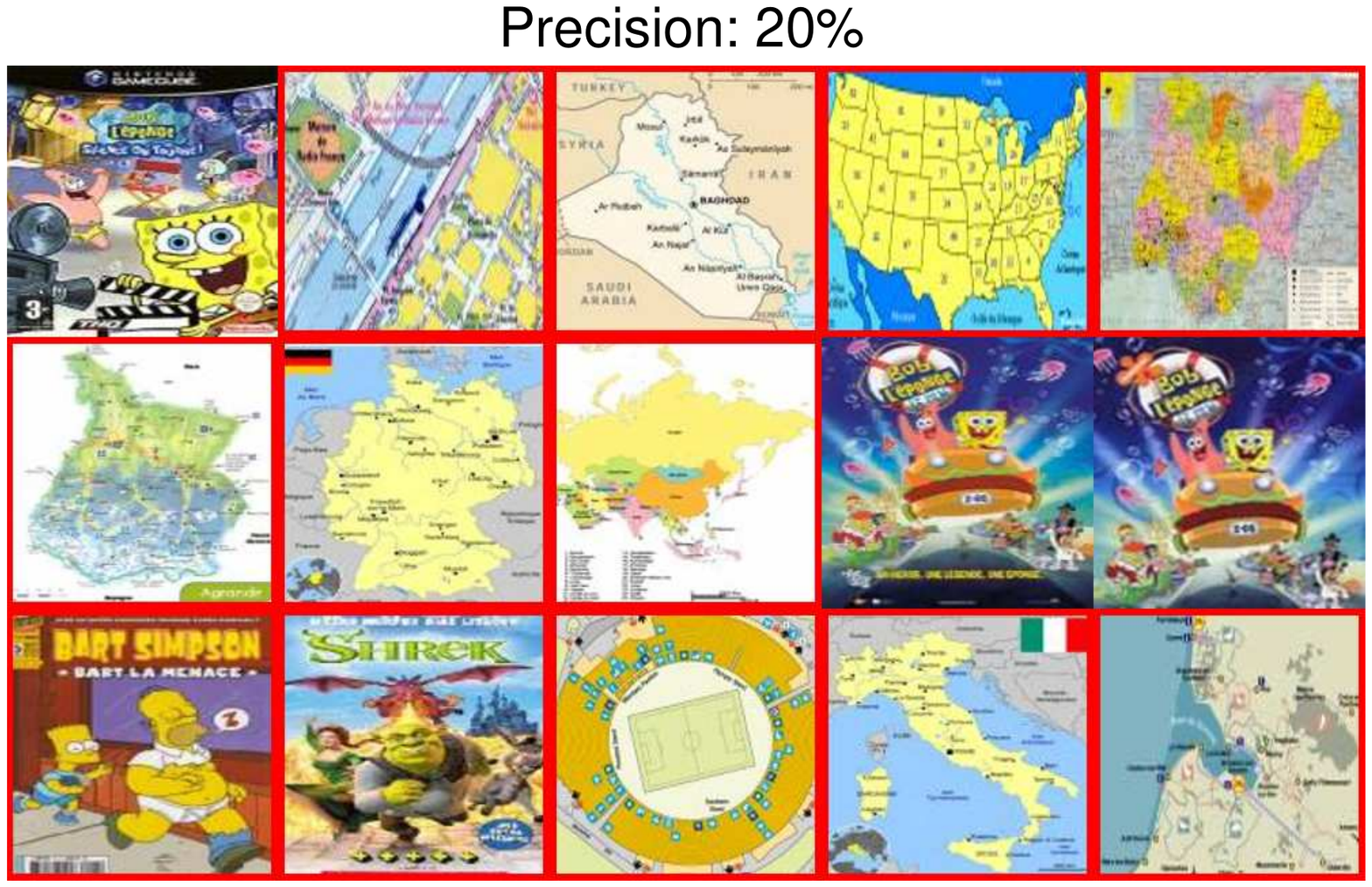}
   \includegraphics[width=1.9in, trim= 15mm 85mm 20mm
75mm]{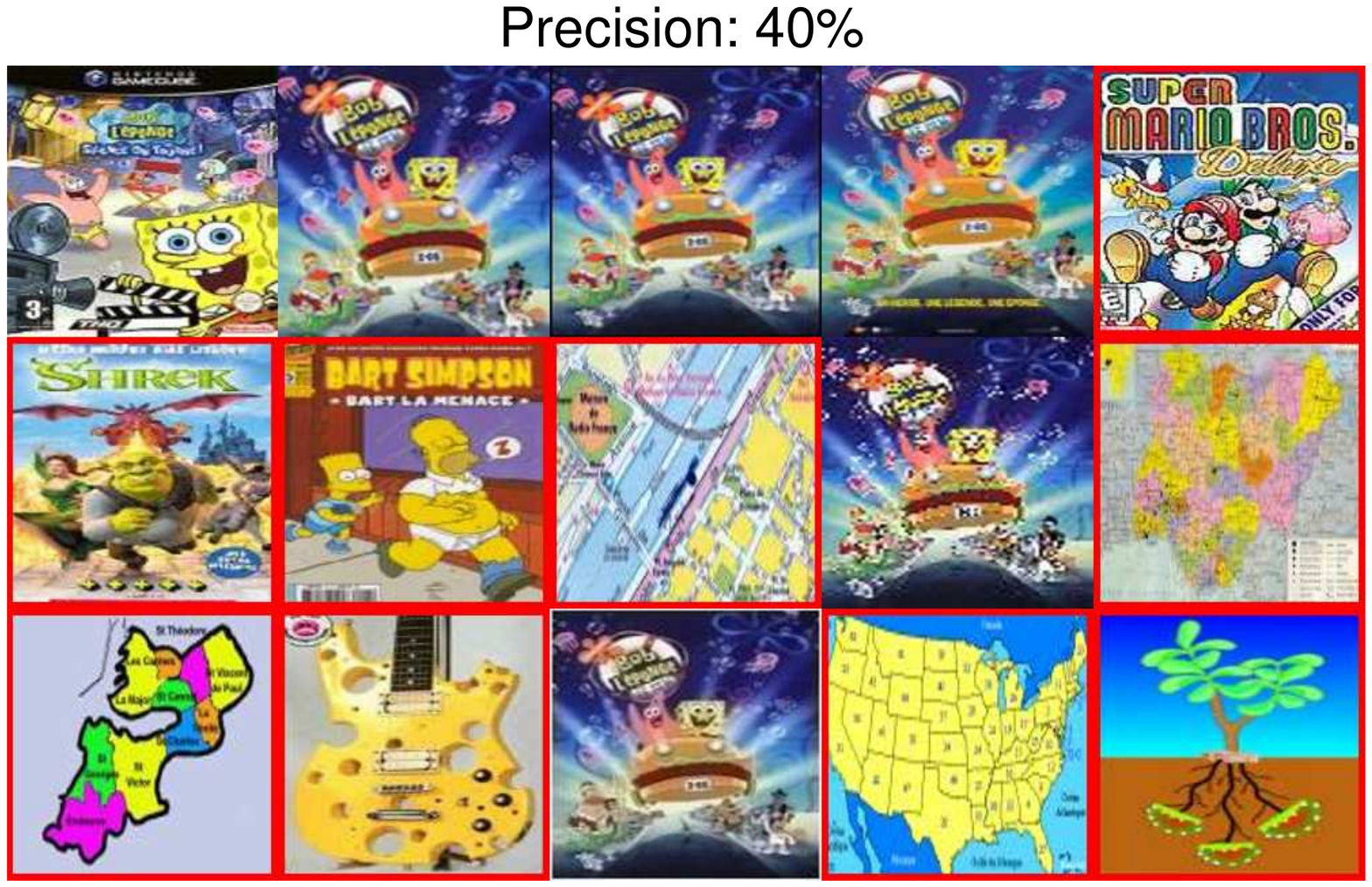}
   \includegraphics[width=0.7in, trim=75mm 80mm 75mm
75mm]{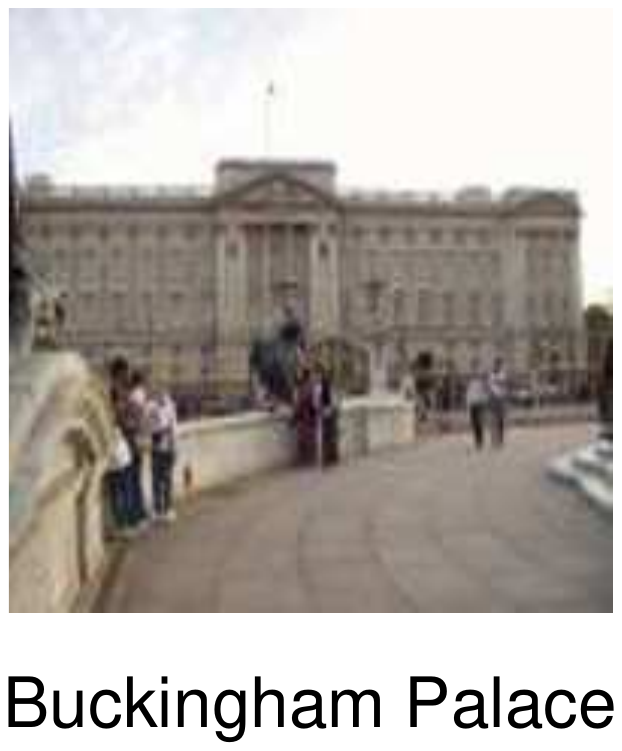}
\subfigure[Original visual feature.]{
   \includegraphics[width=1.9in, trim= 15mm 85mm 15mm
75mm]{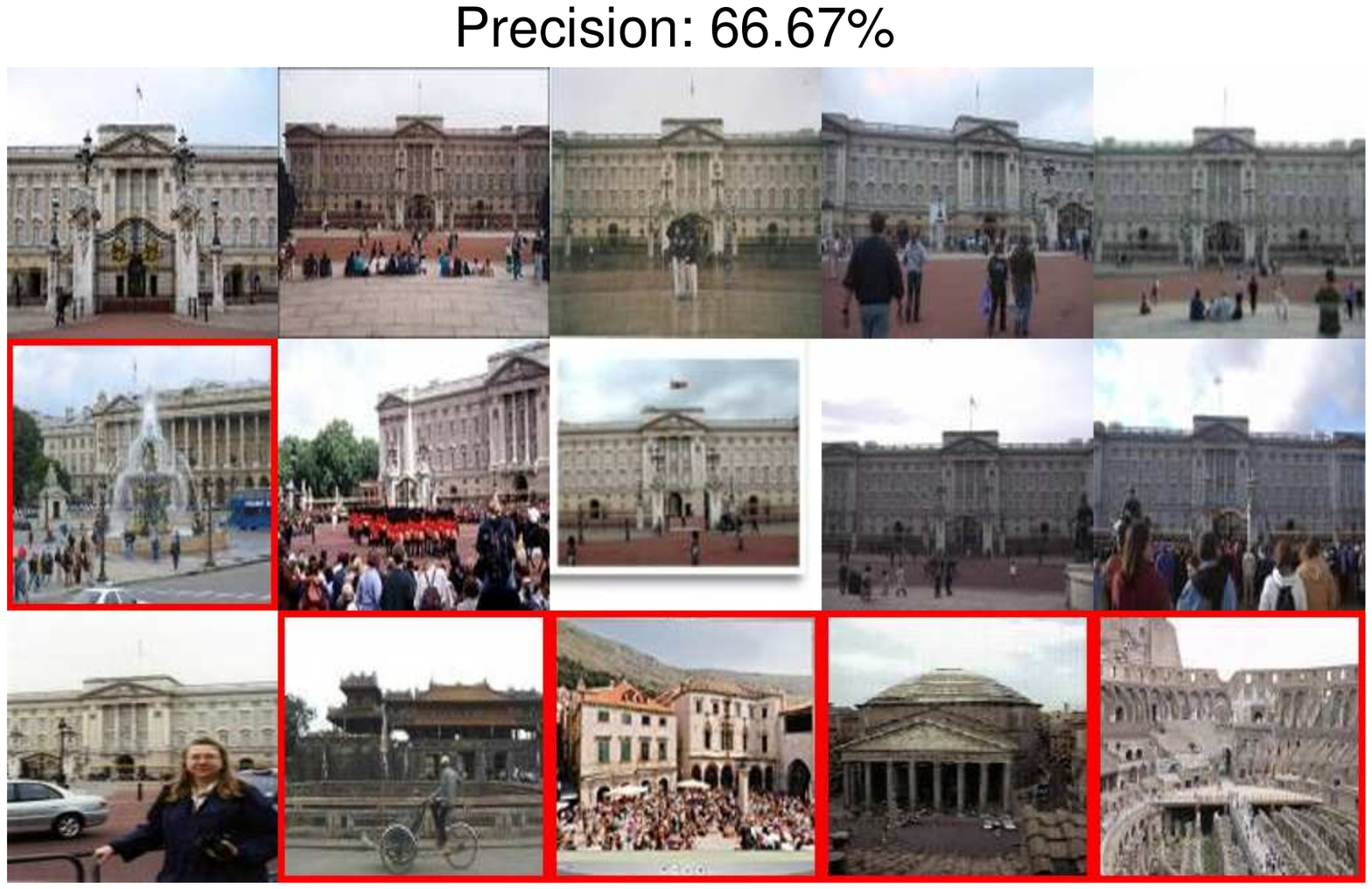}}
\subfigure[CCA (V+T).]{
   \includegraphics[width=1.9in, trim= 15mm 85mm 15mm
75mm]{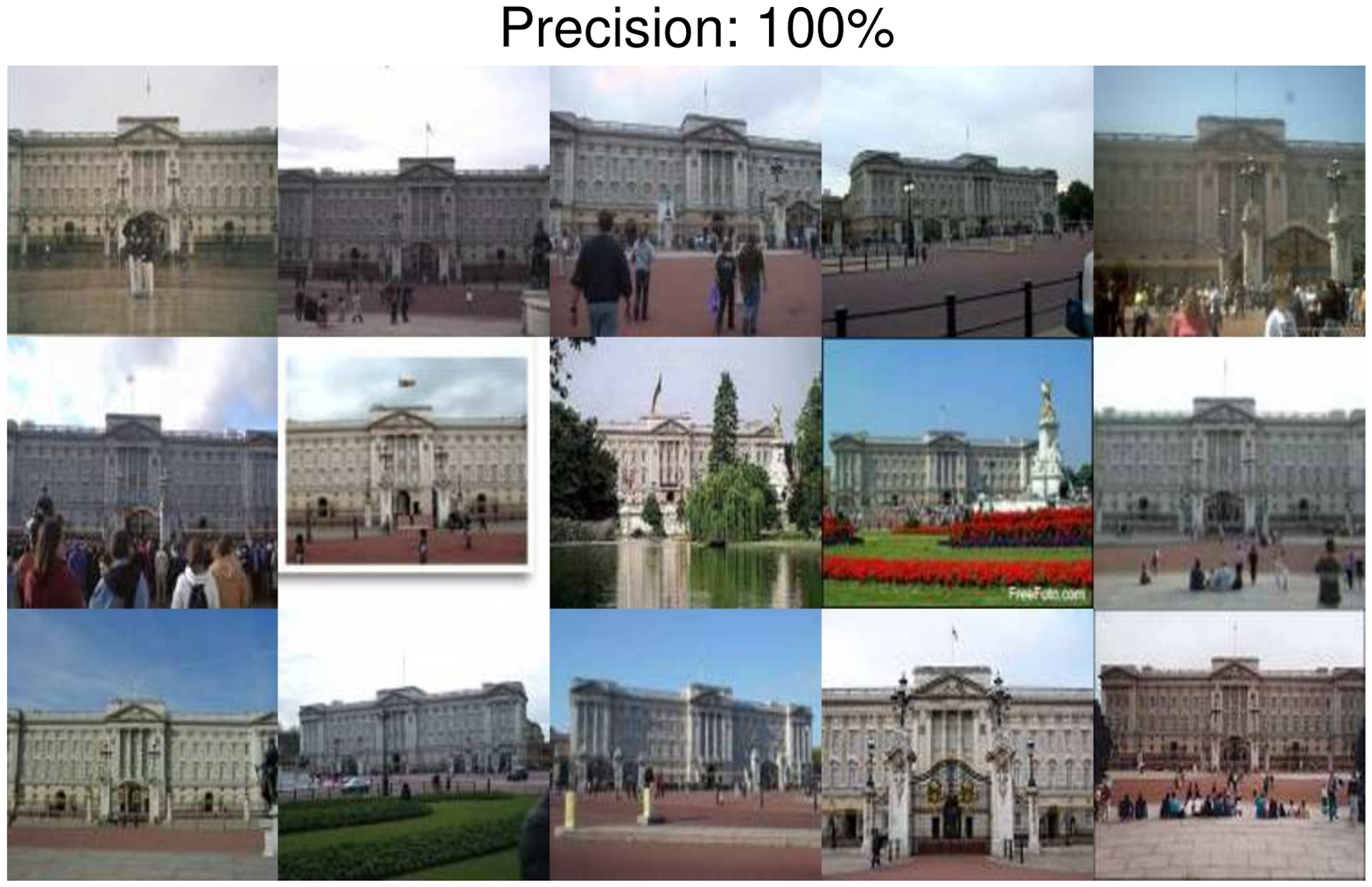}}
\subfigure[CCA (V+T+C).]{
   \includegraphics[width=1.9in, trim= 15mm 85mm 15mm
75mm]{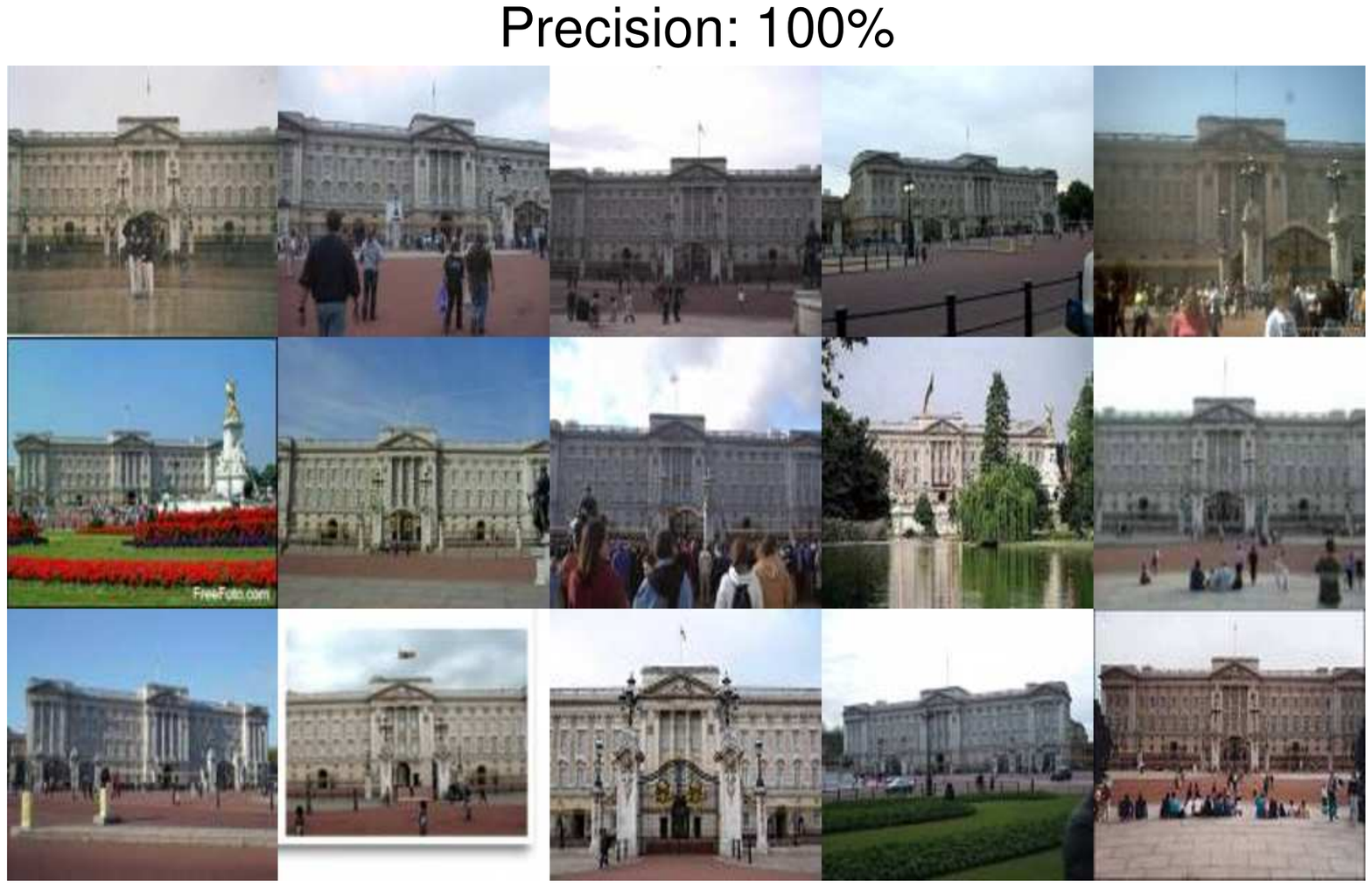}}
   \caption{Sample image-to-image retrieval results on the INRIA-Websearch dataset.  The query is on the left. Red border means false positive.}
      \label{sample_i2i2inria}
\end{figure*}

Table \ref{inria1} reports image-to-image and tag-to-image search results. As this dataset is extremely noisy and diverse, the absolute accuracy for all the methods is low. Precision may be further lowered by the fact that each database image is annotated with its relevance to just a single query concept -- thus, if a retrieved image is relevant for more than one query, this may not show up in the quantitative evaluation. Nevertheless, CCA (V+T+C) still consistently works better than the CCA (V+T) baseline.  As on the NUS-WIDE dataset, the supervised CCA (V+T+K) model works better than CCA (V+T+C).  Also, as on NUS-WIDE, CCA (V+K) works slightly better than CCA (V+T+K) for I2I. Once again, this may be because the tag view (T) is adding noise to the embedding.
Figure \ref{sample_i2i2inria} shows some qualitative image-to-image search results.

Finally, since the second view of this dataset consists not of tags, but of text mined from webpages, we do not evaluate image-to-tag search.

\section{Discussion and Future Work}

This paper has presented a multi-view embedding approach for Internet images, tags, and their semantics. We have started with the two-view visual-textual CCA model popular in several recent works \citep{gong11,cca_survey,Hwang10,hwang2011ijcv,Rasiwasia10} and shown that its performance can be significantly improved by adding a third view based on semantic ground truth labels, image search keywords, or even topics obtained by unsupervised tag clustering. In terms of quantitative results, this is our most significant finding -- both the supervised and unsupervised three-view models, CCA (V+T+K) and CCA (V+T+C), have consistently outperformed the two-view CCA (V+T) model on all three datasets, despite the extremely diverse characteristics shown by these datasets.

For the unsupervised three-view model, CCA (V+T+C), it may appear somewhat unintuitive that the third cluster-based view, which is completely derived from the second textual one, can add any useful information to improve the embedding. There are several ways to understand what the unsupervised third view is doing. Especially in simpler datasets with a few well-separated concepts, such as our Flickr-CIFAR dataset, tag clustering is actually capable of ``recovering'' the underlying class labels. Even in more diverse and ambiguous datasets with overlapping concepts, tag clustering can still find sensible concepts that impose useful high-level structure (Figure \ref{fig:NUS_semantic}). In its attempt to discover this structure, our embedding space may be likened to a non-generative version of a model that connects visual features and noisy tags to latent image-level semantics \citep{Wang09}. From another point of view, we can observe that the output of the clustering process, given by the cluster indicator matrix $C$, is a highly nonlinear transformation of the second view $T$ that either regularizes the embedding or improves its expressive power.

\begin{table}[]
{
\hfill{}
\begin{tabular}{l||cccc}
\hline
method   &    I2I        &   T2I    & K2I   \\
\hline
V-full   &        5.42     &       --   & --      \\
V   &       7.29       &       --    & --    \\
\hline
 CCA (V+T) &    12.66      &   	   25.67  & --   \\
 CCA (V+K)   &	 	  16.84    &   --  & 44.43   \\
CCA (V+T+K)   &	  {15.36}	     &     {32.76} & 41.75	         	\\
CCA (V+C)  &     13.25	          &   --   & --   \\
CCA (V+T+C)  &     {13.61}	        &    {29.57}	& --         \\
\hline
Structural learning  &    8.35	         &       --   & --       \\
Wsabie & 	10.01 		 & 	 --     &    --  \\
\hline
\end{tabular}}
\hfill{}
\caption{Precision@20 for different multi-view models on the INRIA-Websearch dataset. For K2I, the queries correspond to the 353 ground truth concepts. Note that these concepts are no longer necessarily part of the tag vocabulary, so we cannot report K2I results for any embedding that does not include the K view. We have obtained standard deviations from five random database/query splits, and they are around 0.6\% - 1.1\%.
}
\label{inria1}
\end{table}

The quantitative and qualitative results presented in this paper demonstrate that our proposed multi-view embedding space, together with the similarity function specially designed for it, successfully captures visual and semantic consistency in diverse, large-scale datasets. This space can form a good basis for a scalable and flexible retrieval system capable of simultaneously accommodating multiple usage scenarios. The visual and semantic clusters discovered by tag clustering and subsequent CCA projection can be used to summarize and browse the content of Internet photo collections \citep{Berg09,Raguram08}. Figure \ref{fig:NUS_semantic} has shown an example of what such a summary could look like. Furthermore, users can search with images for similar images, or retrieve images based on queries consisting of multiple tags or keywords. As illustrated in Figure \ref{weight}, they can also manually adjust weights corresponding to different keywords according to the importance of those keywords. Finally, our embedding space can also serve as a basis for an automatic image annotation system. However, as discussed in Section \ref{sec:evaluation2}, in order to achieve satisfactory results on this task, we need to develop more sophisticated decoding methods incorporating multi-label consistency constraints.

Besides the application scenarios named above, we are also interested in using our learned latent space as an intermediate representation for recognition tasks. One of these is nonparametric image parsing \citep{liu10,tighe10} where, given a query image, a small number of similar training images is retrieved and labels are transferred from these images to the query. With a better embedding for images and tags, this retrieval step may be able to return training images more consistent with the query and lead to improved accuracy for image parsing. Another problem of interest to us is describing images with sentences \citep{Farhadi10,Kulkarni11,Ordonez11}. Once again, with a good intermediate embedding space linking images and tags, the subsequent step of sentence generation may become easier. \vspace{-20pt}

\vspace{20pt}

\begin{acknowledgements}
We would like to thank the anonymous reviewers for their constructive comments; Jason Weston for advice on implementing the Wsabie method;
Albert Gordo and Florent Perronnin for useful discussions; and Joseph Tighe, Hongtao Huang, Juan Caicedo, and Mariyam Khalid for helping
with manual evaluation of the auto-tagging experiments.
Gong and Lazebnik were supported by NSF
grant IIS 1228082, DARPA Computer Science Study Group
(D12AP00305), and Microsoft Research Faculty Fellowship.
\end{acknowledgements}



\end{document}